\documentclass[11pt]{article}

\usepackage[utf8]{inputenc}
\usepackage[T1]{fontenc}

\usepackage{amsmath}
\usepackage{amsthm}
\usepackage{mathtools}
\usepackage{bm}          % \bm for bold math (used by \mathbold)
\usepackage{upgreek}     % \Upsigma for aggregation operators

\usepackage[preprint]{jmlr2e}

\usepackage{array}
\usepackage{booktabs}
\usepackage{multirow}
\usepackage{rotating}

\usepackage{url}
\usepackage[makeroom]{cancel}
\usepackage{needspace}
\newenvironment{proofsketch}{\par\noindent{\bf Proof sketch\ }}{\hfill\BlackBox\\[2mm]}

\ShortHeadings{Discrimination, Calibration, and Ranking}{Prati and Johari}
\firstpageno{1}

\providecommand{\mathbold}[1]{\bm{#1}}

\newtheorem{mydef}[theorem]{Definition}  % alias for compatibility with section files

\title{Discrimination, Calibration, and Ranking:\\
       A Unified Matrix Framework}

\author{\name Ronaldo C. Prati \email ronaldo.prati@ufabc.edu.br \\
        \addr Centro de Matemática, Computação e Cognição\\
        Universidade Federal do ABC (UFABC)\\
        Santo André, SP, Brazil
        \AND
        \name Farangis Johari \email farangis.johari@ufabc.edu.br \\
        \addr Universidade Federal do ABC (UFABC)\\
        Santo André, SP, Brazil}

\editor{TBD}

\begin{document}
% ============================================================

\maketitle

\begin{abstract}
Discrimination, calibration, and ranking are the three pillars of classifier
evaluation, yet no single algebraic language spans all three.
A recent indicator-matrix account unifies discrimination measures (precision,
recall, $F_1$, Matthews correlation coefficient) across binary, multiclass,
multilabel, and related classification settings, via Boolean cell operations
on indicator label matrices, generalised to soft labels by $t$-norms;
calibration measures (Brier score, binary log-loss as an extended-valued
boundary case) and ranking measures (nDCG, AUC) fall outside it.
This paper closes the gap with two further operator classes:
\emph{deviation operators} $\phi(y - \hat{y})$, yielding a Calibration Framework,
and rank- and slice-dependent operators, including row-positional gains
$Y_{ij}w(\mathrm{rank}_i(j))$ and label-dependent column concordance cells for AUC,
yielding a Ranking Framework; both inherit the micro/macro/exemplar aggregation architecture.
For calibration, the three Brier aggregations collapse to a single
scalar through uniform averaging of the per-cell losses; a matrix-language
Murphy decomposition holds, and within the Brier-type class-indexed
family, a measure is skew-invariant if and only if it depends on the
class-conditional errors $(\alpha_j,\beta_j)$ alone, the calibration
analogues of recall and specificity.
For ranking, raw DCG collapses across aggregation schemes while normalised
nDCG does not; per-class nDCG and per-class AUC are unified by a row--column
duality theorem; and raw positional scores are never strictly proper and are weakly proper among distinct-entry reports when the true marginals are also distinct; with min-rank ties, weak properness can fail (constant discounts are an exception), whereas normalised nDCG and AUC need not be proper.
Across all three frameworks, the skew-invariant per-class building blocks share a class-conditional expectation template over the appropriate positive- or negative-label event, under transformations preserving the conditional distribution of the averaged cell quality. This covers recall, specificity, per-class AUC, per-class nDCG, and $(\alpha_j,\beta_j)$.
Experiments on two multilabel benchmarks illustrate these theoretical distinctions.
\end{abstract}

\begin{keywords}
classifier evaluation, confusion matrix, Brier score, nDCG, AUC,
cell operators, skew-invariance, Murphy decomposition, multilabel learning
\end{keywords}

% ============================================================
%  Section 1 — Introduction
% ============================================================
\section{Introduction}
\label{sec:intro}

Evaluating the performance of classification models is a core step in any empirical machine learning study~\citep{japkowicz2011evaluating}. Many classifiers output scores in $[0,1]$, often interpreted as estimated class probabilities; we collect such scores in a matrix $\mathbold{\hat{P}} \in [0,1]^{n \times m}$ before a decision rule is applied. This output carries distinct kinds of information. First, binary decisions obtained using a specified decision rule determine threshold-dependent discrimination measures such as $F_1$, precision, and recall; the relative ordering of scores alone does not determine these measures. Second, the magnitudes of probability estimates matter for probabilistic predictive quality: calibration concerns whether a forecast such as $\hat{P}_{i,j}=0.8$ corresponds to an event frequency near $80\%$, while proper scoring rules such as Brier and log-loss assess overall probabilistic quality, including calibration and resolution rather than calibration alone. A third, distinct question concerns \emph{ranking}: row-axis nDCG evaluates the positions of relevant labels among the labels of each example, while column-axis AUC evaluates the ordering of positive versus negative examples for a fixed class. All three evaluation paradigms are active in most empirical study of multilabel or probability-producing classifiers, yet they have developed largely in isolation, with separate literatures, separate intuitions, and no common algebraic language.

A unified algebraic framework for discrimination was recently proposed by \citet{prati2026unified}. It represents actual and predicted labels as binary indicator matrices $\mathbold{Y}, \mathbold{\hat{Y}} \in \{0,1\}^{n \times m}$ and defines cell-level true/false positive/negative counts via element-wise Boolean conjunction and negation; soft labels and predictions are accommodated by replacing the Boolean connectives with triangular norms ($t$-norms), which reduce to the Boolean case on binary matrices. Three aggregation operators, global ($\Upsigma_{\mathrm{1}}$), column-wise ($\Upsigma_{\mathrm{m}}$), and row-wise ($\Upsigma_{\mathrm{n}}$), corresponding to micro, macro, and exemplar averaging, then generate entire families of measures for multiclass, multilabel and related classification problems from any single binary formula. The framework characterises skew-invariant measures (functions of recall and specificity only), proves that micro-averaging equals denominator-weighted macro-averaging, and identifies the product $t$-norm as the unique marginally consistent one (which implies the confusion-matrix partition property). However, calibration measures (Brier score, log-loss, Murphy decomposition) and nDCG fall outside the framework of \citet{prati2026unified}: they operate on the raw probability matrix $\mathbold{\hat{P}}$ rather than a thresholded indicator matrix, and their cell values are not expressible as Boolean or $t$-norm outputs. AUC is reachable there only as an integral over thresholds of the coincidence cells.

%\subsection*{Contributions}

This paper extends the indicator-matrix framework by replacing its coincidence cell operator (the Boolean conjunction and its $t$-norm generalisation) with two further operator classes, yielding a three-class taxonomy:

\begin{enumerate}
  \item \emph{Coincidence operators} (Boolean conjunction, generalised by $t$-norms): the cell operator of  \citet{prati2026unified}, quantifying Boolean coincidence of actual and predicted binary labels, or their soft overlap according to a chosen $t$-norm; for soft entries, positivity of both inputs alone need not imply a positive $t$-norm value. These recover the Discrimination Framework of \citet{prati2026unified} with no modification.

  \item \emph{Deviation operators} ($\phi(y - \hat{y})$ for convex $\phi$): measuring the magnitude of the gap between actual and predicted soft labels. This yields the \emph{Calibration Framework},  covering calibration.

  \item \emph{Positional and rank-dependent operators}: row-positional gains $Y_{ij}w(\mathrm{rank}_i(j))$, with non-increasing $w$, give the nDCG family; column-dependent cells compare positive and negative examples of a class to give AUC. A fixed column-rank weight yields an affine identity for aggregated AUC under distinct scores, but does not generally reproduce an individual AUC concordance cell. The row and column ranking axes produce complementary subfamilies of the \emph{Ranking Framework}.
\end{enumerate}

Within this taxonomy we establish the following results:

\begin{itemize}
  \item  For the Calibration Framework, the squared-error deviation operator $D(y,\hat{y}) = (y - \hat{y})^2$ induces the Brier score family. Applying the three aggregation operators yields micro-, macro-, and exemplar-averaged Brier scores; we prove that all three collapse to the same scalar, because the Brier score is a sum of per-cell terms with a fixed denominator (the same holds for any per-cell deviation, but not for the ratio measures of the Discrimination Framework or for normalised nDCG). We derive a matrix-language version of the Murphy~\citep{murphy1973new} reliability--resolution--uncertainty decomposition, $\mathrm{BS}_j = \mathrm{UNC}_j - \mathrm{RES}_j + \mathrm{REL}_j$, via the tower property of conditional expectations. The same identity is also stated for a calibration map estimated by isotonic regression, which needs no bins (Proposition~\ref{prop:murphy_isotonic}).
  
  \item We characterise proper scoring rules within the Calibration Framework via the Savage representation~\citep{savage1971elicitation,gneiting2007strictly}, showing that a symmetric deviation operator $\phi(y-\hat{y})$ is proper exactly when its weight $\phi'(t)/t$ is symmetric about $\tfrac12$ (convexity of $\phi$ alone is not sufficient); the proper ones include squared error and nonquadratic losses, while log-loss satisfies the same weight condition in an extended-valued sense. For skew-sensitivity, we prove that a Brier-type class-indexed measure of the form $M=g(\{\pi_j,\alpha_j,\beta_j\}_{j=1}^{m})$ is skew-invariant if and only if it depends on $\{\alpha_j,\beta_j\}_{j=1}^{m}$ alone, where $\alpha_j = \mathbb{E}[(1-\hat{Y}_{ij})^2 \mid Y_{ij}=1]$ and $\beta_j = \mathbb{E}[\hat{Y}_{ij}^2 \mid Y_{ij}=0]$ are the class-conditional squared errors, analogous to the complements of recall and specificity in the hard-prediction limit (and, in the binary case, the two components of the stratified Brier score~\citep{wallace2012class,wallace2014improving}). General calibration measures are skew-invariant exactly when they depend only on the within-class score distributions (Theorem~\ref{thm:cal_skew}(a)).
  
  \item For the Ranking Framework, the row-dependent positional operator $D_w(Y_{i,j}, \hat{\mathbold{Y}}_{i,\cdot}) = Y_{i,j} \cdot w(\mathrm{rank}_i(j))$ induces the nDCG family. We show that raw DCG collapses across aggregation schemes (micro = macro = exemplar) while normalised nDCG does not, due to heterogeneous IDCG denominators. We propose a per-class nDCG summary within this matrix formalism, with the conditional-mean interpretation: $\mathrm{nDCG}_j^{\mathrm{class}} = \frac{1}{w(1)}\mathbb{E}[w(\mathrm{rank}_i(j)) \mid Y_{ij}=1]$. The column-dependent AUC family instead uses the concordance cell $A_{ij}=Y_{ij}f(i,j)$, where $f(i,j)$ is the fraction of negative examples for class $j$ scored below example $i$ (with half credit for ties). Unlike a fixed positional cell $Y_{ij}v(\mathrm{rank}_j(i))$, its value also depends on the negative-label set. At the per-class aggregate level, AUC is the Wilcoxon--Mann--Whitney statistic~\citep{hanley1982meaning} and, for distinct column scores, an affine function of the mean positive-example rank (Proposition~\ref{prop:auc_meanrank}). This construction yields the per-example $\mathrm{AUC}_i^{\mathrm{ex}}$ summary and makes the per-class skew-invariance transparent. A \emph{row--column duality theorem} unifies the two subfamilies: both $\mathrm{nDCG}_j^{\mathrm{class}}$ and $\mathrm{AUC}_j$ are conditional means over positive examples of rank-based quality measures, with a fixed row-rank discount for nDCG and a label-dependent column-concordance fraction for AUC (whose aggregate is an affine function of mean column rank).
  
  \item Finally, as a cross-framework synthesis, we show that the three skew-invariant per-class bases, namely recall/specificity (Discrimination Framework), $(\alpha_j, \beta_j)$ (Calibration Framework), and $\mathrm{nDCG}_j^{\mathrm{class}}$/$\mathrm{AUC}_j$ (Ranking Framework), all share the same algebraic structure: they are class-conditional means (over the positives for recall, $\alpha_j$, per-class nDCG and AUC; over the negatives for specificity and $\beta_j$). This single template organises the skew-invariance proofs for all three frameworks, under reweightings that preserve the conditional distribution of the averaged cell quality (for per-class nDCG this includes the within-row ranks of the positives, not only each label's score distribution), and reveals that pooled (micro) aggregates are skew-sensitive in general (for the ratio measures of the Discrimination Framework and for micro-AUC because they weight classes by prevalence-dependent denominators; for the Brier score because each per-class term is itself prevalence-dependent). We further argue that the row/column dependency split is specific to the Ranking Framework: the natural ways of introducing rank dependence into the Discrimination or Calibration Frameworks either trivialise or recover the positional operator.
\end{itemize}

The measures of the Calibration and Ranking Frameworks, together with diagnostics that report the covariances behind the gaps between the micro, class and exemplar summaries, are implemented in the experimental subpackage \texttt{aggmetrics.scores} of the \texttt{aggmetrics} package (\url{https://github.com/rcprati/aggmetrics}), available from version~0.3.0.

%\subsection*{Organisation}

This paper is organised as follows. Section~\ref{sec:background} reviews the framework of \citet{prati2026unified}. Section~\ref{sec:related} reviews related work. Section~\ref{sec:taxonomy} introduces the three-class cell operator taxonomy and the framework bridge theorem connecting the hard-prediction limits of the Discrimination and Calibration Frameworks; the hard-prediction limit of the Ranking Framework is treated in Section~\ref{sec:ranking}. Section~\ref{sec:calibration} develops the Calibration Framework in full. Section~\ref{sec:ranking} develops the Ranking Framework, including both the nDCG and AUC subfamilies and the duality theorem. Section~\ref{sec:synthesis} synthesises the three frameworks. Section~\ref{sec:experiments} presents empirical illustrations. Section~\ref{sec:conclusion} concludes.

% ============================================================
%  Section 2 — Background
% ============================================================
\section{Background: The Indicator-Matrix Framework}
\label{sec:background}

This section provides a self-contained review of the base framework of \citet{prati2026unified}, which covers discrimination-based evaluation. Readers familiar with that paper may skip ahead to Section~\ref{sec:taxonomy}.

\subsection{Indicator Matrices}
\label{sec:bg_matrices}

Consider a dataset of $n$ examples, each associated with a subset of $m$ labels. The actual and predicted label assignments are represented as binary matrices.

\begin{mydef}[Indicator matrices]
\label{def:indicator}
The \emph{actual label matrix} $\mathbold{Y} \in \{0,1\}^{n\times m}$ has $Y_{i,j} = 1$ iff example $e_i$ belongs to class or label $l_j$. The \emph{predicted label matrix} $\mathbold{\hat{Y}} \in \{0,1\}^{n \times m}$ has $\hat{Y}_{i,j} = 1$ iff the classifier assigns label $l_j$ to $e_i$.

In \emph{multiclass} problems, each row of $\mathbold{Y}$ (and $\mathbold{\hat{Y}}$) is a one-hot vector: exactly one entry equals 1. In \emph{multilabel} problems, any number of entries per row may equal 1. The binary case is the special case $m = 2$.
\end{mydef}

A binary task can be encoded by one indicator column for a designated positive class or by two complementary one-hot columns; the encodings carry the same label information, but scoring the positive class alone differs from pooling or macro-averaging both columns (Appendix~\ref{app:bg_examples}).

\Needspace{9\baselineskip}
A soft classifier outputs a probability matrix $\mathbold{\hat{P}} \in [0,1]^{n \times m}$; a decision rule converts it into the binary $\mathbold{\hat{Y}}$ of Definition~\ref{def:indicator}: argmax in the multiclass case and thresholding at an arbitrary $\theta$ in the multilabel case. When misclassification costs are available, these rules can be adjusted accordingly, following the cost-sensitive principles of \citet{elkan2001foundations}. When the $t$-norm extension is applied to soft values, the entries of $\mathbold{\hat{Y}}$ in Definition~\ref{def:confusion_matrices} may represent scores in $[0,1]$ rather than hard predictions. The Calibration and Ranking Frameworks instead work explicitly with $\mathbold{\hat{P}}$; a hard prediction matrix is obtained only when a decision rule is required.

\subsection{Aggregation Operators}
\label{sec:bg_aggregation}

\begin{mydef}[Aggregation operators]
\label{def:aggregation}
Let $\mathbold{A} \in \mathbb{R}^{n \times m}$ be any real-valued matrix. Three aggregation operators are defined:
\begin{align*}
\Upsigma_{\mathrm{1}}(\mathbold{A})
  &= \sum_{i=1}^{n}\sum_{j=1}^{m} A_{i,j} \;\in\; \mathbb{R} \qquad\text{(global / micro)},\\
\bigl[\Upsigma_{\mathrm{m}}(\mathbold{A})\bigr]_j
  &= \sum_{i=1}^{n} A_{i,j} \;\in\; \mathbb{R} \qquad\text{(column-wise / per-class)},\\
\bigl[\Upsigma_{\mathrm{n}}(\mathbold{A})\bigr]_i
  &= \sum_{j=1}^{m} A_{i,j}  \;\in\; \mathbb{R} \qquad\text{(row-wise / per-example)}.
\end{align*}
In particular, $\Upsigma_{\mathrm{m}}(\mathbold{A})\in\mathbb{R}^{m}$ and $\Upsigma_{\mathrm{n}}(\mathbold{A})\in\mathbb{R}^{n}$; their indexed components are scalars.
\end{mydef}

These operators correspond to the three most common averaging
schemes in machine learning libraries (e.g., \texttt{average=
'micro'}, \texttt{'macro'}, and \texttt{'samples'} in scikit-learn):
$\Upsigma_{\mathrm{1}}$ pools all counts globally; $\Upsigma_{\mathrm{m}}$
produces a per-class vector from which an unweighted (macro) or
prevalence-weighted average is computed; $\Upsigma_{\mathrm{n}}$
produces a per-example vector whose unweighted mean is the exemplar
average.

\subsection{Coincidence Cell Operators and Confusion Matrices}
\label{sec:bg_tnorm}

In \citet{prati2026unified}, the four confusion matrices are defined by
element-wise Boolean operations (conjunction and negation) on
the binary indicator matrices. Soft labels or soft predictions are
handled by an extension in which the Boolean connectives are
replaced by their fuzzy counterparts, \emph{triangular norms},
with the Boolean definitions recovered exactly on binary matrices.
Since this paper allows soft entries throughout, we state the
general $t$-norm form directly.

\begin{mydef}[$t$-norm]
\label{def:tnorm}
A \emph{triangular norm} \citep{klement2000triangular} is a function $T: [0,1]^2 \to [0,1]$
satisfying commutativity, associativity, monotonicity, and the
boundary condition $T(a,1) = a$ for all $a \in [0,1]$. The
three standard $t$-norms are the Gödel (minimum)
$T_M(a,b) = \min(a,b)$, the product $T_P(a,b) = a \cdot b$,
and the Łukasiewicz $T_L(a,b) = \max(0, a+b-1)$.
\end{mydef}

\begin{mydef}[Confusion cell matrices]
\label{def:confusion_matrices}
Given $\mathbold{Y}, \mathbold{\hat{Y}} \in [0,1]^{n \times m}$,
a $t$-norm $T$, and the standard complement $N(a) = 1-a$, the four
\emph{confusion cell matrices} are:
\begin{align*}
\mathbold{TP}_{i,j} &= T(Y_{i,j},\; \hat{Y}_{i,j}), &
\mathbold{FP}_{i,j} &= T(N(Y_{i,j}),\; \hat{Y}_{i,j}), \\
\mathbold{FN}_{i,j} &= T(Y_{i,j},\; N(\hat{Y}_{i,j})), &
\mathbold{TN}_{i,j} &= T(N(Y_{i,j}),\; N(\hat{Y}_{i,j})).
\end{align*}
For binary $\mathbold{Y}, \mathbold{\hat{Y}} \in \{0,1\}^{n \times m}$,
all three $t$-norms reduce to the Boolean AND, so $\mathbold{TP},
\mathbold{FP}, \mathbold{FN}, \mathbold{TN} \in \{0,1\}^{n \times m}$
are binary indicator matrices whose entries equal 1 exactly where
the corresponding outcome holds for example $e_i$ and label $l_j$.
\end{mydef}

Applying the aggregation operators to these matrices yields the
familiar scalar and per-class confusion counts:
$tp = \Upsigma_{\mathrm{1}}(\mathbold{TP})$, $tp_j =
[\Upsigma_{\mathrm{m}}(\mathbold{TP})]_j$, and so on, from which
any binary measure (precision, recall, $F_1$, MCC \citep{matthews1975comparison}, \ldots) can
be computed. The framework is thus a \emph{measure generator}:
given one binary formula for $M(tp, fp, fn, tn)$ and the three aggregation
operators, a family of multiclass and multilabel
measures follows automatically.

\begin{example}[Running example]
\label{ex:running}
Throughout the paper we use $n=4$ examples and $m=3$ labels $A,B,C$, with
\[
\mathbold{Y}=\begin{pmatrix}1&1&0\\1&0&1\\0&1&0\\0&1&0\end{pmatrix},\qquad
\mathbold{\hat{P}}=\begin{pmatrix}.7&.6&.1\\.6&.1&.4\\.8&.2&.7\\.4&.3&.2\end{pmatrix}.
\]
Scores are pairwise distinct within every row and every column. The classifier scores $A$ highly for
every example, including the two for which $A$ is irrelevant; it separates $B$ from its single negative
example, but ranks $B$ low within examples. Thresholding at $\theta=0.5$ gives
$\hat{Y}_{ij}=\mathbf{1}[\hat{P}_{ij}>0.5]$ and the per-class counts
$(tp_j,fp_j,fn_j,tn_j)=(2,1,0,1)$, $(1,0,2,1)$ and $(0,1,1,2)$ for $A$, $B$ and $C$, so that
$(tp,fp,fn,tn)=(3,2,3,4)$. Micro-precision is $3/5$ and macro-precision is $(2/3+1+0)/3=5/9$: the
per-class denominators $tp_j+fp_j=(3,1,1)$ covary with the per-class precisions
(Theorem~\ref{thm:micro_macro}). Appendix~\ref{app:running} lists every intermediate value of the running
example, and the script \texttt{experiments/running\_example.py} reproduces them.
\end{example}

\subsection{Theoretical Results of the Base Framework}
\label{sec:bg_theorems}

The following results from \citet{prati2026unified} are recalled for use in this paper. For rational measures, all denominators appearing below are assumed nonzero; positive denominators are assumed whenever they are used as averaging weights.

\begin{theorem}[Micro = denominator-weighted macro for aggregation-decomposable measures]
\label{thm:micro_macro}
Let $M$ be an \emph{aggregation-decomposable} binary measure, as in Definition~7 of \citet{prati2026unified}: for the count vector $\mathbold{v}=(tp,fp,fn,tn)^\top$, suppose
\[
M(\mathbold{v})=\frac{\langle\mathbold{a},\mathbold{v}\rangle}{\langle\mathbold{b},\mathbold{v}\rangle}, \qquad \mathbold{a},\mathbold{b}\in\mathbb{R}^{4}.
\]
Let $\mathbold{v}_j$ be the per-class count vectors obtained using $\Upsigma_{\mathrm{m}}$, and assume $\mathrm{den}_j=\langle\mathbold{b},\mathbold{v}_j\rangle>0$ for every included class. Then
\[
  M^{\mathrm{micro}}=\frac{\displaystyle\sum_{j=1}^{m}\mathrm{den}_j M_j}{\displaystyle\sum_{j=1}^{m}\mathrm{den}_j}, \qquad M_j=M(\mathbold{v}_j).
\]
For fixed local denominators, this weighted mean agrees with the unweighted macro-mean for every independently variable vector $(M_1,\ldots,M_m)$ if and only if all denominators are equal. For a given observed vector of scores, the exact equality condition is $\operatorname{Cov}_{j}(\mathrm{den}_j,M_j)=0$, with covariance taken uniformly over the $m$ classes. In general, the identity does \emph{not} hold for measures whose numerators or denominators are not additive in the count vectors.
\end{theorem}

Appendix~\ref{app:bg_examples} gives a non-additive ratio (squared recall) for which the identity fails.

\begin{mydef}[Marginal consistency]
\label{def:marginal_bg}
A $t$-norm $T$, paired with the standard complement $N(x)=1-x$, is \emph{marginally consistent} if, \emph{for all} $a,b\in[0,1]$,
\[
T(a,b)+T(a,1-b)=a,\qquad T(a,b)+T(1-a,b)=b.
\]
Equivalently, both marginal identities $\mathbold{TP}_{i,j}+\mathbold{FN}_{i,j}=Y_{i,j}$ and $\mathbold{TP}_{i,j}+\mathbold{FP}_{i,j}=\hat Y_{i,j}$ must hold for \emph{every possible pair} of soft cell values, and not only for the cells of one fixed data set.
\end{mydef}

\begin{theorem}[Marginal consistency characterises the product $t$-norm]
\label{thm:partition_bg}
A $t$-norm $T$ is marginally consistent in the sense of
Definition~\ref{def:marginal_bg} if and only if $T = T_P$.
\end{theorem}

\begin{corollary}[Partition property]
\label{cor:partition_bg}
Under the product $t$-norm, the four confusion cell matrices
partition the unit measure entrywise,
\[
  \mathbold{TP}_{i,j} + \mathbold{FP}_{i,j}
  + \mathbold{FN}_{i,j} + \mathbold{TN}_{i,j} = 1
\]
for all $(i,j)$. Consequently, the aggregated four counts sum to $nm$ for an $n\times m$ matrix, and to $n$ in a single-column binary encoding. This is a \emph{consequence} of marginal consistency
(sum the two equations of Definition~\ref{def:marginal_bg} for
$Y_{i,j}$ and $N(Y_{i,j})=1-Y_{i,j}$), not an independent
characterisation of $T_P$: the weaker single-equation partition
property alone does not pin down $T_P$ among all $t$-norms without
further argument, so it should not be read as restating
Theorem~\ref{thm:partition_bg}'s uniqueness claim in a different
form.
\end{corollary}

\noindent Definition~\ref{def:marginal_bg} and
Theorem~\ref{thm:partition_bg} are Definition~8 and Theorem~3 of
\citet{prati2026unified}; Corollary~\ref{cor:partition_bg} is that
paper's Corollary~3. Together they provide a
justification for using $T_P$ when evaluating probabilistic
classifiers.

\begin{theorem}[Skew-invariance characterisation]
\label{thm:skew_base}
Define the domain $\mathcal{D}=\{(tp,fp,fn,tn)\in\mathbb{R}_{\geq0}^{4}:tp+fn>0,\ tn+fp>0\}$. A binary measure $M:\mathcal{D}\to\mathbb{R}$ is \emph{skew-invariant}, meaning
\[
  M(\lambda\,tp,\;\mu\,fp,\;\lambda\,fn,\;\mu\,tn)=M(tp,fp,fn,tn)
  \quad\text{for all }\lambda,\mu>0\text{ and all }(tp,fp,fn,tn)\in\mathcal{D},
\]
if and only if $M$ is a function of recall $\mathrm{rec}=tp/(tp+fn)$ and specificity $\mathrm{spec}=tn/(tn+fp)$ alone. The positive factors $\lambda$ and $\mu$ independently rescale the positive and negative class counts; the domain restriction excludes undefined class-conditional rates.
\end{theorem}

\begin{corollary}[Macro- and micro-averaging under skew]
\label{cor:micro_skew_base}
Let $M=g(\mathrm{rec},\mathrm{spec})$ be a skew-invariant binary measure (Theorem~\ref{thm:skew_base}) evaluated class-wise on a collection of classes with positive actual-positive and actual-negative counts. Assume that the classifier's class-conditional behaviour is held fixed under the prevalence shifts being considered.
\begin{enumerate}
  \item If the per-class rates $\mathrm{rec}_j$ and $\mathrm{spec}_j$ remain unchanged under the shifts (as they do for each binary subproblem in the multilabel setting, and under the corresponding condition on one-vs-rest error rates in multiclass problems), then $M^{\mathrm{macro}}$ is skew-invariant when the set of included classes is fixed.
  \item The pooled measure is
  \[
  M^{\mathrm{micro}}=g\!\left(\sum_j\omega_j\,\mathrm{rec}_j,\;\sum_j\tilde\omega_j\,\mathrm{spec}_j\right),\quad
  \omega_j=\frac{tp_j+fn_j}{\sum_k(tp_k+fn_k)},\quad
  \tilde\omega_j=\frac{fp_j+tn_j}{\sum_k(fp_k+tn_k)}.
  \]
  Hence equality of the per-class rates within each argument on which $g$ depends is a \emph{sufficient} condition for micro skew-invariance. It is \emph{not necessary} for a general function $g$: changes in the two pooled arguments can cancel, and the weights may be coupled by the classification setting. In general, micro skew-invariance holds exactly when the displayed composite expression is constant over the admissible prevalence shifts.
\end{enumerate}
\end{corollary}

Appendix~\ref{app:bg_examples} shows such a cancellation for micro balanced accuracy, with unequal class-specific rates and a constant pooled value.

\begin{proposition}[Exemplar-averaging is a mixture over label configurations]
\label{prop:exemplar_skew_bg}
Partition the examples into types $\tau$ by their true label
configuration, with population share $\rho_\tau$. For a
per-example measure $M_i$ (computed via $\Upsigma_{\mathrm{n}}$)
whose conditional mean given the type, $\bar{M}_\tau =
\mathbb{E}[M_i \mid \tau]$, is fixed by the classifier's
conditional behaviour,
\[
  M^{\mathrm{exemplar}} = \sum_{\tau} \rho_\tau\, \bar{M}_\tau.
\]
Hence $M^{\mathrm{exemplar}}$ is invariant under \emph{all} changes in the proportions over a fixed set of admissible types \emph{if and only if} the values $\bar{M}_\tau$ are equal across those types; in general it varies with the
population shares $\boldsymbol{\rho}$ even when every $\bar{M}_\tau$
is held fixed. For example, with $M = $ recall, a type with true
labels $\{A,B\}$ and prediction $\{A\}$ has $\bar{M}_\tau = 0.5$,
while a type with true label $\{A\}$ and the same prediction
$\{A\}$ has $\bar{M}_\tau = 1$; equal shares give $M^{\mathrm{exemplar}}
= 0.75$, while shares $0.8$ and $0.2$ give $0.60$, although neither
type's own predictions change.
\end{proposition}

\noindent Corollary~\ref{cor:micro_skew_base} is
\citeauthor{prati2026unified}'s own Corollary~5;
Proposition~\ref{prop:exemplar_skew_bg} is that paper's
Proposition~7. Binary skew-invariance alone does not
guarantee macro- or exemplar-invariance in every setting; both
require the additional condition stated above.

\noindent Among the standard \emph{binary} measures on $\mathcal{D}$, recall and specificity are
skew-invariant, while precision, $F_1$, accuracy, and MCC are
not. Balanced accuracy and G-mean, as functions of recall and
specificity, are skew-invariant in the binary case but exhibit
residual sensitivity in multiclass settings due to cross-class
contamination of the false-positive counts.

\subsection{The Micro-Collapse Result for Multiclass Discrimination}
\label{sec:bg_collapse}

A fifth result from the same source is included for completeness, since it explains a
well-known practitioner observation left otherwise unaddressed by the results above.

\begin{theorem}[Micro-collapse in multiclass]
\label{thm:micro_collapse}
For a multiclass problem (one-hot $\mathbold{Y}$ and
$\mathbold{\hat{Y}}$), micro-precision, micro-recall, and
micro-$F_1$ are all equal to overall accuracy:
\[
  P^{\mathrm{micro}} = R^{\mathrm{micro}} = F_1^{\mathrm{micro}}
  = \frac{\Upsigma_{\mathrm{1}}(\mathbold{TP})}{n}.
\]
\end{theorem}

\noindent This collapse is a consequence of the one-hot constraint:
each example contributes exactly one actual positive (to its true
class) and one predicted positive (to its predicted class); it is a true positive when they
coincide, and a false negative and a false positive otherwise, so the global
counts satisfy $ppos = tpos = n$ and $tp = $ correct predictions.
The result explains the well-known practitioner observation that
micro-averaged $F_1$ in multiclass settings equals accuracy, which holds without qualification under the indicator-matrix encoding.

\subsection{Boundary of the Base Framework}
\label{sec:bg_boundary}

\citet{prati2026unified} identify a group of measures as
lying at or beyond the boundary of the indicator-matrix framework:
these are measures that operate on the raw probability matrix
$\mathbold{\hat{P}}$ and whose cell values cannot be expressed as
$t$-norm outputs. The base framework reaches AUC as a threshold integral of the coincidence cells, but not as a measure on the
raw score matrix. The two main families are the following.

\needspace{7\baselineskip} % Keep the first scoring-rule item together (layout only).
\begin{enumerate}
  \item Proper scoring rules (Brier score, log-loss):
    $\mathrm{BS}^{\mathrm{micro}} = \frac{1}{nm}\Upsigma_{\mathrm{1}}
    \bigl((\mathbold{Y} - \mathbold{\hat{P}})^{\circ 2}\bigr)$, where
    $\circ 2$ denotes element-wise squaring. These rules assess overall probabilistic predictive quality through the magnitude of prediction errors; they are not pure calibration diagnostics, since the Brier score also reflects uncertainty and resolution.

  \item Ranking measures (nDCG, AUC) evaluate
    the ordering of the predicted scores $\hat{P}_{i,\cdot}$ or
    $\hat{P}_{\cdot,j}$ rather than their absolute values, so their
    cell values depend on the entire predicted score vector, not on
    the individual entry $\hat{P}_{i,j}$.
\end{enumerate}

\citet{prati2026unified} suggest these measures form a direction
for future extension. This paper carries out that extension by
introducing two further cell operator classes (deviation operators
for calibration and positional operators for ranking) and showing
that all three paradigms fit within a common algebraic structure
governed by the same three aggregation operators $\Upsigma_{\mathrm{1}}$,
$\Upsigma_{\mathrm{m}}$, $\Upsigma_{\mathrm{n}}$.

% ============================================================
%  Related Work
% ============================================================
% ============================================================
%  Related Work
%  Draft for: "A Generalised Confusion Matrix Framework"
%  Author: Ronaldo C. Prati
% ============================================================

\section{Related Work}
\label{sec:related}

Research on classification performance evaluation can be grouped
into six broad threads: catalogues of binary measures and their
properties; unified algebraic or decision-theoretic frameworks;
calibration and proper scoring rules; ranking evaluation;
skew-sensitivity and class imbalance; and multilabel-specific
evaluation. We position our contribution with respect to each.

\subsection{Evaluation Surveys and Catalogues}

The literature on binary classification measures is extensive.
\citet{DBLP:journals/ipm/SokolovaL09} analyse
twenty-four measures used in binary, multiclass, multilabel and
hierarchical classification and build a taxonomy of their invariance
under changes of the confusion matrix, but do so
measure-by-measure rather than through a generative mechanism.
\citet{DBLP:journals/prl/FerriHM09} conduct an
experimental comparison showing that rankings of classifiers are
highly sensitive to measure choice.
\citet{DBLP:journals/tkde/PratiBM11} survey graphical
evaluation methods and their scalar counterparts.
\citet{japkowicz2011evaluating} provide a
comprehensive textbook treatment of evaluation methodology,
covering significance testing, protocol design, and practical
considerations that complement the algebraic approach taken here.
These works provide taxonomies, but they analyse measures
individually and do not offer a mechanism for automatically
generating multiclass or multilabel extensions.

\subsection{Unified Algebraic and Decision-Theoretic Frameworks}

The closest work in spirit to ours is
\citet{DBLP:journals/jmlr/Hernandez-OralloFF12},
who propose a unified view of performance metrics as expected
classification losses under threshold variation. Their framework
operates on continuous score-based classifiers and connects
thresholds to cost curves and the ROC framework. It covers the
binary and cost-sensitive cases within a decision-theoretic
foundation, whereas \citet{prati2026unified}'s framework
operates at the level of the confusion matrix and provides an
explicit algebraic mechanism for generating multiclass and
multilabel measures from any binary formula. The two frameworks
are complementary: the approach of Hernández-Orallo et
al.\ gives a decision-theoretic view of discrimination, while the
framework of \citet{prati2026unified} provides structural results on averaging schemes.

More recently, \citet{shirdel2026framework} analyse binary
confusion-matrix metrics through an underlying \emph{reward
matrix}, showing that each metric embeds an implicit
cost--benefit trade-off that reduces to a single break-even
ratio between resources invested and value gained. Their
framework makes the decision-theoretic assumptions behind
discrimination measures explicit and supports the design of new
metrics tailored to a given cost structure; the indicator-matrix
account addresses the complementary, generative question
of extending binary formulae to multiclass and multilabel
settings, which the present paper carries beyond
discrimination.

Our present work extends \citeauthor{prati2026unified}'s framework to calibration and
ranking, which are adjacent to but outside both of the above
frameworks. \citet{flach2003geometry} provides a geometric
treatment of ROC space that complements the algebraic view: his
isometric analysis explains which classifier properties are
preserved under cost variations, while our column-dependent
operator formulation explains why AUC is structurally dual to
nDCG. \citet{provost1997analysis} use ROC analysis to compare
classifiers when class and cost distributions are imprecise,
introducing the ROC convex hull as a robust selection method;
this concern with distribution shift connects to the
skew-sensitivity analysis of Section~\ref{sec:cal_skew}.

\subsection{Calibration and Proper Scoring Rules}

The Brier score \citep{brier1950verification} is the foundational
calibration measure; \citet{murphy1973new} gave the vector partition
into uncertainty, reliability, and resolution components that is now
standard (refining earlier partitions by Sanders and by Murphy himself), which
we reformulate in matrix language (Section~\ref{sec:murphy_col}).
\citet{winkler1968good} already define a proper scoring rule as one
under which an assessor maximises the expected score if and only if the
reported probabilities equal their true judgments, and give the Brier,
logarithmic and spherical rules as examples; their attributes of
predictions are \emph{validity} and \emph{bias}, not yet the
reliability--resolution vocabulary. The theoretical foundations of
proper scoring rules are given by \citet{savage1971elicitation}
and the comprehensive treatment of
\citet{gneiting2007strictly}, whose Savage representation
theorem we embed into the cell operator framework
(Theorem~\ref{thm:savage}). \citet{kull2015novel}
give a four-term proper-score decomposition into adjustment loss, post-adjustment calibration loss, grouping loss, and irreducible loss. Their distinction between score adjustment and calibration is complementary to, but is not identical to, the three-term Murphy decomposition used here. Our contribution is to express column- and row-level Murphy identities in matrix aggregation notation using $\Upsigma_{\mathrm{m}}$ and $\Upsigma_{\mathrm{n}}$.
\citet{brocker2009reliability} shows that the
reliability--resolution decomposition extends from the Brier
score to every proper score; our matrix-language decomposition
is stated for squared error, and Br\"ocker's result delimits how
far it can be generalised within the proper subfamily.
Closest to our binning-free variant, \citet{dimitriadis2021stable} estimate the calibration map by isotonic
regression of the outcomes on the forecasts (the CORP approach, after the pool-adjacent-violators algorithm of
\citealp{ayer1955empirical}) and decompose a proper score into miscalibration, discrimination and uncertainty
as score differences. The calibration--resolution form of the decomposition for a single target is due to \citet{pohle2020murphy},
and \citet{fissler2022model} present it as a user guide for the consistent scoring functions used in regression and classification, with CORP reliability diagrams. Proposition~\ref{prop:murphy_isotonic} is the squared-error case in our notation,
with its relation to the level-set and binned terms; the estimator and its statistical properties are theirs, and
what we add is the matrix reading for the columns and rows of a multilabel matrix, together with the class-conditional pair
$(\alpha_j,\beta_j)$.

Outcome-conditioned decompositions also have a long history in forecast
verification. \citet{murphy1987general} factorise the joint distribution of
forecasts and observations as likelihood times base rate, conditioning on
the observation rather than on the forecast; \citet{yates1982external}
decomposes the Brier score into, among other terms, the covariance between
forecast and outcome and a minimum conditional variance; and
\citet{merkle2018weighted} review the Murphy and Yates decompositions and
extend them to weighted scores. The pair $(\alpha_j,\beta_j)$ belongs to the
same outcome-conditioned family but is a different summary: writing
$\hat Y$ for the forecast,
$\alpha_j=\mathrm{Var}(\hat Y\mid Y{=}1)+(1-\mathbb{E}[\hat Y\mid Y{=}1])^2$
and
$\beta_j=\mathrm{Var}(\hat Y\mid Y{=}0)+\mathbb{E}[\hat Y\mid Y{=}0]^2$, that
is, within each outcome class a dispersion term plus a squared bias, whereas
Yates' covariance, equal for a binary outcome to
$\pi_j(1-\pi_j)\bigl(\mathbb{E}[\hat Y\mid Y{=}1]-\mathbb{E}[\hat Y\mid Y{=}0]\bigr)$,
uses only first moments. What the Calibration Framework adds to this family is
the skew-invariance characterisation (Theorem~\ref{thm:cal_skew}), not a new
decomposition.

The practical importance of calibration for neural networks is
demonstrated by \citet{guo2017calibration}, who show
that modern deep classifiers are systematically overconfident and
that temperature scaling (a single-parameter post-hoc
method) substantially improves calibration. Post-hoc calibration
methods, including Platt scaling \citep{platt1999probabilistic}
for SVMs and isotonic regression for decision
trees \citep{zadrozny2002transforming}, convert raw classifier
scores into calibrated probabilities; the Calibration Framework provides
the evaluation language for assessing the quality of such
conversions at the cell level. \citet{dalpozzolo2015calibrating} document
empirically the applied counterpart of Theorem~\ref{thm:cal_skew} and
Corollary~\ref{cor:bs_skew}: undersampling the majority class to correct
for imbalance, a common practice before training, systematically distorts
predicted probabilities away from the true posterior and requires an
explicit post-hoc correction to restore calibration. This is a prior-shift effect of the same
kind as the prevalence dependence that the Calibration Framework's skew analysis expresses through
$(\alpha_j,\beta_j)$, although theirs concerns the training-set prior and not the evaluation prevalence.
\citet{wallace2012class,wallace2014improving} approach the same problem from
the evaluation side: they show that, under imbalance, class probability
estimates systematically underestimate minority-class probabilities even when
overall calibration looks good, and propose the \emph{stratified Brier
score} (the Brier score computed separately over positive and negative
examples, motivated by analogy with the sensitivity/specificity
decomposition of accuracy) together with a remedy that bags estimators
calibrated over balanced bootstrap samples. In the binary case, the two outcome-stratified Brier components coincide with $(\alpha_j,\beta_j)$ of Definition~\ref{def:cond_cal}; the pair is not being identified with one undifferentiated scalar score. The Calibration Framework makes the analogy
exact (Corollary~\ref{cor:hard_cal_skew}), embeds the pair in the matrix
formalism, and characterises it as the skew-invariant basis
(Theorem~\ref{thm:cal_skew}).
Class-weighted variants of the Brier score also appear in risk prediction:
\citet{gerds2008performance} weight cases and controls differently, in the
form
$\frac1n\sum_i\{w_1Y_i(1-\hat r_i)^2+w_0(1-Y_i)\hat r_i^2\}$ discussed by
\citet{zhu2025weighted}, who contrast it with their own decision-theoretic
weighting. In the present framework such a score is
$w_1\pi_j\alpha_j+w_0(1-\pi_j)\beta_j$, and
Remark~\ref{rem:class_weighted_brier} records when it is skew-invariant.
\citet{widmann2019calibration} propose kernel-based calibration
tests for multi-class classifiers that assess the full
distributional relationship between predicted probabilities and
true labels, going beyond the mean squared error captured by the
Brier score; we identify this distributional calibration direction
as an extension of the present framework.

\citet{degroot1983comparison} give an
early unified treatment of forecaster evaluation that connects
proper scoring rules to information theory and decision-making;
\citet{niculescu2005predicting} show
empirically how different classifier families differ in their
calibration behaviour, providing practical context for the
theoretical results of Section~\ref{sec:calibration}.

\subsection{Ranking Evaluation}

Gain-based ranking evaluation traces to \citet{jarvelin2002cumulated}, who developed cumulative and discounted cumulative gain as alternatives to set-based precision-at-$k$ in information retrieval. We use the now-standard logarithmic discount $w(r)=1/\log_2(r+1)$ in a row-positional construction, without attributing every later normalisation convention to their original formulation. The three aggregation operators
generate per-class, per-example, and global versions
(Section~\ref{sec:dcg_family}). \citet{manning2008introduction} survey the IR evaluation
literature in which nDCG plays a central role.
\citet{wang2013theoretical} analyse nDCG-type measures through the notion of
\emph{consistent distinguishability}: under the authors' asymptotic sampling and ranking assumptions, standard nDCG with a logarithmic discount converges to~1 as the number of ranked items grows even though it retains consistent distinguishability; whether a discount has this property depends on its decay rate; their results
concern the choice of $w$, whereas ours concern the aggregation
structure for a fixed $w$.

The AUC and its connection to the Wilcoxon--Mann--Whitney (WMW)
statistic is established by \citet{wilcoxon1945individual},
\citet{mann1947test}, and \citet{hanley1982meaning}. We exploit this equivalence to
define the label-dependent concordance cells whose per-class mean is AUC (Section~\ref{sec:col_dep}). An individual AUC cell also depends on which examples are negative and need not equal a fixed function of the column rank alone. Rather, the fixed-weight column-rank operator (Definition~\ref{def:col_rank}) yields an affine representation of the \emph{aggregate} AUC through mean positive rank when scores are distinct (Proposition~\ref{prop:auc_meanrank}). The column-dependent fixed-weight rank operator with a general weight $v$ is closely related to the
\emph{two-sample linear rank statistics}
$W^{\phi}_{n,m} = \sum_{i} \phi\bigl(\mathrm{Rank}(X_i)/(N{+}1)\bigr)$
of \citet{clemencon2021concentration}, who show that AUC, DCG, the
p-norm push and local AUC are all instances of this family, indexed
by the score-generating function $\phi$: their rank is computed within
the pooled positive-and-negative sample, exactly as
$\mathrm{rank}_j(i)$ ranks example $i$ within the whole column
$\hat{\mathbold{Y}}_{\cdot,j}$ here, and the linear rank choice $\phi(u)=u$ gives the WMW/AUC relation at the aggregate level, after the appropriate count-dependent affine normalisation (for distinct scores). Their contribution and ours are complementary: they establish concentration inequalities and
generalisation bounds for empirical maximisers of this family within a
single bipartite-ranking population, with no matrix structure or
multiclass/multilabel aggregation; the present paper instead embeds the
same rank-statistic idea into the indicator-matrix architecture, where
it becomes only one of two axes: the row-dependent nDCG family (Section~\ref{sec:disc_op}) ranks labels \emph{within} one example rather than ranking examples in a pooled positive--negative population, an axis not addressed by their bipartite two-sample formulation. The row--column duality of Theorem~\ref{thm:duality} formalises this comparison in the present matrix framework. \citet{hand2001simple} extend AUC to
multi-class problems via pairwise class comparisons; \citet{fawcett2006introduction}
provides a comprehensive practical introduction.

\citet{cortes2004auc} prove that AUC optimisation
and error rate minimisation can favour different classifiers,
which is consistent with the conclusion of Section~\ref{sec:auc_props} that
AUC is not a strictly proper scoring rule and measures a different
property from calibration (their result concerns optimisation, not properness). \citet{hand2009measuring} criticises
AUC on decision-theoretic grounds (interpreted as an expected
minimum loss, it applies cost distributions that depend on the
classifier being evaluated) and proposes the H-measure as a
coherent alternative. That critique is orthogonal to ours: the
column-dependent view characterises the algebraic structure and
skew-invariance of AUC independently of its decision-theoretic
interpretation. \citet{furnkranz2008multilabel} use label ranking (ordering the labels of each
example, with a calibrated threshold) for multilabel classification, the row-axis ranking setting
related to the nDCG family and to the per-example measures of
Definition~\ref{def:auc_family}. \citet{dembczynski2012label} characterise the Bayes-optimal
multilabel classifier under different loss functions, including
ranking losses; their framework and ours are complementary in that
they identify which classifier to train for a given loss, while we
analyse the structure of the loss itself.

\subsection{Skew-Sensitivity and Class Imbalance}

The foundational cost-sensitive view of class imbalance is given
by \citet{elkan2001foundations},
who shows that cost-sensitive classification reduces to reweighted
probability estimation. \citet{luque2019impact} quantify, metric
by metric, how class imbalance distorts confusion-matrix
measures, and \citet{brzezinski2018visual} analyse the same
behaviour through gradient and isometric visualisations in the
spirit of \citet{flach2003geometry}. These studies document
empirically the phenomenon that the algebraic view explains
structurally. \citet{prati2026unified}
establish the algebraic characterisation of skew-invariance
(Theorem~\ref{thm:skew_base}), which we extend to calibration
(Theorem~\ref{thm:cal_skew}) and ranking
(Propositions~\ref{prop:ndcg_class_skew} and~\ref{prop:auc_skew}).
A unifying observation of this paper is that the skew-invariant
per-class bases across all three frameworks, namely recall/specificity
(Discrimination Framework), $(\alpha_j, \beta_j)$ (Calibration Framework), and
$\mathrm{nDCG}_j^{\mathrm{class}}$/$\mathrm{AUC}_j$ (Ranking Framework), share a class-conditional mean template over the appropriate outcome: positive examples for recall, $\alpha_j$, per-class nDCG and AUC, and negative examples for specificity and $\beta_j$. Section~\ref{sec:synthesis} states the hypotheses in full.

Skew-invariance itself, as \citet{prati2026unified} note, is related to but
distinct from the \emph{constant baseline} property of
\citet{gosgens2021good}, also discussed by \citet{hand2024selecting}: constant
baseline requires a measure to score a random assignment consistently
regardless of how many objects are assigned to each class, constraining
behaviour at one reference point, whereas skew-invariance constrains behaviour
over the entire space of classifier configurations. This distinction
transfers unchanged to the Calibration and Ranking Frameworks: a constant-baseline-style
axiom for the Brier score or for nDCG/AUC would fix their value only under a
random or degenerate predictor, which is a different requirement from the
class-conditional invariance that $(\alpha_j,\beta_j)$ and
$\mathrm{nDCG}_j^{\mathrm{class}}$/$\mathrm{AUC}_j$ satisfy throughout this
paper (neither property implies the other). A closely related axiomatic tradition exists in the quantification
literature, whose object is estimating aggregate class prevalence
robustly under prior shift: \citet{sebastiani2020evaluation} develops an
axiomatic account of what makes an evaluation measure appropriate for
quantification, in the same spirit as \citet{gosgens2021good} for
classification but specific to the prior-shift setting that motivates
skew-invariance here. Quantification and skew-invariant classifier
evaluation address the same underlying phenomenon (distributional
change in class or label prevalence) from complementary directions:
quantification estimates the prevalence itself, while the skew-invariant
measures of this paper are constructed to be unaffected by it.

\subsection{Multilabel and Multiclass Evaluation}

The indicator-matrix representation of multilabel problems goes
back to \citet{DBLP:journals/jdwm/TsoumakasK07}
and \citet{zhang2014review}, who catalogue example-based
and label-based measures including Hamming loss, exemplar
precision, and exemplar recall. \citet{godbole2004discriminative} extend precision, recall,
and $F_1$ to multilabel using set-intersection semantics, which the
framework of \citet{prati2026unified} subsumes as a special case of the product $t$-norm
applied to binary matrices. Our extension preserves the indicator-matrix setting across multilabel, multiclass, and binary data, subject to each measure's stated domain and tie conventions. In particular, normalised DCG needs a positive ideal denominator, and per-class AUC needs both positive and negative examples (or an explicit degenerate-case convention); the formulas should not be read as universally defined without such conditions. \citet{wu2017unified} also unify multilabel
performance measures, relating eleven of them through
label-wise and instance-wise margins and proving that maximising these
margins optimises the corresponding measures (they assume tie-free
predictions, the same regime in which our positional results hold). Their unification targets the
learning side (which margins a training algorithm should
optimise), whereas the indicator-matrix account targets the
algebraic structure of the measures themselves; the two views
are complementary. \citet{koyejo2014consistent} characterise which
multilabel losses admit consistent optimisation, connecting our
structural analysis to the learning-theoretic perspective.

The ordinal extension via quadratic weighted kappa
builds on \citet{cohen1960coefficient} and the ordinal
classification framework of \citet{frank2001simple};
we treat QWK as a hybrid of the Calibration Framework (deviation numerator)
and auxiliary column-marginal quantities, in direct parallel to
\citeauthor{prati2026unified}'s treatment of Cohen's $\kappa$.

% ============================================================
%  Section 3 — Cell Operator Taxonomy
% ============================================================
% ============================================================
%  Section 3 — A Taxonomy of Cell Operators
%  Draft for: "A Generalised Confusion Matrix Framework"
%  Author: Ronaldo C. Prati
%
%  Dependencies: Section 2 (background review of Framework 1),
%  in particular Definition 2 (TP/FP/FN/TN matrices via t-norms),
%  the three aggregation operators Sigma_1, Sigma_m, Sigma_n,
%  and Theorem 1 (micro = weighted macro).
% ============================================================

\section{A Taxonomy of Cell Operators}
\label{sec:taxonomy}

The indicator-matrix framework of \citet{prati2026unified} rests on a
specific choice of how to compare actual label $Y_{i,j}$ and
predicted label $\hat{Y}_{i,j}$ at each matrix cell. In
the Discrimination Framework, this comparison is a conjunction
(the Boolean AND on binary matrices, a $t$-norm in the soft case),
measuring the degree to which both values coincide. The resulting cells $\mathbold{TP}$,
$\mathbold{FP}$, $\mathbold{FN}$, $\mathbold{TN}$ measure
discrimination, that is, whether the classifier identifies the correct
label. This section shows that replacing this coincidence operator
with a different cell operator yields two further evaluation paradigms,
each with its own interpretation and its own family of measures.

\begin{mydef}[Cell operator]
\label{def:cell_op}
A \emph{cell operator} is a function that assigns to each cell
$(i,j)$ of the label matrices a real value comparing the actual
label $Y_{i,j} \in [0,1]$ with the prediction. A cell operator is
\emph{pointwise} if this value depends only on the single pair of
entries at $(i,j)$, i.e.\ $D: [0,1] \times [0,1] \to \mathbb{R}$
with $[\mathbold{D}]_{i,j} = D(Y_{i,j}, \hat{Y}_{i,j})$; it is
\emph{row-dependent} if the value at $(i,j)$ may depend on the
full predicted row $\hat{\mathbold{Y}}_{i,\cdot}$, i.e.\
$D_{\mathrm{row}}: [0,1] \times [0,1]^{m} \to \mathbb{R}$ with
$[\mathbold{D}]_{i,j} = D_{\mathrm{row}}(Y_{i,j}, \hat{\mathbold{Y}}_{i,\cdot})$.
A cell operator is \emph{column-dependent} if it may also use the
actual labels in the full column. In general it has the form
$\Phi_{\mathrm{col}}:\{1,\ldots,n\}\times[0,1]^{n}\times[0,1]^{n}\to\mathbb{R}$,
with $[\mathbold{D}]_{i,j}=\Phi_{\mathrm{col}}(i,\mathbold{Y}_{\cdot,j},\hat{\mathbold{Y}}_{\cdot,j})$.
The restricted form $D_{\mathrm{col}}(Y_{i,j},\hat{\mathbold{Y}}_{\cdot,j})$
is included, but is insufficient for concordance-based AUC cells, which
also depend on which other examples are negative. In every
case the \emph{cell matrix} induced by $D$ is
\[
  \mathbold{D}(\mathbold{Y}, \mathbold{\hat{Y}}) \;\in\; \mathbb{R}^{n \times m}.
\]
Applying the aggregation operators $\Upsigma_{\mathrm{1}}$,
$\Upsigma_{\mathrm{m}}$, $\Upsigma_{\mathrm{n}}$ to
$\mathbold{D}(\mathbold{Y}, \mathbold{\hat{Y}})$ yields scalar,
per-class, and per-example summaries of the evaluation
criterion encoded by $D$. Coincidence and deviation operators
(Sections~\ref{sec:tax_coincidence}--\ref{sec:tax_deviation}) are
pointwise; positional operators
(Section~\ref{sec:tax_discount}) are row- or column-dependent.
\end{mydef}

\begin{figure}[!t]
\centering
\includegraphics[width=\textwidth]{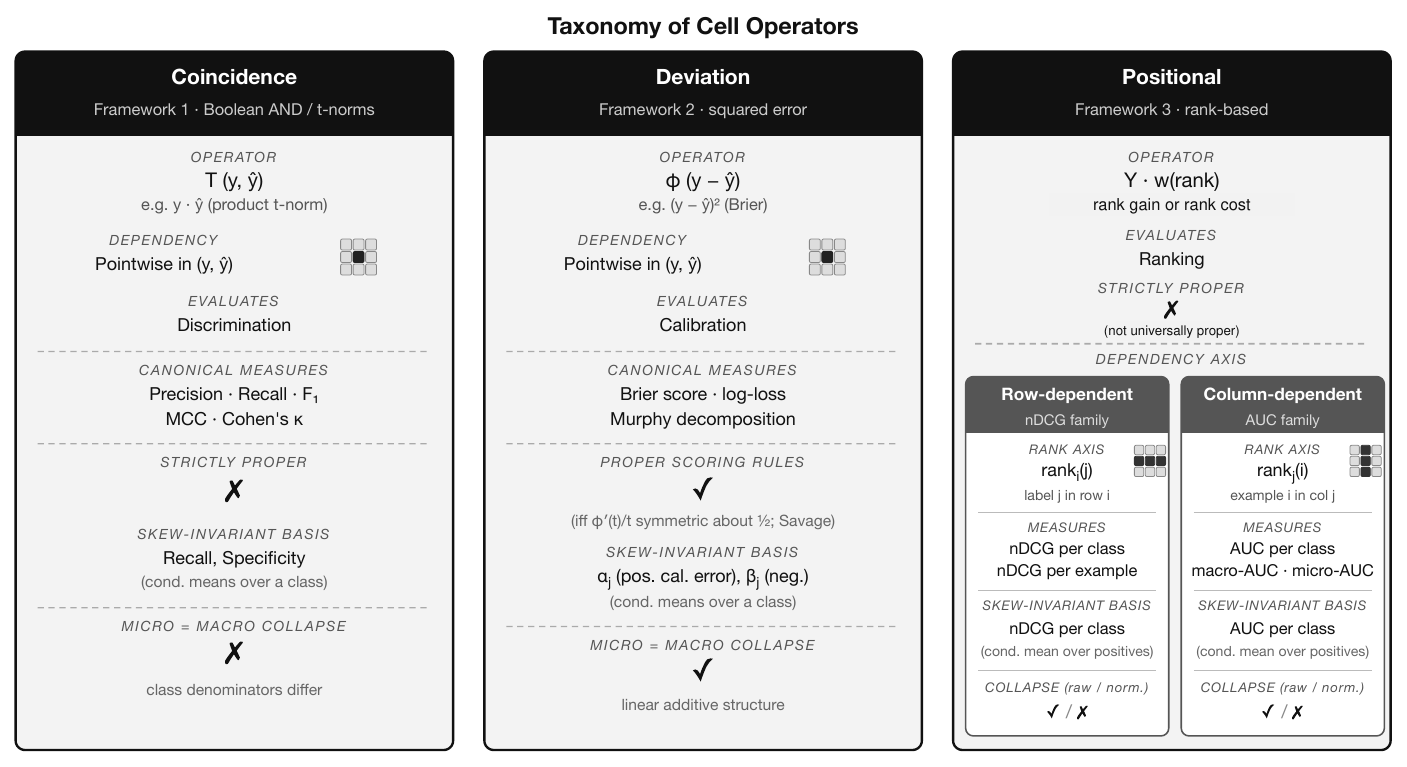}
\caption{Taxonomy of the three cell operator classes.
Each panel shows the operator form, its dependency structure
(illustrated by the highlighted cell in the $3\times3$ matrix icon),
canonical evaluation measures, proper scoring rule status,
skew-invariant basis, and micro\,=\,macro collapse behaviour.
Framework~3 (Positional) splits into a row-dependent subfamily
(nDCG, rank within a row) and a column-dependent subfamily
(AUC, rank within a column), unified by the row--column duality
theorem (Section~\ref{sec:ranking}).
$T$ denotes a triangular norm; $\phi$ is a convex symmetric function
with $\phi(0)=0$; $w(\cdot)$ denotes a non-increasing discount weight for row nDCG, whereas the column AUC mean-rank construction admits the increasing rank cost $v(r)=r$.
The cross under ``strictly proper'' in the Positional panel covers both subfamilies:
raw positional scores are never strictly proper. For non-increasing discounts and pairwise distinct true marginals, truthful reporting maximises their expected raw score among competing reports with distinct entries (Proposition~\ref{prop:disc_improper}). Under the min-rank convention, allowing tied reports can invalidate weak properness; constant discounts are an exception, since every report then has the same expected raw score.
Normalised nDCG and AUC need not be proper (Remark~\ref{rem:ndcg_not_proper}, Proposition~\ref{prop:auc_improper}).}
\label{fig:taxonomy}

\end{figure}

We identify three qualitatively distinct classes of cell
operator, each giving rise to a different evaluation framework.
Each class generalises one elementary comparison between an actual and a
predicted value: \emph{conjunction} (do they coincide?),
\emph{deviation} (how far apart are they?), or \emph{position}
(where does the prediction rank the label?). Each
elementary comparison anchors one evaluation paradigm, namely
discrimination, calibration, and ranking, respectively.
Figure~\ref{fig:taxonomy} summarises the three classes; the
remainder of the paper develops the Calibration and Ranking
Frameworks in detail.

\subsection{Coincidence Operators (Discrimination Framework)}
\label{sec:tax_coincidence}

A coincidence operator measures the degree to which $Y_{i,j}$ and
$\hat{Y}_{i,j}$ are simultaneously large. In the binary case it is
the Boolean AND of \citet{prati2026unified}; for soft values the
canonical family is the $t$-norms adopted there for soft labels:
$T_P(y,\hat{y}) = y\hat{y}$, $T_M(y,\hat{y}) = \min(y,\hat{y})$,
and $T_L(y,\hat{y}) = \max(0, y+\hat{y}-1)$. For binary
$Y,\hat{Y}$, all three recover the Boolean AND, giving the
crisp confusion-matrix cells $\mathbold{TP}$, $\mathbold{FP}$,
$\mathbold{FN}$, $\mathbold{TN}$ of the Discrimination Framework. For soft
$Y,\hat{Y} \in [0,1]$, the product $t$-norm is singled out by
marginal consistency (the characterisation theorem of \citealt{prati2026unified}, restated as
Theorem~\ref{thm:partition_bg}): it is the unique $t$-norm for which
$\mathbold{TP}_{i,j} + \mathbold{FN}_{i,j} = Y_{i,j}$ and
$\mathbold{TP}_{i,j} + \mathbold{FP}_{i,j} = \hat{Y}_{i,j}$, and it then satisfies the
partition property
$\mathbold{TP}_{i,j} + \mathbold{FP}_{i,j} + \mathbold{FN}_{i,j}
+ \mathbold{TN}_{i,j} = 1$ (Corollary~\ref{cor:partition_bg}).

Coincidence operators are designed to reward correct hard
decisions, and the measures they generate (precision, recall,
$F_1$, MCC, \ldots) are appropriate when only the final class
assignment matters. They do not reward confidence once the
predictions are thresholded: a classifier that predicts
$\hat{Y}_{i,j} = 0.51$ and one that predicts $\hat{Y}_{i,j} = 0.99$
receive identical credit when $Y_{i,j} = 1$ and the threshold is $0.5$. (Applied to the
soft predictions themselves, every $t$-norm gives $T(1,\hat{y}) = \hat{y}$, so the
true-positive credit does grow with $\hat{y}$, but it measures coincidence, not calibration.)

\subsection{Deviation Operators (Calibration Framework)}
\label{sec:tax_deviation}

A deviation operator measures the magnitude of the gap between
$Y_{i,j}$ and $\hat{Y}_{i,j}$. The natural requirement is that
$D(y,\hat{y}) = 0$ when $y = \hat{y}$ and $D(y,\hat{y}) > 0$
otherwise, with larger deviations incurring larger penalties.
The prototypical family is parametrised by a convex, symmetric
function $\phi: \mathbb{R} \to \mathbb{R}_{\geq 0}$ with
$\phi(0) = 0$ and $\phi(t)>0$ for $t\neq0$:
\begin{equation}
  D_{\phi}(y, \hat{y}) \;=\; \phi(y - \hat{y}).
  \label{eq:dev_op}
\end{equation}
The two most important special cases are:
\begin{itemize}
  \item $\phi(t) = t^{2}$: the squared-error operator
    $D(y,\hat{y}) = (y-\hat{y})^{2}$, inducing the Brier score
    family (Section~\ref{sec:calibration}).
  \item $\phi(t) = |t|$: the absolute-error operator
    $D(y,\hat{y}) = |y - \hat{y}|$, inducing a mean absolute
    calibration error family.
\end{itemize}
Unlike thresholded coincidence operators, deviation operators reward
confident correct predictions: a prediction of $0.99$ for a true
positive is penalised far less than a prediction of $0.51$. This is accuracy together with
sharpness at the cell level; calibration proper, whether predicted probabilities match frequencies,
is measured by the reliability term of Section~\ref{sec:calibration}. The
Savage representation theorem (Section~\ref{sec:proper_scoring})
characterises which members of this family yield proper scoring
rules: a symmetric deviation operator is proper exactly when its
weight $\phi'(t)/t$ is symmetric about $\tfrac12$
(Corollary~\ref{cor:savage_cases}(c)). Squared error is the simplest
such operator, and the only power-law one; it is not the only one. Log-loss is an extended-valued variant of the deviation construction: for binary outcomes, it corresponds to $\phi(t)=-\log(1-|t|)$, with $\phi(\pm1)=+\infty$, and therefore lies outside the finite-valued family defined above.

\subsection{Positional Operators (Ranking Framework)}
\label{sec:tax_discount}

A positional operator weights the actual label $Y_{i,j}$ by a
function of the \emph{rank} that the prediction assigns to the
entry $(i,j)$ within a chosen slice of $\mathbold{\hat{Y}}$. The
slice may be a \emph{row} or a \emph{column}, and this choice is
itself a degree of freedom, unique to this operator class,
which gives rise to two dual subfamilies.

The row-dependent subfamily (nDCG) ranks label $j$ among the other
labels of example $i$'s prediction row:
\[
  D_w\bigl(Y_{i,j}, \hat{\mathbold{Y}}_{i,\cdot}\bigr)
  \;=\; Y_{i,j} \cdot w\!\bigl(\mathrm{rank}_i(j)\bigr),
  \qquad
  \mathrm{rank}_i(j) = \bigl|\{j' : \hat{Y}_{i,j'} > \hat{Y}_{i,j}\}\bigr| + 1,
\]
with $w: \mathbb{N} \to \mathbb{R}_{\geq 0}$ non-increasing; the
canonical nDCG weight is $w(r) = 1/\log_2(r+1)$.

The column-dependent subfamily (AUC) ranks example $i$ among the other
examples of class $j$'s prediction column:
\[
  D_v\bigl(Y_{i,j}, \hat{\mathbold{Y}}_{\cdot,j}\bigr)
  \;=\; Y_{i,j} \cdot v\!\bigl(\mathrm{rank}_j(i)\bigr),
  \qquad
  \mathrm{rank}_j(i) = \bigl|\{i' : \hat{Y}_{i',j} > \hat{Y}_{i,j}\}\bigr| + 1,
\]
with a fixed nonnegative rank function $v:\mathbb{N}\to\mathbb{R}_{\geq0}$. Unlike a direct gain weight, $v$ need not be non-increasing: the mean-rank construction of AUC takes the increasing rank cost $v(r)=r$ (Proposition~\ref{prop:auc_meanrank}).

The fixed-rank template applies directly to the row DCG gain cells $Y_{ij}w(\mathrm{rank}_i(j))$ and to the column \emph{mean-rank cost} cells $Y_{ij}v(\mathrm{rank}_j(i))$. These use a fixed, exogenous rank function, but they must not be identified cell by cell with AUC's concordance cells $A_{ij}=Y_{ij}f(i,j)$: the latter also depend on which other examples are negative (Equation~\eqref{eq:auc_cell}). The common geometric distinction is the \emph{axis} of ranking (row $\hat{\mathbold{Y}}_{i,\cdot}$ versus column $\hat{\mathbold{Y}}_{\cdot,j}$). The row DCG aggregate is normalised by its ideal DCG to give nDCG. For columns with both positive and negative examples and distinct scores, the mean positive column rank is instead related to per-class AUC by an affine Wilcoxon--Mann--Whitney transformation \emph{of the aggregate} (Section~\ref{sec:col_dep}, Proposition~\ref{prop:auc_meanrank}), not by an identity of the underlying cells.
The normalisations act differently: nDCG divides by an ideal DCG, whereas AUC applies a decreasing affine transformation of mean rank whose coefficients depend on class counts (Proposition~\ref{prop:auc_meanrank}). This produces skew-invariant \emph{per-class} nDCG and AUC, but normalisation alone does not guarantee skew-invariance of micro- or exemplar-level summaries (Corollary~\ref{cor:rank_skew_summary}; Proposition~\ref{prop:auc_skew}).

 The AUC concordance fraction $f(i,j)$ is \emph{not} a fixed function of an individual example's column rank: it counts outranked \emph{negative} examples and therefore depends on the other labels in column $j$. Only the resulting per-class mean admits the affine rank representation under the stated assumptions. The duality of Section~\ref{sec:ranking} (Theorem~\ref{thm:duality}) records the shared endpoint: both $\mathrm{nDCG}_j^{\mathrm{class}}$ and $\mathrm{AUC}_j$ are conditional means over positive examples.

Positional operators differ from the first two
classes: the cell value depends on the \emph{full predicted slice}, that is, the entire row (row-dependent) or column (column-dependent), not only on the single entry $\hat{Y}_{i,j}$. For concordance-based AUC cells, the negative-label membership within the column is also needed. The cell matrix is therefore not pointwise in the sense of Definition~\ref{def:cell_op}. The aggregation operators
$\Upsigma_{\mathrm{m}}$ and $\Upsigma_{\mathrm{n}}$ still apply to
the resulting cell matrix, but the interpretation changes:
correcting a single prediction $\hat{Y}_{i,j}$ may alter the rank
of every other entry in its slice, changing multiple cell values
at once. The full development of both subfamilies (the DCG and
AUC measure families, their collapse behaviour, skew-invariance,
and the duality theorem) is deferred to
Section~\ref{sec:ranking}.

\begin{example}[Running example: four cell matrices]
\label{ex:running_cells}
The pair $(\mathbold{Y},\mathbold{\hat{P}})$ of Example~\ref{ex:running} produces one cell matrix per
operator class. With $w_2=1/\log_2 3\approx0.631$ and $w_3=1/2$, the product $t$-norm, the squared
deviation, the row-positional gain $Y_{ij}w(\mathrm{rank}_i(j))$ and the column concordance fraction give
\[
\begin{pmatrix}.7&.6&0\\.6&0&.4\\0&0&0\\0&0&0\end{pmatrix},\quad
\begin{pmatrix}.09&.16&.01\\.16&.01&.36\\.64&.64&.49\\.16&.49&.04\end{pmatrix},\quad
\begin{pmatrix}1&w_2&0\\1&0&w_2\\0&w_3&0\\0&w_2&0\end{pmatrix},\quad
\begin{pmatrix}1/2&1&0\\1/2&0&2/3\\0&1&0\\0&1&0\end{pmatrix}.
\]
The first two depend on $(Y_{ij},\hat{P}_{ij})$ alone. The positional gain of cell $(1,B)$ needs the
whole first row ($B$ is ranked second there), and the concordance cell $(2,C)$ needs the whole column $C$
and its negatives: the score $0.4$ exceeds $0.1$ and $0.2$ but not $0.7$.
\end{example}

\subsection{Framework Bridge and Axis Duality}
\label{sec:tax_bridge}

A complete comparison of the three operator classes is deferred to
Table~\ref{tab:op_properties} in Section~\ref{sec:synthesis}, once all
frameworks have been fully developed. Two structural results are noted
here, as they motivate the design of Sections~\ref{sec:calibration}
and~\ref{sec:ranking}.

The first is a bridge between the Discrimination and Calibration
Frameworks: although the two frameworks use different cell operators
in general, they coincide exactly for binary hard predictions.

\begin{theorem}[Framework bridge]
\label{thm:bridge_s3}
Let $\mathbold{Y}, \mathbold{\hat{Y}} \in \{0,1\}^{n \times m}$
be binary indicator matrices. Under the product $t$-norm
$T_P$ for the Discrimination Framework and the squared-error operator for
the Calibration Framework:
\begin{enumerate}
  \item[\textup{(a)}] $(Y_{i,j} - \hat{Y}_{i,j})^{2}
    = \mathbold{FP}_{i,j} + \mathbold{FN}_{i,j}$
    for every $(i,j)$.
  \item[\textup{(b)}] $\mathrm{BS}^{\mathrm{micro}}
    = \mathrm{HammingLoss}(\mathbold{Y}, \mathbold{\hat{Y}})$.
  \item[\textup{(c)}] $\mathrm{BS}^{\mathrm{micro}}
    + \mathrm{LabelAccuracy}(\mathbold{Y}, \mathbold{\hat{Y}}) = 1$.
\end{enumerate}
\end{theorem}

\begin{proof}
See Section~\ref{sec:calibration}, Theorem~\ref{thm:bridge}.
\end{proof}

\noindent Theorem~\ref{thm:bridge_s3} establishes that the squared-error
cell operator is the natural soft extension of the binary error
operator $Y \oplus \hat{Y}$ used for Hamming loss. The two
frameworks diverge precisely in the soft setting: when
$\hat{Y}_{i,j} \in (0,1)$, deviation operators penalise
wrong or hesitant confidence, while thresholded coincidence operators
reward only the sign of the prediction relative to a threshold.
This divergence is the reason both frameworks are needed: a
classifier that is highly discriminative (high $F_1$) may still
be poorly calibrated (high Brier score), and vice versa.

\begin{remark}[Row/column symmetry in the Discrimination and Calibration Frameworks]
\label{rem:row_col_symmetry}
The row/column duality of the Ranking Framework has no direct analogue for coincidence and deviation
operators. These act pointwise, so $D(Y_{i,j},\hat{Y}_{i,j})$ is unchanged up to reindexing when rows and
columns are exchanged; choosing an axis changes only the aggregation, not the cell operator, and a
separable loss such as squared error aggregated along a row is the sum of its per-cell losses. Ranking is
relational: $\mathrm{rank}_i(j)$ compares scores within row~$i$ (how an example orders its labels, nDCG
family), whereas $\mathrm{rank}_j(i)$ compares them within column~$j$ (how a class separates positives from
negatives, AUC family). Adding a rank-based selection rule to coincidence gives the positional operator
$Y_{i,j}\,\mathbf{1}[\mathrm{rank}_i(j)\le k]$, whose row sum, for $1\le k\le m$ and an ordering that
selects exactly $k$ positions, divided by $k$ is Precision@$k$
(Remark~\ref{rem:hard_ranking_general_k}). This describes the constructions studied here, not a
classification of all cell operators: a softmax cross-entropy, for instance, depends on the whole
prediction row without using ranks and lies outside the present taxonomy.
\end{remark}

% ============================================================
%  Section 4 — Calibration Framework
% ============================================================
% ============================================================
%  Calibration Framework: Calibration via Squared Deviation
%  Draft for: "A Generalised Confusion Matrix Framework"
%  Author: Ronaldo C. Prati
%
%  Dependencies: assumes the notation and theorems of the base
%  paper (Discrimination Framework), in particular Definition 2
%  (TP/FP/FN/TN matrices), the three aggregation operators, and
%  Theorem 1 (micro = weighted macro).
% ============================================================

\section{The Calibration Framework: Deviation Operators}
\label{sec:calibration}

The Discrimination Framework evaluates whether a classifier
correctly identifies which class or label applies to each example.
Its cell operators (Boolean conjunction, $t$-norms) ask whether
the actual and predicted labels \emph{coincide}. This section
introduces a complementary framework that evaluates
\emph{calibration}: whether the predicted soft probabilities
faithfully represent the true label frequencies. Calibration
matters whenever the raw probability $\hat{P}_{i,j}$ is used
downstream (for ranking, risk assessment, or decision-making
under uncertainty), and not only the hard decision derived from it.
Miscalibration is widespread in modern classifiers, including
deep neural networks~\citep{guo2017calibration}, and is typically
corrected by post-hoc methods such as Platt
scaling~\citep{platt1999probabilistic} or isotonic
regression~\citep{zadrozny2002transforming}; these methods are easiest
to analyse when calibration is formalised as a
matrix-level operation, as we do here.

\subsection{The Calibration Cell Operator}
\label{sec:cal_cell}

Replace the conjunction-based cell operators of the Discrimination Framework with
a single squared-deviation operator.

\begin{mydef}[Calibration cell operator]
\label{def:cal_cell}
For $Y_{i,j},\,\hat{Y}_{i,j} \in [0,1]$, define the
\emph{calibration entry}
\begin{equation}
  \mathrm{Cal}_{i,j} \;=\; \bigl(Y_{i,j} - \hat{Y}_{i,j}\bigr)^{2}.
  \label{eq:cal_cell}
\end{equation}
The \emph{calibration matrix} is
$\mathbold{\mathrm{Cal}} = (\mathbold{Y} - \mathbold{\hat{Y}})^{\circ 2}$,
where $\circ 2$ denotes element-wise squaring (Hadamard square).
\end{mydef}

Applying the three aggregation operators $\Upsigma_{\mathrm{1}}$,
$\Upsigma_{\mathrm{m}}$, $\Upsigma_{\mathrm{n}}$ to
$\mathbold{\mathrm{Cal}}$ yields a family of Brier-score variants.

\begin{mydef}[Brier score family]
\label{def:brier_family}
\begin{align}
  \mathrm{BS}^{\mathrm{micro}}
    &= \frac{1}{nm}\,\Upsigma_{\mathrm{1}}(\mathbold{\mathrm{Cal}})
     = \frac{1}{nm}\sum_{i,j}(Y_{i,j}-\hat{Y}_{i,j})^{2}
  \label{eq:bs_micro}\\[4pt]
  \mathrm{BS}_{j}^{\mathrm{class}}
    &= \frac{1}{n}\,\bigl[\Upsigma_{\mathrm{m}}(\mathbold{\mathrm{Cal}})\bigr]_{j}
     = \frac{1}{n}\sum_{i}(Y_{i,j}-\hat{Y}_{i,j})^{2},
  \quad j \in [1,m]
  \label{eq:bs_class}\\[4pt]
  \mathrm{BS}_{i}^{\mathrm{example}}
    &= \frac{1}{m}\,\bigl[\Upsigma_{\mathrm{n}}(\mathbold{\mathrm{Cal}})\bigr]_{i}
     = \frac{1}{m}\sum_{j}(Y_{i,j}-\hat{Y}_{i,j})^{2},
  \quad i \in [1,n]
  \label{eq:bs_example}
\end{align}
The \emph{macro-Brier} score is the unweighted average of the
per-class values,
$\mathrm{BS}^{\mathrm{macro}} = \frac{1}{m}\sum_{j}\mathrm{BS}_{j}^{\mathrm{class}}$,
and the \emph{exemplar-Brier} score is the unweighted average of
the per-example values,
$\mathrm{BS}^{\mathrm{exemplar}} = \frac{1}{n}\sum_{i}\mathrm{BS}_{i}^{\mathrm{example}}$.
\end{mydef}

\noindent Equation~(\ref{eq:bs_micro}) recovers, up to the normalisation by $m$, the standard
multiclass Brier score~\citep{brier1950verification} used in the
multilabel literature when $\mathbold{Y}$ is binary and
$\mathbold{\hat{Y}}=\mathbold{\hat{P}}$ is a soft probability
matrix. Equation~(\ref{eq:bs_class}) gives a
per-class Brier score useful for diagnosing which labels are
poorly calibrated. Equation~(\ref{eq:bs_example}) gives a
per-example Brier score measuring calibration quality for
individual examples' label vectors.

% ============================================================
\subsection{Connection to the Discrimination Framework: The Framework Bridge}
\label{sec:bridge}
% ============================================================

The following theorem establishes an exact algebraic connection
between the Calibration Framework and the Hamming-loss / label-accuracy
measures of the Discrimination Framework.

\begin{theorem}[Framework bridge]
\label{thm:bridge}
Let $\mathbold{Y}, \mathbold{\hat{Y}} \in \{0,1\}^{n \times m}$
be binary indicator matrices. Then:
\begin{enumerate}
  \item[\textup{(a)}]
    $\mathrm{Cal}_{i,j} = \mathbold{FP}_{i,j} + \mathbold{FN}_{i,j}$
    for every $(i,j)$, i.e.\
    $\mathbold{\mathrm{Cal}} = \mathbold{FP} + \mathbold{FN}$
    (entry-wise).
  \item[\textup{(b)}]
    $\mathrm{BS}^{\mathrm{micro}} = \mathrm{HammingLoss}(\mathbold{Y},\mathbold{\hat{Y}})$.
  \item[\textup{(c)}]
    $\mathrm{BS}^{\mathrm{micro}} + \mathrm{LabelAccuracy}(\mathbold{Y},\mathbold{\hat{Y}}) = 1$.
\end{enumerate}
\end{theorem}

\begin{proof}
\textit{(a)} For binary $a, b \in \{0,1\}$:
$(a - b)^{2} = 1$ iff $a \neq b$, and $(a-b)^{2} = 0$ iff
$a = b$. Thus $(a-b)^{2} = a \oplus b$ (XOR). In the indicator
matrix language, $Y_{i,j} \oplus \hat{Y}_{i,j} = 1$ iff exactly
one of the two is~1, which corresponds to either
$\mathbold{FP}_{i,j} = 1$ (predicted positive, actually negative)
or $\mathbold{FN}_{i,j} = 1$ (predicted negative, actually
positive); both cases are mutually exclusive by definition. Hence
$\mathrm{Cal}_{i,j} = \mathbold{FP}_{i,j} + \mathbold{FN}_{i,j}$.

\textit{(b)} By (a) and the definition of Hamming loss of \citet{prati2026unified},
\[
  \mathrm{BS}^{\mathrm{micro}}
  = \frac{1}{nm}\Upsigma_{\mathrm{1}}(\mathbold{\mathrm{Cal}})
  = \frac{1}{nm}\Upsigma_{\mathrm{1}}(\mathbold{FP}+\mathbold{FN})
  = \frac{1}{nm}\Upsigma_{\mathrm{1}}(\mathbold{Y}\oplus\mathbold{\hat{Y}})
  = \mathrm{HammingLoss}.
\]

\textit{(c)} For binary indicator matrices,
$\mathrm{LabelAccuracy} = (\Upsigma_{\mathrm{1}}(\mathbold{TP}) +
\Upsigma_{\mathrm{1}}(\mathbold{TN}))/(nm)$. Since the four
matrices $\mathbold{TP}$, $\mathbold{FP}$, $\mathbold{FN}$,
$\mathbold{TN}$ partition every entry (for binary matrices this is
immediate; in general it is Corollary~\ref{cor:partition_bg}),
$\Upsigma_{\mathrm{1}}(\mathbold{TP}) +
\Upsigma_{\mathrm{1}}(\mathbold{FP}) +
\Upsigma_{\mathrm{1}}(\mathbold{FN}) +
\Upsigma_{\mathrm{1}}(\mathbold{TN}) = nm$,
so $\mathrm{LabelAccuracy} = 1 -
(\Upsigma_{\mathrm{1}}(\mathbold{FP}) +
\Upsigma_{\mathrm{1}}(\mathbold{FN}))/(nm) = 1 -
\mathrm{BS}^{\mathrm{micro}}$.
\end{proof}

\begin{remark}
Theorem~\ref{thm:bridge}(c) is the calibration analogue of the
trivial identity $\mathrm{AccuracyRate} + \mathrm{ErrorRate} = 1$.
It shows that for binary hard predictions, the Brier score carries
no information beyond Hamming loss. The two frameworks diverge
precisely when predictions are \emph{soft} ($\hat{Y}_{i,j} \in
(0,1)$): in that regime $(Y_{i,j} - \hat{Y}_{i,j})^{2}$ no
longer equals $Y_{i,j} \oplus \hat{Y}_{i,j}$, and the calibration
matrix $\mathbold{\mathrm{Cal}}$ captures confidence quality that
the Boolean operators of the Discrimination Framework discard.
\end{remark}

Remark~\ref{rem:qwk} in Appendix~\ref{app:qwk} extends the deviation view to ordinal labels and quadratic weighted kappa.

% ============================================================
\subsection{Micro, Macro, and Exemplar Brier: A Collapse}
\label{sec:brier_collapse}
% ============================================================

The Calibration Framework differs from the Discrimination Framework in
the behaviour of the three aggregation operators on the resulting
scalar measures.

\begin{theorem}[Micro-macro-exemplar collapse for Brier]
\label{thm:brier_collapse}
\[
  \mathrm{BS}^{\mathrm{micro}}
  \;=\;
  \mathrm{BS}^{\mathrm{macro}}
  \;=\;
  \mathrm{BS}^{\mathrm{exemplar}}.
\]
\end{theorem}

\begin{proof}
By commutativity of finite summation:
\begin{align*}
  \mathrm{BS}^{\mathrm{macro}}
  &= \frac{1}{m}\sum_{j} \frac{1}{n}\sum_{i}(Y_{i,j}-\hat{Y}_{i,j})^{2}
   = \frac{1}{nm}\sum_{i,j}(Y_{i,j}-\hat{Y}_{i,j})^{2}
   = \mathrm{BS}^{\mathrm{micro}}.
\end{align*}
The argument for exemplar is identical, swapping the order of summation over $i$ and $j$.
\end{proof}

\noindent This collapse follows from the additivity of the aggregation operators applied to a fixed matrix of per-cell losses, together with uniform normalisation: the Brier score is a \emph{sum} of per-cell squared deviations divided by the fixed constant $nm$, so the order of aggregation is irrelevant. The squared-deviation cell operator itself need not be linear.

\begin{remark}[Contrast with the Discrimination Framework]
\label{rem:contrast_collapse}

\begin{enumerate}
\item In the Discrimination Framework, micro- and macro-averages generally differ because many performance measures are \emph{ratios} of aggregated quantities. For precision, provided that each classwise denominator $[\Upsigma_{\mathrm{m}}(\mathbold{TP}+\mathbold{FP})]_j$ is strictly positive,
$M^{\mathrm{micro}} = \Upsigma_{\mathrm{1}}(\mathbold{TP})/\Upsigma_{\mathrm{1}}(\mathbold{TP}+\mathbold{FP})$
is a ratio of global sums, whereas
$M^{\mathrm{macro}}=(1/m)\sum_j [\Upsigma_{\mathrm{m}}(\mathbold{TP})]_j/[\Upsigma_{\mathrm{m}}(\mathbold{TP}+\mathbold{FP})]_j$
is an average of per-class ratios. The two coincide for every choice of classwise numerators when the positive per-class denominators are equal (cf. Corollary~1 of \citealt{prati2026unified}); for a given data set, they coincide if and only if the denominators and the per-class precision values are empirically uncorrelated.

\item For Brier, the per-class and per-example losses are averages of the same per-cell squared deviations with fixed denominators $n$ and $m$, respectively. Their subsequent unweighted averages therefore each equal the single sum normalised by $nm$, as established directly in Theorem~\ref{thm:brier_collapse}. This identity follows from uniform finite-sum aggregation; it does not require applying the ratio-based weighted-macro theorem of \citet{prati2026unified} to the Brier score.

\item Consequently, the three aggregation operators provide no additional information at the level of the scalar Brier score itself. Their contribution lies one level deeper: column- and row-wise aggregation yield two \emph{different Murphy decompositions} of the same scalar, each revealing a different aspect of the predictions, as developed in Sections~\ref{sec:murphy_col} and~\ref{sec:three_murphy}, respectively.
\end{enumerate}

\end{remark}

\needspace{8\baselineskip}

% ============================================================
\subsection{Murphy Decomposition via Column Aggregation}
\label{sec:murphy_col}
% ============================================================

The Murphy decomposition~\citep{winkler1968good,murphy1973new} expresses the
Brier score for a single binary variable as the sum of three
interpretable terms: \emph{uncertainty} (inherent variability of
the labels), \emph{reliability} (calibration error), and
\emph{resolution} (discriminative power of the forecasts). We
re-derive this decomposition within the matrix framework,
connecting each term to column-wise aggregation $\Upsigma_{\mathrm{m}}$.

\begin{mydef}[Column-level Murphy terms]
\label{def:murphy_col}
For class $j \in [1,m]$, let
$\bar{y}_{j} = [\Upsigma_{\mathrm{m}}(\mathbold{Y})]_{j}/n$
denote the empirical label frequency for class~$j$, and let
\begin{equation}
  f_{j}(p)
  \;=\;
  \mathbb{E}\!\left[Y_{i,j} \;\middle|\; \hat{Y}_{i,j} = p\right]
  \label{eq:cal_fn}
\end{equation}
be the \emph{calibration function} for class~$j$: the true
conditional probability that label~$j$ is relevant, given that
the classifier predicts probability~$p$. In the finite-sample
setting, $f_{j}$ is defined empirically as
$f_{j}(p) = n_{p}^{-1}\sum_{i:\hat{Y}_{i,j}=p} Y_{i,j}$, where
$n_{p} = |\{i : \hat{Y}_{i,j} = p\}|$.

\noindent Define:
\begin{align}
  \mathrm{UNC}_{j}
    &= \bar{y}_{j}(1 - \bar{y}_{j})
    \label{eq:unc_col}\\
  \mathrm{RES}_{j}
    &= \frac{1}{n}\sum_{i=1}^{n}
       \bigl(f_{j}(\hat{Y}_{i,j}) - \bar{y}_{j}\bigr)^{2}
    \label{eq:res_col}\\
  \mathrm{REL}_{j}
    &= \frac{1}{n}\sum_{i=1}^{n}
       \bigl(\hat{Y}_{i,j} - f_{j}(\hat{Y}_{i,j})\bigr)^{2}
    \label{eq:rel_col}
\end{align}
\end{mydef}

\noindent $\mathrm{UNC}_{j}$ is the Bernoulli variance of class
$j$'s label distribution: it is maximised when
$\bar{y}_{j} = 0.5$ (maximally uncertain label) and zero for a
label that is always present or always absent.
$\mathrm{RES}_{j}$ measures how much the calibration function
$f_{j}(\hat{Y}_{i,j})$ varies across examples. It measures variability of
the calibration map and ignores its direction: when $f_j$ is increasing, a classifier with high
resolution assigns systematically higher predicted probabilities to positive examples than to
negative ones; a classifier with $\hat{Y}=1-Y$ has maximal $\mathrm{RES}_j=\mathrm{UNC}_j$ (and
$\mathrm{REL}_j=1$).
$\mathrm{REL}_{j}$ measures the mean-squared gap between the
predicted probability $\hat{Y}_{i,j}$ and the calibration
function $f_{j}(\hat{Y}_{i,j})$; a perfectly calibrated
classifier satisfies $\hat{Y}_{i,j} = f_{j}(\hat{Y}_{i,j})$ for
all~$i$, giving $\mathrm{REL}_{j} = 0$.

\begin{lemma}[Group-mean residuals are orthogonal to group-level functions]
\label{lem:group_mean_orthogonal}
Fix $j$ and suppress the subscript. For \emph{any} function $h$ of the predicted value alone,
\begin{equation}
  \sum_{i=1}^{n} h(\hat{Y}_i)\,\bigl(Y_i - f(\hat{Y}_i)\bigr) \;=\; 0.
  \label{eq:group_mean_orthogonal}
\end{equation}
\end{lemma}
\begin{proof}
Group summands by unique predicted value $v \in
\mathrm{range}(\hat{Y}_{\cdot,j})$, letting
$\mathcal{I}_{v} = \{i : \hat{Y}_{i,j} = v\}$:
\[
  \sum_{i \in \mathcal{I}_{v}} h(v)\bigl(Y_i - f(v)\bigr)
  = h(v) \sum_{i \in \mathcal{I}_{v}}(Y_i - f(v))
  = 0,
\]
since by definition of $f$, $\sum_{i \in \mathcal{I}_{v}} Y_i =
|\mathcal{I}_{v}| \cdot f(v)$, so $\sum_{i \in
\mathcal{I}_{v}}(Y_i - f(v)) = 0$ regardless of $h(v)$. Summing over all $v$ gives
Eq.~(\ref{eq:group_mean_orthogonal}).
\end{proof}

\noindent The Murphy decomposition below uses this orthogonality of group-mean residuals to
every function of the group index twice, once against $h(v) = f(v) - v$ and once
against $h(v) = f(v) - \bar{y}$.

\begin{theorem}[Column-level Murphy decomposition]
\label{thm:murphy_col}
For every class $j \in [1,m]$, for binary $Y_{i,j} \in \{0,1\}$
and soft predictions $\hat{Y}_{i,j} \in [0,1]$, with $f_j$ as in
Definition~\ref{def:murphy_col}:
\begin{equation}
  \mathrm{BS}_{j}^{\mathrm{class}}
  \;=\;
  \mathrm{UNC}_{j} \;-\; \mathrm{RES}_{j} \;+\; \mathrm{REL}_{j}.
  \label{eq:murphy_col}
\end{equation}
\end{theorem}

\begin{proof}
Fix $j$ and suppress the subscript. Add and subtract
$f(\hat{Y}_i)$ inside the squared loss:
\begin{equation}
  (Y_i - \hat{Y}_i)^{2}
  = \underbrace{(Y_i - f(\hat{Y}_i))^{2}}_{\text{irreducible residual}}
  + \underbrace{(f(\hat{Y}_i) - \hat{Y}_i)^{2}}_{\text{calibration error}}
  + 2\underbrace{(Y_i - f(\hat{Y}_i))(f(\hat{Y}_i) - \hat{Y}_i)}_{\text{cross term}}.
  \label{eq:three_terms}
\end{equation}

The cross term vanishes: applying Lemma~\ref{lem:group_mean_orthogonal} with $h(v) = f(v) - v$ gives
$\sum_{i}(f(\hat{Y}_i) - \hat{Y}_i)(Y_i - f(\hat{Y}_i)) = 0$.
Averaging (\ref{eq:three_terms}) over $i$ therefore gives
\[
  \mathrm{BS} \;=\; \frac{1}{n}\sum_i(Y_i - f(\hat{Y}_i))^{2} + \mathrm{REL}.
\]

It remains to show that the first term equals $\mathrm{UNC} - \mathrm{RES}$.
Apply the finite-sample law of total variance to $Y$ and $f(\hat{Y})$:
\begin{equation}
  \underbrace{\frac{1}{n}\sum_{i}(Y_i - \bar{y})^{2}}_{\mathrm{UNC}}
  \;=\;
  \underbrace{\frac{1}{n}\sum_{i}(f(\hat{Y}_i) - \bar{y})^{2}}_{\mathrm{RES}}
  \;+\;
  \underbrace{\frac{1}{n}\sum_{i}(Y_i - f(\hat{Y}_i))^{2}}_{\displaystyle A}.
  \label{eq:total_var}
\end{equation}
The left side is $\mathrm{UNC} = \bar{y}(1-\bar{y})$ for binary
$Y$. The finite-sample law of total variance holds exactly:
\begin{equation*}
  \frac{1}{n}\sum_i(Y_i - \bar{y})^{2}
  = \frac{1}{n}\sum_i(f(\hat{Y}_i)-\bar{y})^{2}
  + \frac{1}{n}\sum_i(Y_i - f(\hat{Y}_i))^{2},
\end{equation*}
which follows from the identity
$(Y_i - \bar{y})^2 = (f(\hat{Y}_i)-\bar{y})^2 +
(Y_i - f(\hat{Y}_i))^2 + 2(f(\hat{Y}_i)-\bar{y})(Y_i-f(\hat{Y}_i))$
together with Lemma~\ref{lem:group_mean_orthogonal} applied once more, now with
$h(v) = f(v) - \bar{y}$: the cross term $\sum_i(f(\hat{Y}_i)-\bar{y})(Y_i - f(\hat{Y}_i))$ vanishes
by the same argument. Hence $A = \mathrm{UNC} - \mathrm{RES}$, and
\[
  \mathrm{BS}
  = A + \mathrm{REL}
  = (\mathrm{UNC} - \mathrm{RES}) + \mathrm{REL}. \qedhere
\]
\end{proof}

\begin{remark}[Relation to binned reliability diagrams]
\label{rem:binning}
In practice, the population calibration function $f_j(p)$ cannot be determined exactly from a finite sample. If all observed predictions for class $j$ are distinct, each empirical level set $\mathcal{I}_{v}=\{i:\hat{Y}_{i,j}=v\}$ contains exactly one example for every observed value $v$. In that case $f_j(\hat{Y}_{i,j})=Y_{i,j}$, $\mathrm{REL}_j=\mathrm{BS}_j$ and $\mathrm{RES}_j=\mathrm{UNC}_j$, so the unbinned empirical decomposition is trivial. A continuous prediction scale does not by itself ensure distinct observations. For example, with $\hat{Y}_{\cdot,j}=(0.2,0.5,0.5,0.8)$ and $Y_{\cdot,j}=(0,0,1,1)$, the two examples predicted at $0.5$ form the same level set, giving $f_j(0.5)=0.5$ even though their individual labels are $0$ and $1$.
The standard approach is to group predictions into $K$ bins $G_1,\ldots,G_K$ (e.g.\ equal-width or equal-frequency) and approximate $f_j(p)\approx\bar{y}_{j,k}=n_{j,k}^{-1}\sum_{i\in G_k}Y_{i,j}$ for $p\in G_k$.
Here and in the binned sums below, bin-level means are evaluated only for nonempty bins ($n_{j,k}>0$); an empty bin has no defined sample mean and contributes no observations.
Instead of splicing this approximation into the formulae of Definition~\ref{def:murphy_col}, we use the binwise orthogonality argument underlying Theorem~\ref{thm:murphy_col}, keeping the bin identities distinct, to obtain an exact decomposition for the \emph{discretised predictor} obtained by replacing every $\hat{Y}_{i,j}$ with its own bin mean $\bar{p}_{j,\kappa(i)}$, where $\kappa(i)$ denotes the bin containing example $i$.
A binning-free alternative is to estimate $f_j$ by isotonic regression (Proposition~\ref{prop:murphy_isotonic}), which does not degenerate when the scores are distinct.
\end{remark}

\needspace{7\baselineskip}

\begin{proposition}[Exact binned Murphy decomposition]
\label{prop:murphy_binned}
Let $\widehat{\mathrm{BS}}_{j}
= \frac{1}{n}\sum_{i}\bigl(Y_{i,j}-\bar{p}_{j,\kappa(i)}\bigr)^{2}$
denote the Brier score of the discretised predictor. Then, exactly and
with no residual term,
\begin{equation}
  \widehat{\mathrm{BS}}_{j}
  \;=\;
  \mathrm{UNC}_{j}
  \;-\;
  \underbrace{\frac{1}{n}\sum_{k}n_{j,k}\bigl(\bar{y}_{j,k}-\bar{y}_{j}\bigr)^{2}}_{\widehat{\mathrm{RES}}_{j}}
  \;+\;
  \underbrace{\frac{1}{n}\sum_{k}n_{j,k}\bigl(\bar{y}_{j,k}-\bar{p}_{j,k}\bigr)^{2}}_{\widehat{\mathrm{REL}}_{j}}.
  \label{eq:murphy_binned}
\end{equation}
\end{proposition}

\begin{proof}
Fix $j$ and abbreviate $a_k=\bar{y}_{j,k}$ and
$b_k=\bar{p}_{j,k}$ for each nonempty bin $G_k$. Since
$\sum_{i\in G_k}(Y_{i,j}-a_k)=0$ and $Y_{i,j}$ is binary,
expansion within each bin gives
\begin{equation*}
\sum_{i\in G_k}(Y_{i,j}-b_k)^2
=n_{j,k}a_k(1-a_k)+n_{j,k}(a_k-b_k)^2.
\end{equation*}
Moreover, $\sum_k n_{j,k}a_k=n\bar{y}_j$, and hence
\begin{equation*}
\sum_k n_{j,k}a_k(1-a_k)
=n\bar{y}_j(1-\bar{y}_j)
-\sum_k n_{j,k}(a_k-\bar{y}_j)^2.
\end{equation*}
Summing the first identity over the bins, inserting the second,
and dividing by $n$ proves~(\ref{eq:murphy_binned}).
The argument groups observations by \emph{bin identity}, not by
unique discretised prediction value. It therefore remains valid even
if two different bins have the same mean prediction $b_k$; no
identification of those bins is required.
\end{proof}

\needspace{18\baselineskip}
\begin{remark}[\texorpdfstring{$\widehat{\mathrm{BS}}_j$}{BS-hat\_j} is the score of the discretised predictor, not of \texorpdfstring{$\hat{\mathbold{Y}}$}{Yhat} itself]
\label{rem:binning_gap}

\begin{enumerate}
\item Proposition~\ref{prop:murphy_binned} decomposes $\widehat{\mathrm{BS}}_j$ exactly, but $\widehat{\mathrm{BS}}_j$ is generally \emph{not} equal to the true per-class Brier score $\mathrm{BS}_j^{\mathrm{class}}$ of Equation~(\ref{eq:bs_class}), which uses the original, continuous $\hat{Y}_{i,j}$: replacing $\hat{Y}_{i,j}$ by its bin mean discards the within-bin variation of the predictor.

\item The gap $\mathrm{BS}_j^{\mathrm{class}}-\widehat{\mathrm{BS}}_j$ is \emph{not} the within-bin prediction variance $B_j=\frac{1}{n}\sum_k\sum_{i\in G_k}(\hat{Y}_{i,j}-\bar{p}_{j,k})^2\geq0$ of a naive residual accounting. Writing $\hat{Y}_{i,j}=\bar{p}_{j,k}+\delta_i$ within bin $G_k$ (so that the $\delta_i$ sum to zero in each bin), the gap is exactly $\mathrm{BS}_j^{\mathrm{class}}-\widehat{\mathrm{BS}}_j=B_j-2C_j$, where $C_j=\frac1n\sum_k\sum_{i\in G_k}\delta_i\,(Y_{i,j}-\bar{y}_{j,k})$ is the within-bin covariance between the prediction and the label. $C_j$ is not sign-definite (it is positive when the predictor still discriminates inside the bins, and can be negative), so the gap does not admit the ``approximation-error'' reading that $B_j$ would give on its own.

\item The gap shrinks as the bins narrow: $\widehat{\mathrm{BS}}_j=\mathrm{BS}_j^{\mathrm{class}}$ exactly, and Proposition~\ref{prop:murphy_binned} reduces to Theorem~\ref{thm:murphy_col}, whenever each bin contains only examples with identical predicted values. For a fixed finite $K$, however, Proposition~\ref{prop:murphy_binned} should be read as an exact statement about the \emph{binned} score (the quantity a reliability diagram with $K$ bins displays) rather than as a controlled, signed approximation to $\mathrm{BS}_j^{\mathrm{class}}$. The reliability diagram plots $(\bar{p}_{j,k},\,\bar{y}_{j,k})$ for each bin $k$; $\widehat{\mathrm{REL}}_j$ is the area-weighted mean-squared deviation of these points from the diagonal. Explicitly, as already defined in Proposition~\ref{prop:murphy_binned},
\begin{equation*}
  \widehat{\mathrm{REL}}_j=\frac{1}{n}\sum_{k:\,n_{j,k}>0}n_{j,k}\bigl(\bar{y}_{j,k}-\bar{p}_{j,k}\bigr)^2.
\end{equation*}
\end{enumerate}

\end{remark}

\needspace{12\baselineskip}

\begin{proposition}[Isotonic Murphy decomposition]
\label{prop:murphy_isotonic}
Fix a class $j$ and write $p_i=\hat{Y}_{i,j}$, $y_i=Y_{i,j}$, $\bar{y}=\bar{y}_j$. Let $\hat{f}_j$ be the
least-squares non-decreasing regression of $y$ on $p$ (the pool-adjacent-violators fit, computed on the
level sets of $p$), and put
\[
  \mathrm{RES}_j^{\mathrm{iso}}=\frac1n\sum_i\bigl(\hat{f}_j(p_i)-\bar{y}\bigr)^2,\qquad
  \mathrm{REL}_j^{\mathrm{iso}}=\mathrm{BS}_j^{\mathrm{class}}-\frac1n\sum_i\bigl(y_i-\hat{f}_j(p_i)\bigr)^2 .
\]
Then:
\begin{enumerate}
\item[\textup{(a)}] $\mathrm{BS}_j^{\mathrm{class}}=\mathrm{UNC}_j-\mathrm{RES}_j^{\mathrm{iso}}+\mathrm{REL}_j^{\mathrm{iso}}$
exactly, and $\mathrm{RES}_j^{\mathrm{iso}}=\mathrm{UNC}_j-\frac1n\sum_i(y_i-\hat{f}_j(p_i))^2$.
\item[\textup{(b)}] $0\le\mathrm{REL}_j^{\mathrm{iso}}\le\mathrm{REL}_j$ and $0\le\mathrm{RES}_j^{\mathrm{iso}}\le\mathrm{RES}_j$, where
$\mathrm{REL}_j$ and $\mathrm{RES}_j$ are the level-set terms of Theorem~\ref{thm:murphy_col}.
\item[\textup{(c)}] $\mathrm{REL}_j^{\mathrm{iso}}=\frac1n\sum_i(p_i-\hat{f}_j(p_i))^2+\frac2n\sum_i(y_i-\hat{f}_j(p_i))(\hat{f}_j(p_i)-p_i)$,
and the second term need not vanish, so $\mathrm{REL}_j^{\mathrm{iso}}$ is in general not the mean squared distance between the
scores and the fitted calibration map.
\end{enumerate}
\end{proposition}

\begin{proof}
The fit $\hat{f}_j$ is constant on blocks of adjacent level sets, and its value on a block is the mean of $y$ over that block.
Hence $\sum_i(y_i-\hat{f}_j(p_i))g(p_i)=0$ for every $g$ that is constant on the blocks, in particular for
$g=\hat{f}_j-\bar{y}$. Expanding $y-\hat{f}_j=(y-\bar{y})-(\hat{f}_j-\bar{y})$ and using this orthogonality gives
$\frac1n\sum_i(y_i-\hat{f}_j(p_i))^2=\mathrm{UNC}_j-\mathrm{RES}_j^{\mathrm{iso}}$, which is the second claim of (a); the first is the
definition of $\mathrm{REL}_j^{\mathrm{iso}}$. For (b), the identity map is a non-decreasing function of $p$, so the isotonic fit cannot
have a larger squared error than $p$ itself, which gives $\mathrm{REL}_j^{\mathrm{iso}}\ge0$. The level-set mean $f_j$ of Theorem~\ref{thm:murphy_col}
minimises the squared error over all functions of $p$, so it is no worse than $\hat{f}_j$; the same orthogonality for $f_j$ gives
$\mathrm{RES}_j=\mathrm{UNC}_j-\mathrm{BS}(f_j)$ and $\mathrm{REL}_j=\mathrm{BS}_j^{\mathrm{class}}-\mathrm{BS}(f_j)$, from which both inequalities follow.
(c) is the expansion of $(y-p)^2=((y-\hat{f}_j)+(\hat{f}_j-p))^2$.
\end{proof}

\begin{remark}[When to use the isotonic terms]
\label{rem:isotonic}
When all scores of a column are distinct, the level-set terms degenerate ($\mathrm{REL}_j=\mathrm{BS}_j^{\mathrm{class}}$,
$\mathrm{RES}_j=\mathrm{UNC}_j$; Remark~\ref{rem:binning}), whereas $\mathrm{REL}_j^{\mathrm{iso}}$ and $\mathrm{RES}_j^{\mathrm{iso}}$ are
generally informative and need neither bins nor a bin count. The price is the monotonicity assumption: $\hat{f}_j$ is the best non-decreasing
calibration map, so a score that is miscalibrated in a non-monotone way has part of that error absorbed into
$\mathrm{RES}_j^{\mathrm{iso}}$ (Proposition~\ref{prop:murphy_isotonic}(b)). The estimator is the one used by the CORP
approach of \citet{dimitriadis2021stable}, for which they establish its statistical properties; Proposition~\ref{prop:murphy_isotonic} states the
squared-error case in the notation of this paper, for the columns and, with the same proof, the rows of a multilabel matrix. Like the level-set terms,
the isotonic terms are not skew-invariant in general, since $\hat{f}_j$ depends on the prevalence of the class; replicating the positives of a class changes both of them in
a numerical check, and we do not state a sharper result. The pair $(\alpha_j,\beta_j)$ remains the skew-invariant calibration summary.
\end{remark}

% ============================================================
\subsection{Column- and Row-Level Murphy Decompositions}
\label{sec:three_murphy}
% ============================================================

Theorem~\ref{thm:murphy_col} decomposes the per-class Brier score
via $\Upsigma_{\mathrm{m}}$. The same decomposition can be applied
along the \emph{row axis} using $\Upsigma_{\mathrm{n}}$, yielding
a qualitatively different calibration diagnosis.

\begin{mydef}[Row-level Murphy terms]
\label{def:murphy_row}
For example $i \in [1,n]$, let
$\bar{Y}_{i} = [\Upsigma_{\mathrm{n}}(\mathbold{Y})]_{i}/m$
denote the empirical label density of example~$i$ (fraction of
relevant labels), and let
\begin{equation}
  g_{i}(p)
  \;=\;
  \mathbb{E}\!\left[Y_{i,j} \;\middle|\; \hat{Y}_{i,j} = p\right]
  \label{eq:cal_fn_row}
\end{equation}
be the \emph{per-example calibration function}: the conditional
probability that a label is relevant for example~$i$ given that
the classifier predicts~$p$ for that label. In the
finite-sample setting,
$g_{i}(p) = m_{p}^{-1}\sum_{j:\hat{Y}_{i,j}=p} Y_{i,j}$, where
$m_{p} = |\{j : \hat{Y}_{i,j} = p\}|$.

\noindent Define:
\begin{align}
  \mathrm{UNC}_{i}^{\mathrm{ex}}
    &= \bar{Y}_{i}(1 - \bar{Y}_{i})
    \label{eq:unc_row}\\
  \mathrm{RES}_{i}^{\mathrm{ex}}
    &= \frac{1}{m}\sum_{j=1}^{m}
       \bigl(g_{i}(\hat{Y}_{i,j}) - \bar{Y}_{i}\bigr)^{2}
    \label{eq:res_row}\\
  \mathrm{REL}_{i}^{\mathrm{ex}}
    &= \frac{1}{m}\sum_{j=1}^{m}
       \bigl(\hat{Y}_{i,j} - g_{i}(\hat{Y}_{i,j})\bigr)^{2}
    \label{eq:rel_row}
\end{align}
\end{mydef}

\begin{theorem}[Row-level Murphy decomposition]
\label{thm:murphy_row}
For every example $i \in [1,n]$, for binary $Y_{i,j} \in \{0,1\}$
and soft predictions $\hat{Y}_{i,j} \in [0,1]$, with $g_i$ as in
Definition~\ref{def:murphy_row}:
\begin{equation}
  \mathrm{BS}_{i}^{\mathrm{example}}
  \;=\;
  \mathrm{UNC}_{i}^{\mathrm{ex}}
  \;-\; \mathrm{RES}_{i}^{\mathrm{ex}}
  \;+\; \mathrm{REL}_{i}^{\mathrm{ex}}.
  \label{eq:murphy_row}
\end{equation}
\end{theorem}

\begin{proof}
The proof is structurally identical to that of
Theorem~\ref{thm:murphy_col}, with the roles of $n$ (examples)
and $m$ (labels) exchanged. Fix $i$ and suppress its subscript.
Add and subtract $g(\hat{Y}_{i,j})$ in the squared loss, group by
unique predicted values within the $m$ labels of example~$i$, and
apply the same two steps: (1) the cross term vanishes because
$g(p)$ is the conditional mean of $Y_{i,j}$ over labels~$j$ with
$\hat{Y}_{i,j} = p$; (2) the finite-sample law of total variance
gives $\frac{1}{m}\sum_j(Y_{i,j}-g(\hat{Y}_{i,j}))^{2} =
\mathrm{UNC}_{i}^{\mathrm{ex}} - \mathrm{RES}_{i}^{\mathrm{ex}}$.
\end{proof}

The row axis also admits a class-conditional decomposition of
$\mathrm{BS}_i^{\mathrm{example}}$ analogous to
Proposition~\ref{prop:bs_decomp}, obtained by conditioning on the label
\emph{value} $Y_{i,j}$ instead of the class index: the row-axis counterpart
of the class prevalence $\pi_j$ is $k_i/m$, example $i$'s own label
density.

\begin{mydef}[Row-axis calibration errors]
\label{def:cond_cal_row}
For example $i$ with label cardinality
$k_i = [\Upsigma_{\mathrm{n}}(\mathbold{Y})]_i = \sum_j Y_{i,j}$, and
$0 < k_i < m$, define
\begin{align}
  \alpha_i
    &= \frac{1}{k_i}\sum_{j:\,Y_{i,j}=1}(1-\hat{Y}_{i,j})^{2}
     = \mathbb{E}\!\left[(1-\hat{Y}_{i,j})^{2}\;\middle|\;Y_{i,j}=1\right],
  \label{eq:alpha_row}\\[4pt]
  \beta_i
    &= \frac{1}{m-k_i}\sum_{j:\,Y_{i,j}=0}\hat{Y}_{i,j}^{2}
     = \mathbb{E}\!\left[\hat{Y}_{i,j}^{2}\;\middle|\;Y_{i,j}=0\right].
  \label{eq:beta_row}
\end{align}
(For the boundary cases $k_i \in \{0, m\}$, the empty conditioning class
contributes no term to~(\ref{eq:bs_decomp_row}) below.)
\end{mydef}

\begin{proposition}[Prevalence decomposition of per-example Brier]
\label{prop:bs_decomp_row}
For binary $Y_{i,j} \in \{0,1\}$:
\begin{equation}
  \mathrm{BS}_i^{\mathrm{example}}
  \;=\; \frac{k_i}{m}\,\alpha_i \;+\; \frac{m-k_i}{m}\,\beta_i.
  \label{eq:bs_decomp_row}
\end{equation}
\end{proposition}

\begin{proof}
Identical to the proof of Proposition~\ref{prop:bs_decomp}, with the roles of
$n$ (examples) and $m$ (labels) exchanged. Split
$\mathrm{BS}_i^{\mathrm{example}} = \tfrac{1}{m}\sum_j(Y_{i,j}-\hat{Y}_{i,j})^2$
by the value of $Y_{i,j}$:
\begin{align*}
  \mathrm{BS}_i^{\mathrm{example}}
  &= \frac{1}{m}\!\!\sum_{j:\,Y_{i,j}=1}\!\!(1-\hat{Y}_{i,j})^{2}
   + \frac{1}{m}\!\!\sum_{j:\,Y_{i,j}=0}\!\!\hat{Y}_{i,j}^{2}
   = \frac{k_i}{m}\,\alpha_i + \frac{m-k_i}{m}\,\beta_i. \qedhere
\end{align*}
\end{proof}

\begin{remark}[Not paired with a skew-invariance theorem]
\label{rem:bs_decomp_row_no_skew}
Unlike its column-axis counterpart (Theorem~\ref{thm:cal_skew}),
Proposition~\ref{prop:bs_decomp_row} is not paired with a skew-invariance
result: the skew transformation of Definition~\ref{def:cal_skew} rescales
class subpopulations \emph{across} examples and has no natural analogue
acting on a single example's own $m$ labels
(Remark~\ref{rem:cal_skew_exemplar_scope}). Decomposition
(\ref{eq:bs_decomp_row}) is used here descriptively, to separate an
example's miss-calibration on its relevant labels ($\alpha_i$) from its
false-alarm calibration on its irrelevant ones ($\beta_i$); these are the
quantities that Experiment~6 (Section~\ref{sec:exp_murphy_example}) reports.
\end{remark}

The two decompositions have distinct interpretations,
summarised in Table~\ref{tab:murphy_comparison}.

\begin{table}[!t]
\centering
\small
\setlength{\tabcolsep}{5pt}
\begin{tabular}{@{}l|p{1.55in}|p{1.55in}|p{1.55in}@{}}
\hline
\textbf{Term} & \textbf{Column axis ($\Upsigma_{\mathrm{m}}$)} &
\textbf{Row axis ($\Upsigma_{\mathrm{n}}$)} &
\textbf{Interpretation} \\
\hline
UNC
  & $\bar{y}_{j}(1-\bar{y}_{j})$
  & $\bar{Y}_{i}(1-\bar{Y}_{i})$
  & Inherent difficulty: class prevalence (col)
    vs.\ example label density (row) \\[3pt]
RES
  & Spread of the label frequency across the groups of equal predicted value in a column
  & Same, across the groups of equal predicted value in a row
  & Discrimination: separates pos.\ from neg.\ labels? \\[3pt]
REL
  & Mean-sq.\ calibration error per class
  & Mean-sq.\ calibration error per example
  & Calibration: confidence matches frequency? \\[3pt]
\hline
\end{tabular}
\caption{The two Murphy decompositions induced by the column and
row aggregation operators. Both partition the same per-cell
Brier score, but condition on different axes. Column-axis UNC
is the class-prevalence Bernoulli variance; row-axis UNC is the
example label-density Bernoulli variance. The two are generally
unequal (see Proposition~\ref{prop:unc_inequality}).}
\label{tab:murphy_comparison}
\end{table}

\begin{proposition}[Uncertainty terms are generally unequal]
\label{prop:unc_inequality}
In general,
\[
  \frac{1}{m}\sum_{j=1}^{m}\mathrm{UNC}_{j}
  \;\neq\;
  \frac{1}{n}\sum_{i=1}^{n}\mathrm{UNC}_{i}^{\mathrm{ex}}.
\]
A sufficient condition for equality is that the label matrix
$\mathbold{Y}$ is \emph{doubly balanced}: all class frequencies
are equal ($\bar{y}_{j} = c$ for all $j$) \emph{and} all example
label densities are equal ($\bar{Y}_{i} = c$ for all $i$), for
the same constant $c$. The condition is not necessary.
\end{proposition}

\begin{proof}
Write $f(x) = x(1-x)$. The column-axis macro-UNC is
$\frac{1}{m}\sum_{j}f(\bar{y}_{j})$ and the row-axis exemplar-UNC
is $\frac{1}{n}\sum_{i}f(\bar{Y}_{i})$.

If $\mathbold{Y}$ is doubly balanced with
common constant $c$, both sides equal $f(c)$, which proves sufficiency.

To see that the two sides differ in general, take $m = 1$ and
$\mathbold{Y} = (1, 0, 0)^{\top}$: each row mean
$\bar{Y}_{i} \in \{0, 1\}$ gives $f(\bar{Y}_{i}) = 0$, so the
exemplar side vanishes, while the class side is
$f(1/3) = 2/9 > 0$.

For non-necessity, consider
\[
  \mathbold{Y} =
  \begin{pmatrix} 1 & 1 & 0 \\ 0 & 0 & 1 \\ 0 & 1 & 0 \end{pmatrix},
\]
for which the column means are $(1/3,\, 2/3,\, 1/3)$ and the row means are
$(2/3,\, 1/3,\, 1/3)$; both sides average $f$ to $2/9$, yet
$\mathbold{Y}$ is not doubly balanced.
\end{proof}

\begin{remark}[Calibration vs.\ discrimination diagnostics]
\label{rem:cal_disc}
Theorem~\ref{thm:brier_collapse} shows that all three aggregation
operators yield the same scalar Brier score. But when the score is
decomposed via Murphy, the two axes diverge on the sub-terms:
\begin{itemize}
  \item The column decomposition (via $\Upsigma_{\mathrm{m}}$)
    reveals which \emph{labels} are poorly calibrated ($\mathrm{REL}_{j}$ large),
    which are hard to discriminate ($\mathrm{RES}_{j}$ small), and
    which are intrinsically uncertain ($\mathrm{UNC}_{j}$ high due
    to near-$0.5$ prevalence).
  \item The row decomposition (via $\Upsigma_{\mathrm{n}}$) reveals
    which \emph{examples} are poorly calibrated
    ($\mathrm{REL}_{i}^{\mathrm{ex}}$ large), which have
    ambiguous label sets ($\mathrm{UNC}_{i}^{\mathrm{ex}}$ large,
    when $\bar{Y}_{i} \approx 0.5$), and for which
    the predicted-score groups exhibit different empirical relevance rates, without implying that the ordering is correct
    ($\mathrm{RES}_{i}^{\mathrm{ex}}$ large).
\end{itemize}
A caveat applies to the row decomposition as defined: within one example there are only $m$
labels, and when the observed predicted values are pairwise distinct, the empirical calibration
function $g_i$ is degenerate ($g_i(\hat{Y}_{i,j})=Y_{i,j}$), which gives
$\mathrm{REL}_{i}^{\mathrm{ex}}=\mathrm{BS}_i^{\mathrm{example}}$ and
$\mathrm{RES}_{i}^{\mathrm{ex}}=\mathrm{UNC}_{i}^{\mathrm{ex}}$. Informative row-level terms
need discretised scores (the row analogue of Proposition~\ref{prop:murphy_binned}), an isotonic calibration map
(Proposition~\ref{prop:murphy_isotonic}, whose proof applies to a row, although with only $m$ labels the fit is coarse), or the
$(\alpha_i,\beta_i)$ split of Proposition~\ref{prop:bs_decomp_row}, which is what
Experiment~6 uses. With that understanding, a model can be well-calibrated per class (low
$\mathrm{REL}_{j}$) while being poorly calibrated per example (high
$\mathrm{REL}_{i}^{\mathrm{ex}}$), because the former asks
whether the \emph{marginal} label frequencies match predicted
probabilities, while the latter asks whether, within each example, the predicted values match the frequency of relevant
labels among the labels that share them (a statement about the example's own $m$ labels, not about the
\emph{joint} law of the label vector). This distinction is invisible to
the scalar Brier score and unavailable in the Discrimination Framework.
\end{remark}

% ============================================================
\subsection{Proper Scoring Rules and the Distinction from
the Discrimination Framework}
\label{sec:proper_scoring}
% ============================================================

This subsection concerns \emph{properness}:
a proper scoring rule rewards honest probability estimates and is
minimised in expectation only when the predicted probability
equals the true label probability. Discrimination Framework measures lack this
property; Calibration Framework measures can satisfy it. We
characterise which cell operators yield proper scoring rules.

\begin{mydef}[Proper scoring rule induced by a cell operator]
\label{def:proper}
A cell operator $D: [0,1] \times [0,1] \to \mathbb{R}$ induces a
proper scoring rule if, for every $p \in [0,1]$ and every
$\hat{p} \in [0,1]$:
\[
  p\,D(1, p) + (1-p)\,D(0, p)
  \;\leq\;
  p\,D(1, \hat{p}) + (1-p)\,D(0, \hat{p}).
\]
That is, the expected loss under the true Bernoulli distribution
$\mathrm{Ber}(p)$ is minimised when the predicted value equals
$p$. The rule is \emph{strictly proper} if equality holds only
when $\hat{p} = p$.
\end{mydef}

\begin{proposition}[t-norm operators are not proper]
\label{prop:tnorm_improper}
For any $t$-norm $T$, in particular for the three standard ones
$T \in \{T_M, T_P, T_L\}$, the cell operator $D(y, \hat{y}) = T(y, \hat{y})$ does not induce
a proper scoring rule.
\end{proposition}

\begin{proof}
Every $t$-norm has $T(1,q)=q$ and $T(0,q)=0$ for $q\in[0,1]$.
Consequently, when $Y\sim\mathrm{Ber}(p)$, its expected \emph{loss} as
specified in Definition~\ref{def:proper} is
\[
  L_p(q)=p\,T(1,q)+(1-p)\,T(0,q)=pq.
\]
For any $p\in(0,1)$, the truthful report $q=p$ incurs loss $p^2>0$,
whereas the admissible report $q=0$ incurs loss $0$.
Thus truth-telling does not minimise the expected loss, and $D=T$ is not
proper. The argument applies to every $t$-norm, including $T_M$, $T_P$,
and $T_L$, because their binary-outcome sections coincide.
\end{proof}

\begin{theorem}[Brier score is strictly proper]
\label{thm:brier_proper}
The calibration cell operator $D(y, \hat{y}) = (y - \hat{y})^{2}$
induces a strictly proper scoring rule: for any $p \in (0,1)$ and any $\hat{p} \in [0,1]$,
letting $L(\hat{p}) := p(1-\hat{p})^{2} + (1-p)\hat{p}^{2}$ denote the expected loss,
\[
  L(\hat{p}) \;\ge\; L(p), \qquad \text{with equality if and only if } \hat{p} = p.
\]
\end{theorem}

\begin{proof}
Completing the square directly, for every $\hat{p}$,
\[
  L(\hat{p}) = p(1-\hat{p})^{2} + (1-p)\hat{p}^{2} = (\hat{p}-p)^{2} + p(1-p),
\]
an identity verified by expanding both sides. Since $L(p) = p(1-p)$, this gives
$L(\hat{p}) - L(p) = (\hat{p}-p)^{2} \ge 0$, with equality iff $\hat{p} = p$: the expected loss
decomposes exactly into a squared-bias term and the irreducible variance $p(1-p)$. The same
decomposition reappears, one level down, in the Murphy decomposition of
Theorem~\ref{thm:murphy_col}. (The usual calculus argument is an immediate corollary of this
identity: $L'(\hat{p}) = 2(\hat{p}-p)$ vanishes uniquely at $\hat{p}=p$, and
$L''(\hat{p}) = 2 > 0$ confirms it is a minimum.)
\end{proof}

\begin{theorem}[Savage representation in the matrix framework
  {\normalfont\citep{savage1971elicitation,gneiting2007strictly,kull2015novel}}]
\label{thm:savage}
A cell operator $D$ whose two sections $D(1,\cdot)$ and $D(0,\cdot)$ are
continuous on $[0,1]$ induces a
proper scoring rule if and only if there exists a concave function
$G: [0,1] \to \mathbb{R}$, differentiable on $[0,1]$ (one-sidedly at the endpoints), such that, for
$y\in\{0,1\}$,
\begin{equation}
  D(y, \hat{y})
  \;=\;
  G(\hat{y})
  \;+\; (y - \hat{y})\,G'(\hat{y}).
  \label{eq:savage}
\end{equation}
The rule is strictly proper iff $G$ is strictly concave. Only the values of
$D$ at $y\in\{0,1\}$ enter the expected loss, so the representation is a
statement about those two sections; nothing is asserted about $y\in(0,1)$.
\end{theorem}

\begin{proofsketch}
This is the representation theorem of \citet{savage1971elicitation}
for binary outcomes, stated in the cell-operator notation
of this paper. Savage and
\citet{gneiting2007strictly} state the representation for
positively-oriented scores (maximised at the truth) with a
\emph{convex} generator; Definition~\ref{def:proper} is stated
for losses (minimised at the truth), so the generator here is the
negative of theirs and is \emph{concave}.

For sufficiency, substitute (\ref{eq:savage}) into the expected loss:
$\mathbb{E}_{Y \sim \mathrm{Ber}(p)}[D(Y,\hat{p})] =
G(\hat{p}) + (p - \hat{p})\,G'(\hat{p})$, and the gradient
inequality for concave $G$ gives
$G(\hat{p}) + (p-\hat{p})G'(\hat{p}) \geq G(p)$, with the minimum
attained at $\hat{p} = p$.

For necessity, we use the Bayes risk and its supporting slopes. Write $\ell_1=D(1,\cdot)$, $\ell_0=D(0,\cdot)$
and $L_p(x)=p\,\ell_1(x)+(1-p)\,\ell_0(x)$, which is affine in $p$. Properness says
$G(p):=L_p(p)=\min_{x\in[0,1]}L_p(x)$. Put $d(x)=\ell_1(x)-\ell_0(x)$, a continuous function;
since $L_{p'}(x)=G(x)+(p'-x)\,d(x)$, properness gives the supergradient inequality
\begin{equation}
  G(p')\;\le\;G(x)+(p'-x)\,d(x)\qquad\text{for all }p',x\in[0,1].
  \label{eq:supergradient}
\end{equation}
The function $G$ is concave: for $r=\lambda p+(1-\lambda)p'$, affinity in $p$ gives
$G(r)=\lambda L_p(r)+(1-\lambda)L_{p'}(r)\ge\lambda G(p)+(1-\lambda)G(p')$.
It is also differentiable: applying (\ref{eq:supergradient}) to the pairs
$(y,x)$ and $(x,y)$ gives $(y-x)\,d(y)\le G(y)-G(x)\le(y-x)\,d(x)$, so the slope of $G$ between $x$
and $y$ lies between $d(y)$ and $d(x)$, and differs from $d(x)$ by at most $|d(y)-d(x)|$, which
tends to $0$ as $y\to x$ by the continuity of $d$. (Only the continuity of the two sections is used.)
Finally, solving $G=p\,\ell_1+(1-p)\,\ell_0$ at $p=x$ together with $G'=\ell_1-\ell_0$ gives
$\ell_1=G+(1-x)G'$ and $\ell_0=G-xG'$, which is (\ref{eq:savage}).
For the strict case, if the rule is strictly proper then (\ref{eq:supergradient}) is strict for
$p'\neq x$, so for $p\neq p'$ and $r$ strictly between them, $G(r)=\lambda L_p(r)+(1-\lambda)L_{p'}(r)
>\lambda G(p)+(1-\lambda)G(p')$: $G$ is strictly concave; conversely a strictly concave $G$ makes
the tangent inequality strict. Both directions, with the strict versions, are checked in Lean
(Section~\ref{sec:conclusion}).
\end{proofsketch}

\begin{corollary}[Special cases of the Savage representation]
\label{cor:savage_cases}
\begin{enumerate}
\item[(a)] The strictly concave generator
$G(q)=q(1-q)$ yields the Brier score $D(y,q)=(y-q)^2$ for binary $y$.

\item[(b)] The strictly concave binary-entropy
generator $G(q)=-q\log q-(1-q)\log(1-q)$ gives, for $0<q<1$, the binary log-loss
\[
  D(y,q)=-y\log q-(1-y)\log(1-q)=-\log(1-|y-q|),\qquad y\in\{0,1\}.
\]
The loss extends to $q=0,1$ by the standard endpoint conventions,
including $+\infty$ for a confident wrong prediction. Because its two
sections are not finite and continuous on $[0,1]$, it is an
\emph{extended-valued} example, not a literal instance of the hypotheses
of Theorem~\ref{thm:savage}.

\item[(c)] For symmetric deviation losses, suppose
$D_\phi(y,q)=\phi(y-q)$, with $\phi$ even, convex and finite-valued and
continuous on $[-1,1]$, and differentiable on $(0,1)$. Put
$w(t)=\phi'(t)/t$ for $0<t<1$. Then $D_\phi$ is proper if and only if
\[
  w(t)=w(1-t)\qquad(0<t<1).
\]
It is strictly proper whenever, in addition, $w(t)>0$ on $(0,1)$.
In particular, squared error is the unique proper member of the
convex power-loss family $\phi(t)=|t|^s$ ($s\ge1$), but not the unique
proper symmetric deviation loss.

\item[(d)] For any $t$-norm $T$,
$D(y,q)=T(y,q)$ has no concave Savage representation of the form
(\ref{eq:savage}) under the assumptions of Theorem~\ref{thm:savage}.
Proper probability elicitation is a property of the \emph{proper-scoring
subset} of Calibration Framework operators, not of every operator in
that framework. The Discrimination and Calibration Frameworks thus
serve complementary evaluation purposes, without asserting that their
full operator classes are disjoint or that all calibration losses are proper.
\end{enumerate}
\end{corollary}

\begin{proof}
\noindent\textit{(a)}
Since $G'(q)=1-2q$, the Savage expression is
\[
G(q)+(y-q)G'(q)=q(1-q)+(y-q)(1-2q)
=y-2yq+q^2=(y-q)^2,
\]
using $y^2=y$ for $y\in\{0,1\}$. Also $G''=-2<0$.

\smallskip
\noindent\textit{(b)}
For $0<q<1$,
$G'(q)=\log((1-q)/q)$, and substitution into
(\ref{eq:savage}) gives
$D(1,q)=-\log q$ and $D(0,q)=-\log(1-q)$.
For binary $y$, these are exactly $-\log(1-|y-q|)$.
The corresponding even convex deviation function is
$\phi(t)=-\log(1-|t|)$ for $|t|<1$; it has a kink at $t=0$ and
is unbounded as $|t|\uparrow1$. Thus it belongs to the
\emph{extended-valued} deviation family, whereas the finite-section
Savage equivalence is invoked only on the interior.

\smallskip
\noindent\textit{(c)}
Evenness gives $\phi(-q)=\phi(q)$, so at a true Bernoulli probability
$p$ the expected loss from report $q$ is
\[
 L_p(q)=p\,\phi(1-q)+(1-p)\,\phi(q).
\]
For $0<q<1$ its derivative is
\begin{equation*}
 L_p'(q)=-p(1-q)w(1-q)+(1-p)q\,w(q).
\end{equation*}
If the rule is proper, then for each $0<p<1$ the interior minimiser
$q=p$ satisfies $L_p'(p)=0$. Hence
$p(1-p)[w(p)-w(1-p)]=0$, or $w(p)=w(1-p)$.
Conversely, if $w$ is symmetric, then
\[
 L_p'(q)=w(q)(q-p).
\]
For an even convex $\phi$, $\phi'(q)\ge0$ on $(0,1)$, hence
$w(q)\ge0$. The expected loss therefore decreases (weakly) toward
$q=p$ and increases (weakly) thereafter. Continuity on the closed
interval supplies the endpoint cases $p,q\in\{0,1\}$.
This proves properness; if $w(q)>0$ on $(0,1)$ the monotonicity is
strict away from $p$, giving strict properness.

The examples of part~(c), including the power-loss statement, are verified in Appendix~\ref{app:savage_examples}.

\smallskip
\noindent\textit{(d)}
Any concave Savage representation of the finite-valued sections would
imply properness by Theorem~\ref{thm:savage}. Proposition~\ref{prop:tnorm_improper}
shows that $T(y,q)$ is not proper, giving the claimed exclusion.
\end{proof}

\begin{remark}[Implications for model evaluation]
\label{rem:proper_implications}

\begin{enumerate}
\item Proposition~\ref{prop:tnorm_improper} concerns the expected value of
an individual $t$-norm cell operator; it does not, on its own, establish
a properness statement for aggregate ratios such as precision, $F_1$
or MCC. In general, maximising such decision-oriented measures as a
training or selection criterion does \emph{not guarantee} calibrated
probability reports. Conversely, minimising Brier score or log-loss
alone does not guarantee high precision or recall at a given decision
threshold.

\item The two frameworks therefore provide complementary information, and a thorough evaluation should consider
at least one relevant discrimination measure alongside a suitable
probabilistic loss. Theorem~\ref{thm:savage} expresses the Brier and
log-loss examples through their quadratic and entropy generators,
respectively, and characterises proper losses with continuous binary
sections through concave Bayes-risk generators.
\end{enumerate}

\end{remark}

\begin{remark}[Absolute deviation and the boundary of the Calibration Framework]
\label{rem:mae_nonproper}

\begin{enumerate}
\item The Calibration Framework admits convex deviation operators,
including $\phi(t)=|t|$. This gives
\[
\mathrm{MAE}=(nm)^{-1}\Upsigma_{\mathrm{1}}\bigl(|\mathbold{Y}-\mathbold{\hat{Y}}|\bigr).
\]
Its micro, macro and exemplar averages coincide, as in
Theorem~\ref{thm:brier_collapse}, because uniform aggregation
of fixed per-cell losses is additive; no additivity of $\phi$
itself is required.

\item MAE is not proper. For a binary outcome with true probability $p$ and reported probability
$q\in[0,1]$,
\[
 L_p(q)=p(1-q)+(1-p)q=p+(1-2p)q.
\]
Its minimisers are $q=0$ for $p<1/2$, $q=1$ for $p>1/2$,
and \emph{every} $q\in[0,1]$ when $p=1/2$.
Thus MAE generally elicits a conditional median, not the true
probability. Not every extreme report is optimal: the appropriate
endpoint depends on $p$. The kink of $|t|$ at zero does not by itself
cause this failure: the extended-valued log-loss deviation has a kink
there and is proper, whereas MAE has the nonsymmetric weight
$w(t)=1/t$ (Corollary~\ref{cor:savage_cases}(c)).

\item The exact quadratic identity of Theorem~\ref{thm:murphy_col} uses
conditional means and square-loss variance decomposition and does not
carry over unchanged to MAE. This does \emph{not} preclude analogous
reliability--resolution identities for other proper losses; for example,
log-loss admits an entropy-based decomposition with different component
functions. Nondifferentiability of a deviation function alone does not
rule out such a probability-theoretic interpretation.

\item Within the finite convex symmetric family of
Corollary~\ref{cor:savage_cases}(c), properness is characterised by
symmetry of $w(t)=\phi'(t)/t$. Squared error $\phi(t)=ct^2$
($c>0$) is the simplest example and the only proper power loss,
but nonquadratic proper losses also exist. MAE and quartic error
retain the aggregation collapse while failing properness.
Binary log-loss is another proper deviation loss in the
\emph{extended-valued} sense of Corollary~\ref{cor:savage_cases}(b).
The specific decomposition of Theorem~\ref{thm:murphy_col} is the
square-loss case, not a universal formula for all these operators.
\end{enumerate}

\end{remark}

% ============================================================
\subsection{Skew-Sensitivity of Calibration Measures}
\label{sec:cal_skew}

\citet{prati2026unified} characterise which
Discrimination Framework measures are immune to class imbalance: a measure is
skew-invariant if and only if it is a function of recall and
specificity alone. This section establishes the calibration
analogue. For Brier-type class-indexed measures, the relevant
skew-invariant summaries are the class-conditional squared errors;
for general measures, Theorem~\ref{thm:cal_skew}(a) instead requires
the full within-class score distributions.

\subsubsection*{Setup}

Fix a class $j$ and suppress the subscript. Let
$n^{+} = \Upsigma_{\mathrm{1}}(\mathbold{Y}_{\cdot j})$ and
$n^{-} = n - n^{+}$ denote the counts of positive and negative
examples for class $j$, and let $\pi = n^{+}/n$ be the class
prevalence. The prevalence $\pi$ coincides with the empirical label
frequency $\bar{y}_{j}$ of Definition~\ref{def:murphy_col}; we
write $\pi$ in this section to emphasise its role as the
prevalence parameter that skew transformations act upon.
Throughout this subsection, the conditional means are defined for classes
with both positive and negative observations: $n_j^+>0$, $n_j^->0$,
or equivalently $0<\pi_j<1$.

\begin{mydef}[Class-conditional calibration errors]
\label{def:cond_cal}
For class $j$, define the \emph{positive calibration error}
and \emph{negative calibration error} as
\begin{align}
  \alpha_{j}
    &= \frac{1}{n^{+}}\sum_{i:\,Y_{i,j}=1}(1 - \hat{Y}_{i,j})^{2}
     = \mathbb{E}\!\left[(1 - \hat{Y}_{i,j})^{2}\;\middle|\;Y_{i,j}=1\right],
  \label{eq:alpha}\\[4pt]
  \beta_{j}
    &= \frac{1}{n^{-}}\sum_{i:\,Y_{i,j}=0}\hat{Y}_{i,j}^{2}
     = \mathbb{E}\!\left[\hat{Y}_{i,j}^{2}\;\middle|\;Y_{i,j}=0\right].
  \label{eq:beta}
\end{align}
\end{mydef}

\noindent $\alpha_j$ measures how far predicted probabilities are
from 1 among the positive examples; a classifier that assigns
probability~$1$ to every positive example ($\hat{Y}_{i,j} = 1$ for all
$i$ with $Y_{i,j}=1$) has $\alpha_j = 0$. Symmetrically, $\beta_j$
measures how far predictions are from 0 among the negative
examples. Both are defined as conditional means and are therefore
independent of the class prevalence $\pi_j$.

In the binary case, $\alpha_j$ and $\beta_j$ are the Brier score
computed separately over the positive and over the negative examples
(Brier$^{+}$ and Brier$^{-}$), i.e., the two components of the
\emph{stratified Brier score} proposed by
\citet{wallace2012class,wallace2014improving}, who introduced it to expose
the systematic underestimation of minority-class probabilities that the
overall Brier score can mask under imbalance. They motivate the split by
analogy with decomposing accuracy into sensitivity and specificity; within
the present framework this analogy becomes exact: for hard predictions the
pair reduces to the class-conditional error rates of the Discrimination
Framework (Corollary~\ref{cor:hard_cal_skew}), it characterises the skew-invariant Brier-type class-indexed measures
that factor through $(\pi_j,\alpha_j,\beta_j)$ (Theorem~\ref{thm:cal_skew}(b)),
and it extends to $m$ classes and to multilabel data through the
aggregation operators under the per-class skew convention.

The following proposition decomposes $\mathrm{BS}_j^{\mathrm{class}}$
into $\alpha_j$ and $\beta_j$.

\begin{proposition}[Prevalence decomposition of per-class Brier]
\label{prop:bs_decomp}
For binary $Y_{i,j} \in \{0,1\}$:
\begin{equation}
  \mathrm{BS}_{j}^{\mathrm{class}}
  \;=\; \pi_{j}\,\alpha_{j} \;+\; (1-\pi_{j})\,\beta_{j}.
  \label{eq:bs_mix}
\end{equation}
\end{proposition}

\begin{proof}
Split the sum in $\mathrm{BS}_j^{\mathrm{class}} =
\tfrac{1}{n}\sum_i(Y_{i,j}-\hat{Y}_{i,j})^2$ by the value of
$Y_{i,j}$:
\begin{align*}
  \mathrm{BS}_{j}^{\mathrm{class}}
  &= \frac{1}{n}\!\!\sum_{i:\,Y_{i,j}=1}\!\!(1-\hat{Y}_{i,j})^{2}
   + \frac{1}{n}\!\!\sum_{i:\,Y_{i,j}=0}\!\!\hat{Y}_{i,j}^{2}
   = \frac{n^{+}}{n}\,\alpha_{j} + \frac{n^{-}}{n}\,\beta_{j}
   = \pi_{j}\,\alpha_{j} + (1-\pi_{j})\,\beta_{j}. \qedhere
\end{align*}
\end{proof}

\begin{example}[Running example: Brier collapse and the conditional errors]
\label{ex:running_brier}
Averaging the squared-deviation cells of Example~\ref{ex:running_cells} by columns gives
$\mathrm{BS}_j^{\mathrm{class}}=(21/80,\,13/40,\,9/40)$, and by rows
$\mathrm{BS}_i^{\mathrm{example}}=(13/150,\,53/300,\,59/100,\,23/100)$; both average to the micro Brier
score $13/48$ (Theorem~\ref{thm:brier_collapse}). The class-conditional errors are
$\alpha=(1/8,\,43/100,\,9/25)$ and $\beta=(2/5,\,1/100,\,9/50)$: the classifier is overconfident on the
negatives of $A$ and underconfident on the positives of $B$. For $A$, $\pi_A=1/2$ and
$\mathrm{BS}_A=\frac12\cdot\frac18+\frac12\cdot\frac25=\frac{21}{80}$ (Proposition~\ref{prop:bs_decomp}).
Two companion predictors separate calibration from ranking. Squaring every score preserves every row and
column order, so all ranking measures of Section~\ref{sec:ranking} are unchanged, yet the Brier score rises
to $14039/40000\approx0.351$. Conversely, the predictor with rows $(.8,.7,.2)$, $(.9,.4,.1)$, $(.4,.2,.3)$,
$(.2,.1,.6)$ has the same Brier score $13/48$ but macro-AUC $4/9$ instead of $13/18$.
\end{example}

\noindent Equation~(\ref{eq:bs_mix}) expresses $\mathrm{BS}_j$
as a convex mixture of $\alpha_j$ and $\beta_j$ with prevalence
weights. If $\alpha_j\neq\beta_j$, varying $\pi_j$ over $(0,1)$ makes
$\mathrm{BS}_j$ range over the open interval with endpoints $\alpha_j$
and $\beta_j$; if $\alpha_j=\beta_j$, the value is constant.
This mixture is the main source of skew-sensitivity. The identity itself is elementary (it
is the law of total expectation over the label) and is stated only to make
this mechanism explicit; the contribution of this section is the
characterisation below.

Even a perfectly calibrated forecaster has non-zero $\alpha_j$ and $\beta_j$:
if it issues the same prediction $p$ for every unit and $p$ equals the true base
rate, then $\alpha_j=(1-p)^2$ and $\beta_j=p^2$. The pair therefore
mixes calibration with sharpness, and is best read as a pair of class-conditional
squared errors.

\subsubsection*{Skew-invariance}

Following \citet{prati2026unified}, we define skew-invariance by asking
how a measure changes when the counts of positive and negative
examples are independently rescaled while their within-class
prediction distributions are held fixed.

\begin{mydef}[Skew-invariant calibration measure]
\label{def:cal_skew}
A calibration measure $M$ is \emph{skew-invariant} if, for any
class $j$ and any $\lambda, \mu > 0$, replacing $n^{+}_{j}$
positive examples with $\lambda n^{+}_{j}$ copies and $n^{-}_{j}$
negative examples with $\mu n^{-}_{j}$ copies (while keeping the
within-class distributions of $\hat{Y}_{i,j}$ unchanged) leaves
$M$ unchanged. It is \emph{skew-sensitive} otherwise.
\end{mydef}

\noindent Replicating whole examples also changes the columns
of the other classes. As in the base paper, skew-invariance is therefore understood
at the level of the per-class summaries: a measure is evaluated on the data
$\{(n_j^{+},n_j^{-},F_j^{+},F_j^{-})\}_{j=1}^{m}$, and the transformation changes
the two counts of one class, all the other quantities being held fixed. A factor
$\lambda$ or $\mu$ that is not an integer is read as the corresponding weighting of
the empirical distribution; for integer factors this is literal replication, and
every pair of data sets with the same $\{F_j^{\pm}\}$ is then connected by a chain of
such replications (pass through a common multiple of the counts). The theorem below
holds on this representation, and is proved in Lean in the following model: for one class, the
predicted scores of the positive and of the negative examples are two finite multisets, and the
transformation replicates them $k$- and $l$-fold; exemplar-indexed measures are outside it
(Remark~\ref{rem:cal_skew_exemplar_scope}).

\begin{theorem}[Characterisation of skew-invariant calibration
measures]
\label{thm:cal_skew}
Let $F_j^{+}$ and $F_j^{-}$ denote the within-class empirical
distributions of the predicted scores $\hat{Y}_{i,j}$ over the
positive and negative examples of class $j$, respectively.
Both statements use the per-class skew transformations of
Definition~\ref{def:cal_skew}, not arbitrary replications of entire
multilabel rows.
\begin{enumerate}
  \item[\textup{(a)}] A calibration measure $M$ is skew-invariant
    if and only if it can be written as a function of
    $\{F_j^{+}, F_j^{-}\}_{j=1}^{m}$ alone.
  \item[\textup{(b)}] Restrict attention to \emph{Brier-type class-indexed} measures: functions
    $g(\{\pi_j,\alpha_j,\beta_j\}_{j})$ of the class prevalences and of the squared-error
    class-conditional errors (the quantities through which the per-class values
    $\mathrm{BS}_j^{\mathrm{class}}$ depend on the data, by the mixture identity
    (\ref{eq:bs_mix})). Measures built from another deviation operator have their own conditional
    errors $\alpha_j^{\phi},\beta_j^{\phi}$ and obey the same statement with them. This
    includes $\mathrm{BS}_j^{\mathrm{class}}$ itself,
    $\mathrm{BS}^{\mathrm{macro}}$, and (by
    Theorem~\ref{thm:brier_collapse}) $\mathrm{BS}^{\mathrm{micro}}$. Within
    this class, $M$ is skew-invariant if and only if it is a function of
    $\{\alpha_j,\,\beta_j\}_{j=1}^{m}$ alone. (Exemplar-indexed measures such
    as $\mathrm{BS}_i^{\mathrm{example}}$ are outside this scope; see
    Remark~\ref{rem:cal_skew_exemplar_scope}.)
\end{enumerate}
\end{theorem}

\begin{proofsketch}
(a) A skew transformation changes only the counts $(n_j^+,n_j^-)$ and leaves every $F_j^{\pm}$ fixed, and
two data sets with the same $\{F_j^+,F_j^-\}$ are linked by a chain of such transformations; invariance
is therefore exactly dependence on $\{F_j^+,F_j^-\}$. (b) $\alpha_j$ and $\beta_j$ are means under
$F_j^{\pm}$; conversely, rescaling sweeps $\pi_j$ over $(0,1)$ with $(\alpha_j,\beta_j)$ fixed. The full
proof is in Appendix~\ref{app:proofs_cal}.
\end{proofsketch}

\begin{remark}[Scope of the characterisation]
\label{rem:cal_skew_scope}
Part (a) shows that the full invariant of the skew relation is
the pair of within-class score distributions, not the pair
$(\alpha_j, \beta_j)$: measures such as the within-class variance
of predicted scores are skew-invariant without being functions of
$(\alpha_j, \beta_j)$. Within the restricted class of Brier-type measures that factor
through the conditional mean squared errors, as specified in
Theorem~\ref{thm:cal_skew}(b), the pair $(\alpha_j,\beta_j)$
is sufficient to characterise skew-invariant measures. It is not
a complete invariant for arbitrary functionals of squared-error cells.
\end{remark}

\begin{remark}[Exemplar-indexed measures are outside part (b)'s scope]
\label{rem:cal_skew_exemplar_scope}
Part (b) concerns class-indexed measures only. Exemplar-indexed quantities such as
$\mathrm{BS}_i^{\mathrm{example}}$ (Equation~\ref{eq:bs_example}) or the row-level Murphy terms of
Definition~\ref{def:murphy_row} depend on how errors are distributed within each example's label vector,
which the per-class summaries $\{\pi_j,\alpha_j,\beta_j\}_j$ do not determine; moreover, a per-class skew
transformation duplicates or removes whole examples rather than acting on a fixed one. A skew-invariance
theory for exemplar-indexed measures is not developed here.
\end{remark}

\begin{corollary}[Skew-sensitivity of $\mathrm{BS}^{\mathrm{class}}_j$]
\label{cor:bs_skew}

Under the per-class skew transformations of
Definition~\ref{def:cal_skew}, $\mathrm{BS}_j^{\mathrm{class}}$
is skew-sensitive precisely when $\alpha_j\neq\beta_j$.
If this holds for at least one class, the class-averaged
$\mathrm{BS}^{\mathrm{macro}}$ is skew-sensitive, and so are the
numerically identical $\mathrm{BS}^{\mathrm{micro}}$ and
$\mathrm{BS}^{\mathrm{exemplar}}$ by Theorem~\ref{thm:brier_collapse}.
This statement concerns independent per-class changes of prevalence;
it does not assert sensitivity under every joint replication of entire
multilabel examples.

\end{corollary}

\begin{proof}
By Proposition~\ref{prop:bs_decomp}, $\mathrm{BS}_j = \pi_j
\alpha_j + (1-\pi_j)\beta_j$ depends explicitly on $\pi_j$, and
hence on the class counts $n^+_j$ and $n^-_j$. It is invariant
to their rescaling only if $\alpha_j = \beta_j$, i.e., the
classifier has identical calibration error on positive and
negative examples. Skew-sensitivity of $\mathrm{BS}^{\mathrm{macro}}$
follows since it is an unweighted average of the (skew-sensitive)
$\mathrm{BS}_j$; skew-sensitivity of $\mathrm{BS}^{\mathrm{micro}}$ and of
$\mathrm{BS}^{\mathrm{exemplar}}$ then follows from Theorem~\ref{thm:brier_collapse}, since both
equal $\mathrm{BS}^{\mathrm{macro}}$ as scalars for every dataset,
skewed or not.
\end{proof}

\begin{remark}[Class-weighted Brier scores]
\label{rem:class_weighted_brier}
A Brier score with class weights $w_1,w_0>0$ for positive and negative
examples,
$\frac1n\sum_i\{w_1Y_{i,j}(1-\hat{Y}_{i,j})^2+w_0(1-Y_{i,j})\hat{Y}_{i,j}^2\}$
(the form discussed by \citet{gerds2008performance}, as described by
\citet{zhu2025weighted}), equals $w_1\pi_j\alpha_j+w_0(1-\pi_j)\beta_j$ by the
argument of Proposition~\ref{prop:bs_decomp}. With fixed $w_1,w_0$ it depends
on $\pi_j$ and is skew-sensitive unless $w_1\alpha_j=w_0\beta_j$; with
$w_1=w_0=1$ it is $\mathrm{BS}_j^{\mathrm{class}}$ itself. By
Theorem~\ref{thm:cal_skew}(b), it is skew-invariant for every
$(\alpha_j,\beta_j)$ if and only if $w_1\pi_j$ and $w_0(1-\pi_j)$ are constant
in $\pi_j$, i.e.\ if and only if the weights are inverse-prevalence weights
$w_1=c_1/\pi_j$ and $w_0=c_0/(1-\pi_j)$, in which case the score is
$c_1\alpha_j+c_0\beta_j$. The choice $c_1=c_0=\tfrac12$ gives the unweighted
average $(\alpha_j+\beta_j)/2$ of the two stratified components.
Skew-invariance should not be confused with properness: these
prevalence-adjusted weights do not, in general, make the truthful
probability the minimiser of expected weighted squared loss.
\end{remark}

The scalar Brier score is therefore skew-sensitive. The next corollary relates
$(\alpha_j,\beta_j)$ to the skew-invariant quantities already established for the
Discrimination Framework; the connection is exact in the
hard-prediction limit.

\begin{corollary}[Connection to the Discrimination Framework for hard predictions]
\label{cor:hard_cal_skew}
For binary hard predictions $\hat{Y}_{i,j} \in \{0,1\}$:
\begin{align*}
  \alpha_{j} &= 1 - \mathrm{recall}_{j},\\
  \beta_{j}  &= 1 - \mathrm{specificity}_{j}.
\end{align*}
Consequently, $\mathrm{BS}_j^{\mathrm{class}} = \pi_j(1 -
\mathrm{recall}_j) + (1-\pi_j)(1-\mathrm{specificity}_j)$, which
is the prevalence-weighted error rate, and the skew-invariant
calibration measures $(\alpha_j, \beta_j)$ are exactly the
class-conditional error rates of the Discrimination Framework.
\end{corollary}

\begin{proof}
For binary $\hat{Y}_{i,j} \in \{0,1\}$, $(1-\hat{Y}_{i,j})^2 =
1-\hat{Y}_{i,j}$, so
$\alpha_j = \mathbb{E}[1-\hat{Y}_{i,j}\mid Y_{i,j}=1] =
1 - \mathbb{E}[\hat{Y}_{i,j}\mid Y_{i,j}=1] = 1 - \mathrm{recall}_j$.
Similarly, $\hat{Y}_{i,j}^2 = \hat{Y}_{i,j}$ for binary values,
so $\beta_j = \mathbb{E}[\hat{Y}_{i,j}\mid Y_{i,j}=0] =
\mathrm{FPR}_j = 1 - \mathrm{specificity}_j$.
\end{proof}

Given the skew-sensitivity of $\mathrm{BS}_j$ itself (Corollary~\ref{cor:bs_skew}), the
following corollary examines, term by term, whether the individual terms of the Murphy
decomposition inherit the same prevalence dependence.

\begin{corollary}[Skew-sensitivity of Murphy terms]
\label{cor:murphy_skew}
For binary $Y_{i,j} \in \{0,1\}$, of the three Murphy terms:
\begin{enumerate}
  \item[\textup{(a)}]
    $\mathrm{UNC}_j = \pi_j(1-\pi_j)$ is skew-sensitive:
    it depends only on $\pi_j$ and carries no information about
    the classifier.
  \item[\textup{(b)}]
    
$\mathrm{RES}_j$ is skew-sensitive if and only if the
    class-conditional score distributions differ, $F_j^+\neq F_j^-$.
    If $F_j^+=F_j^-$, then $\mathrm{RES}_j=0$ at every prevalence
    and is skew-invariant. In the other case, $\mathrm{RES}_j$
    varies with $\pi_j$ according to the Bayes formula of the proof.

  \item[\textup{(c)}]
    $\mathrm{REL}_j = \mathbb{E}[(\hat{Y}_{i,j} -
    f_j(\hat{Y}_{i,j}))^2]$ is skew-sensitive in general, because
    $f_j$ changes with $\pi_j$. Perfect calibration
    ($\mathrm{REL}_j = 0$) is \emph{not} preserved either: it holds at two
    different prevalences if and only if every predicted value is attained by
    one class alone and equals $1$ (positives only) or $0$ (negatives only). In
    particular, if some predicted value is attained by both positive and negative
    examples, then $\mathrm{REL}_j=0$ at one prevalence implies $\mathrm{REL}_j>0$
    at every other.
\end{enumerate}
The class-conditional pair $(\alpha_j,\beta_j)$ is invariant
under the per-class skew transformation, whereas the individual Murphy
terms are not invariant in general. Degenerate exceptions exist: $\mathrm{RES}_j\equiv0$ when
$F_j^+=F_j^-$ (part (b)), and, when every predicted value is attained by one class alone,
$\mathrm{REL}_j=\pi_j\alpha_j+(1-\pi_j)\beta_j$, which is invariant if $\alpha_j=\beta_j$.
\end{corollary}

\begin{proofsketch}
(a) is immediate. (b) By Bayes' theorem,
$\mathrm{RES}_j=\pi_j^2(1-\pi_j)^2\sum_p(a_p-b_p)^2/(\pi_ja_p+(1-\pi_j)b_p)$, where $a_p$ and $b_p$ are
the probabilities of the predicted value $p$ under $F_j^+$ and $F_j^-$; it vanishes identically iff
$F_j^+=F_j^-$, and is otherwise positive yet bounded by $\pi_j(1-\pi_j)$. (c) For a value attained by both
classes, the calibration map is strictly increasing in the replication factor. The full proof, with the
concrete instance $\mathrm{REL}_j=0\to1/45$, is in Appendix~\ref{app:proofs_cal}.
\end{proofsketch}

\begin{remark}[Practical consequences]
\label{rem:cal_skew_practical}
Corollary~\ref{cor:bs_skew} implies that
reporting $\mathrm{BS}^{\mathrm{macro}}$ on class-imbalanced
multilabel data mixes calibration quality with class prevalence
information. A rare label with $\pi_j \approx 0.01$ can have a much lower Brier
score than a label with $\pi_j \approx 0.5$, whose
$\mathrm{UNC}_j=\pi_j(1-\pi_j)$ is $0.25$ against $0.0099$, even if both are equally
miscalibrated. Practitioners requiring a prevalence-independent
calibration summary should report $(\alpha_j, \beta_j)$ or their
aggregate $\alpha_j + \beta_j$. Corollary~\ref{cor:hard_cal_skew}
shows that, in the hard-prediction limit, these become the
complements $(1-\mathrm{recall}_j,1-\mathrm{specificity}_j)$ of the
skew-invariant recall/specificity rates identified by
\citet{prati2026unified}. Thus the two frameworks share the same
class-conditional invariant structure, with the appropriate error-rate orientation.
\end{remark}

% ============================================================

% ============================================================
%  Section 5 — Ranking Framework
% ============================================================
% ============================================================
%  Section 5: Ranking Framework: Ranking via Discount Operators
%  Draft for: "A Generalised Confusion Matrix Framework"
%  Author: Ronaldo C. Prati
% ============================================================

\section{The Ranking Framework: Positional Operators}
\label{sec:ranking}

The first two frameworks evaluate a classifier's ability to
identify which labels are correct (discrimination) and how
accurately its confidence values match empirical frequencies
(calibration). Both rest on \emph{pointwise} cell operators:
the cell value $D(Y_{i,j}, \hat{Y}_{i,j})$ depends only on
the pair at position $(i,j)$. Ranking evaluation breaks this
assumption in two orthogonal ways. First: how useful is it to
predict label $j$ at rank~1 for example $i$ vs.\ rank~5? The
answer depends on the full \emph{row} $\hat{\mathbold{Y}}_{i,\cdot}$,
not just on $\hat{Y}_{i,j}$, which gives rise to \emph{row-dependent}
positional operators and the nDCG family. Second: how highly does
the classifier rank example $i$ among all examples for class
$j$? This depends on the full \emph{column}
$\hat{\mathbold{Y}}_{\cdot,j}$, giving rise to
\emph{column-dependent} positional operators and the AUC family.
This section develops both families within a unified operator
framework, establishes the duality between them, and shows that
the two families share the same structural properties (collapse
behaviour, skew-invariance, and failure of strict properness),
derived from the same matrix machinery.

% ============================================================
\subsection{Discount Cell Operators and the DCG Matrix}
\label{sec:disc_op}
% ============================================================

\begin{mydef}[Rank function]
\label{def:rank}
For example $i$ and a predicted score vector
$\hat{\mathbold{Y}}_{i,\cdot} \in [0,1]^{m}$, the rank of
label $j$ is
\begin{equation}
  \mathrm{rank}_{i}(j)
  \;=\;
  \bigl|\{j' : \hat{Y}_{i,j'} > \hat{Y}_{i,j}\}\bigr| + 1,
  \label{eq:rank}
\end{equation}
i.e.\ the position of label $j$ in the descending ordering of
$\hat{\mathbold{Y}}_{i,\cdot}$. Under this convention tied scores
share the same rank, namely the smallest position among them
(min-rank, or ``competition'', ranking). Labels predicted most
confidently receive rank~1.
\end{mydef}

\begin{mydef}[Positional operator and DCG cell matrix]
\label{def:disc_op}
A \emph{positional weight function} is a non-increasing map
$w: \mathbb{N} \to \mathbb{R}_{\geq 0}$ with $w(1) > 0$.
The \emph{positional cell operator} with weight $w$ is
\begin{equation}
  D_{w}\bigl(Y_{i,j},\, \hat{\mathbold{Y}}_{i,\cdot}\bigr)
  \;=\;
  Y_{i,j} \cdot w\!\bigl(\mathrm{rank}_{i}(j)\bigr).
  \label{eq:disc_op}
\end{equation}
The \emph{DCG cell matrix} $\mathbold{D}_{w} \in \mathbb{R}^{n\times m}$
is defined entry-wise by
$[\mathbold{D}_{w}]_{i,j} = D_{w}(Y_{i,j}, \hat{\mathbold{Y}}_{i,\cdot})$.
\end{mydef}

\noindent Two canonical weight functions define the families
developed in this section, both fixed a priori. For the
\emph{row-dependent} case (nDCG), $w(r) = 1/\log_{2}(r+1)$:
labels ranked first receive full credit $w(1)=1$, with credit
decaying logarithmically so that $w(2)\approx 0.63$ and
$w(r)\to 0$ as $r\to\infty$. For the \emph{column-dependent}
case, the fixed rank function can instead act as a cost: the AUC construction uses the increasing rank cost $v(r)=r$ to obtain the mean rank of positive examples, followed by a decreasing affine transformation
(Section~\ref{sec:col_dep}, Proposition~\ref{prop:auc_meanrank}).
In both cases, cells with
$Y_{i,j} = 0$ contribute zero regardless of rank: the
positional weight rewards relevant items ranked high and gives a cell
nothing for an irrelevant label, so a false alarm (an irrelevant label ranked
high) is penalised only indirectly, by pushing relevant labels down the
ranking.

The positional operator is \emph{row-dependent}: changing
$\hat{Y}_{i,j'}$ for any $j' \neq j$ can alter
$\mathrm{rank}_i(j)$ and therefore $[\mathbold{D}_w]_{i,j}$.
This violates the pointwise structure of
Definition~\ref{def:cell_op} and has structural consequences
developed in Sections~\ref{sec:dcg_family}
and~\ref{sec:ranking_collapse}.

% ============================================================
\subsection{The DCG Family via the Three Aggregation Operators}
\label{sec:dcg_family}
% ============================================================

Applying the three aggregation operators $\Upsigma_{\mathrm{1}}$,
$\Upsigma_{\mathrm{m}}$, $\Upsigma_{\mathrm{n}}$ to
$\mathbold{D}_{w}$ yields three families of measures; they are
summarised in Table~\ref{tab:dcg_family}.

\begin{mydef}[Raw DCG family]
\label{def:dcg_family}
For the DCG cell matrix $\mathbold{D}_{w}$ with weight $w$:
\begin{align}
  \mathrm{DCG}^{\mathrm{micro}}
    &= \frac{1}{nm}\,\Upsigma_{\mathrm{1}}(\mathbold{D}_{w})
     = \frac{1}{nm}\sum_{i=1}^{n}\sum_{j=1}^{m}
       Y_{i,j}\, w\!\bigl(\mathrm{rank}_{i}(j)\bigr),
  \label{eq:dcg_micro}\\[4pt]
  \mathrm{DCG}_{j}^{\mathrm{class}}
    &= \frac{1}{n}\bigl[\Upsigma_{\mathrm{m}}(\mathbold{D}_{w})\bigr]_{j}
     = \frac{1}{n}\sum_{i=1}^{n}
       Y_{i,j}\, w\!\bigl(\mathrm{rank}_{i}(j)\bigr),
  \label{eq:dcg_class}\\[4pt]
  \mathrm{DCG}_{i}^{\mathrm{ex}}
    &= \frac{1}{m}\bigl[\Upsigma_{\mathrm{n}}(\mathbold{D}_{w})\bigr]_{i}
     = \frac{1}{m}\sum_{j=1}^{m}
       Y_{i,j}\, w\!\bigl(\mathrm{rank}_{i}(j)\bigr).
  \label{eq:dcg_ex}
\end{align}
\end{mydef}

\noindent $\mathrm{DCG}_j^{\mathrm{class}}$ is a per-label
ranking measure novel to this framework: it asks how highly
the classifier ranks label $j$ across all examples in which
$j$ is relevant. $\mathrm{DCG}_i^{\mathrm{ex}}$ is the standard per-example DCG of the IR literature scaled by $1/m$
\citep{jarvelin2002cumulated}, here derived as the row-wise
aggregation of the cell matrix.

\begin{mydef}[Ideal DCG and normalisation]
\label{def:idcg}
The \emph{ideal DCG} for example $i$ is achieved when all
$k_{i} = \Upsigma_{\mathrm{1}}(\mathbold{Y}_{i,\cdot})$
relevant labels are ranked first:
\begin{equation}
  \mathrm{IDCG}_{i} \;=\; \sum_{r=1}^{k_{i}} w(r),
  \label{eq:idcg_ex}
\end{equation}
and for class $j$ the ideal is achieved when every positive
example ranks label $j$ at position~1:
\begin{equation}
  \mathrm{IDCG}_{j}^{\mathrm{class}}
  \;=\; \pi_{j} \cdot w(1),
  \label{eq:idcg_class}
\end{equation}
where $\pi_j = n_j^+ / n$ with $n_j^+ = \sum_i Y_{ij}$.
The ratios below are defined whenever their denominators are positive: $k_i\geq 1$ for $\mathrm{nDCG}_i$, $n_j^+\geq 1$ (equivalently $\pi_j>0$) for $\mathrm{nDCG}_j^{\mathrm{class}}$, and $\sum_i\mathrm{IDCG}_i>0$ for micro-nDCG. Zero-ideal instances or classes require an explicit exclusion or zero-score convention rather than an unqualified division.
The \emph{normalised measures} are:
\begin{align}
  \mathrm{nDCG}_{i}
    &= \frac{\mathrm{DCG}_{i}^{\mathrm{ex}}}
            {\mathrm{IDCG}_{i}/m}
     = \frac{\sum_{j}Y_{i,j}w(\mathrm{rank}_{i}(j))}
            {\mathrm{IDCG}_{i}},
  \label{eq:ndcg_ex}\\[4pt]
  \mathrm{nDCG}_{j}^{\mathrm{class}}
    &= \frac{\mathrm{DCG}_{j}^{\mathrm{class}}}
            {\mathrm{IDCG}_{j}^{\mathrm{class}}}
     = \frac{1}{\pi_{j}\, w(1)}
       \cdot \frac{1}{n}\sum_{i} Y_{i,j}\,
       w\!\bigl(\mathrm{rank}_{i}(j)\bigr).
  \label{eq:ndcg_class}
\end{align}
The \emph{micro-nDCG} normalises the global DCG by the global
ideal:
\begin{equation}
  \mathrm{nDCG}^{\mathrm{micro}}
  = \frac{\mathrm{DCG}^{\mathrm{micro}}}
         {\overline{\mathrm{IDCG}}}
  = \frac{\sum_{i,j} Y_{i,j}w(\mathrm{rank}_{i}(j))}
         {\sum_{i}\mathrm{IDCG}_{i}},
  \label{eq:ndcg_micro}
\end{equation}
where $\overline{\mathrm{IDCG}} = \frac{1}{nm}\sum_i
\mathrm{IDCG}_i$ is the average per-example ideal DCG.
\end{mydef}

\begin{remark}[Ties and the range of nDCG]
\label{rem:ties}
Under the min-rank convention of Definition~\ref{def:rank}, tied
scores share the smallest position among them, so a degenerate
predictor that assigns identical scores to all labels places every
label at rank~1 and collects the full weight $w(1)$ on each
positive cell. The realised DCG can then exceed
$\mathrm{IDCG}_i$, which is computed for a strict ordering in
which the $k_i$ positives occupy positions $1, \ldots, k_i$, and
$\mathrm{nDCG}_i$ can exceed~1. Throughout we therefore assume
that the predicted scores within each row (and, for the AUC
family of Section~\ref{sec:col_dep}, within each column) are
distinct (for example, almost surely under a joint score distribution assigning probability zero to exact ties), so that $\mathrm{nDCG} \in [0,1]$ for all
aggregations. When exact ties do occur in practice, they can be
resolved by any fixed tie-breaking permutation applied before
ranks are computed, which restores the standard range.
\end{remark}

\begin{table}[!t]
\centering
\small
\setlength{\tabcolsep}{5pt}
\begin{tabular}{l|lll}
\hline
\textbf{Aggregation} &
  \textbf{Raw DCG} &
  \textbf{Normalisation} &
  \textbf{Normalised nDCG} \\
\hline
$\Upsigma_{\mathrm{1}}$ (micro) &
  $\mathrm{DCG}^{\mathrm{micro}}$ &
  $\overline{\mathrm{IDCG}}$ &
  $\mathrm{nDCG}^{\mathrm{micro}}$ \\
$\Upsigma_{\mathrm{m}}$ (class) &
  $\mathrm{DCG}_{j}^{\mathrm{class}}$ &
  $\pi_{j}\,w(1)$ &
  $\mathrm{nDCG}_{j}^{\mathrm{class}}$ (novel) \\
$\Upsigma_{\mathrm{n}}$ (exemplar) &
  $\mathrm{DCG}_{i}^{\mathrm{ex}}$ &
  $\mathrm{IDCG}_{i}/m$ &
  $\mathrm{nDCG}_{i}$ (standard) \\
\hline
\end{tabular}
\caption{DCG family via the three aggregation operators.
``Novel'' indicates measures not previously expressed within
the matrix framework. The standard nDCG of the IR literature
\citep{jarvelin2002cumulated} corresponds to the exemplar axis.}
\label{tab:dcg_family}
\end{table}

% ============================================================
\subsection{Collapse for Raw DCG, Non-Collapse for nDCG}
\label{sec:ranking_collapse}
% ============================================================

The main structural difference between the Ranking Framework and
the Discrimination and Calibration Frameworks lies in the behaviour of micro- and
macro-averaging.

\begin{theorem}[Raw DCG collapses; nDCG does not]
\label{thm:dcg_collapse}
Assume that the ideal denominators appearing below are positive. Put
$D_{ij}=Y_{ij}w(\mathrm{rank}_i(j))$,
$S_i=\sum_jD_{ij}$, $b_i=\mathrm{IDCG}_i$,
$r_i=S_i/b_i=\mathrm{nDCG}_i$,
$T_j=\sum_iD_{ij}$, $N_j=n_j^+w(1)$ and $B=\sum_i b_i$.
\begin{enumerate}
\item[\textup{(a)}] The three raw DCG
aggregations are equal as scalars:
\begin{equation}
\mathrm{DCG}^{\mathrm{micro}}
=\frac1m\sum_j\mathrm{DCG}_j^{\mathrm{class}}
=\frac1n\sum_i\mathrm{DCG}_i^{\mathrm{ex}}.
\label{eq:dcg_collapse}
\end{equation}

\item[\textup{(b)}] The micro-nDCG, a global ratio, is a $b_i$-weighted mean of the $r_i$, whereas the
exemplar mean of nDCG is unweighted. They coincide if and only if
\begin{equation}
\widehat{\mathrm{Cov}}(\mathrm{IDCG}_i,\mathrm{nDCG}_i)=0.
\label{eq:ndcg_cov}
\end{equation}
In particular, equality holds when all $k_i$ are equal.
\item[\textup{(c)}] The micro-nDCG and the class mean of nDCG coincide if and only if
\begin{equation}
\sum_j T_j\left(\frac1B-\frac1{mN_j}\right)=0.
\label{eq:ndcg_class_micro}
\end{equation}
For arbitrary class-cell sums $T_j$, this is an identity if and only if
$mN_j=B$ for every $j$. These coefficients cannot all vanish when
some example has at least two relevant labels and $w(2)<w(1)$.
\item[\textup{(d)}] The class mean and the exemplar mean of nDCG coincide if and only if
\begin{equation}
\sum_{i,j}D_{ij}\left(\frac1{mN_j}-\frac1{nb_i}\right)=0.
\label{eq:ndcg_class_ex}
\end{equation}
For arbitrary cell values supported on the relevant positions
($D_{ij}=0$ when $Y_{ij}=0$), this equality holds identically
if and only if $mN_j=nb_i$ at every relevant cell. These coefficient
equalities cannot all hold when some example has at least two relevant
labels and $w(2)<w(1)$, assuming every class and every example has a
relevant entry.

\end{enumerate}

Consequently, the three normalised summaries
\begin{equation}
\mathrm{nDCG}^{\mathrm{micro}},\qquad
\frac1m\sum_j\mathrm{nDCG}_j^{\mathrm{class}},\qquad
\frac1n\sum_i\mathrm{nDCG}_i
\label{eq:ndcg_no_collapse}
\end{equation}
\emph{need not coincide}; their equality on a particular data set is
governed by (b)--(d), not ruled out universally.

\end{theorem}

\begin{proofsketch}
(a) is finite-sum rearrangement, as for the Brier score (Theorem~\ref{thm:brier_collapse}). Parts (b)--(d)
compare weighted and unweighted means of the same cells: (b) is the covariance identity for weighted means,
and (c)--(d) reduce equality for all freely chosen cells to coefficient conditions that contradict
$\sum_i\mathrm{IDCG}_i<w(1)\sum_ik_i$ when some $k_i\ge2$ and $w(2)<w(1)$. The full proof is in
Appendix~\ref{app:proofs_rank}.
\end{proofsketch}

\needspace{8\baselineskip}

\begin{remark}[Comparison with the Discrimination and Calibration Frameworks]
\label{rem:collapse_comparison}
The Discrimination Framework (t-norm operators) does not collapse: micro-$F_1
\neq$ macro-$F_1$ in general, because the denominator of each
per-class ratio depends on class-specific TP, FP, FN counts.
The Calibration Framework (squared-error) collapses completely: all three
scalar Brier scores are identical (Theorem~\ref{thm:brier_collapse}),
because the same per-cell squared losses are summed in either order and the resulting total is normalised by the common factor $nm$.
The Ranking Framework is split: raw DCG collapses (like
the Calibration Framework), but normalised nDCG does not (like the Discrimination Framework).
The source of non-collapse is the normalisation, which
introduces per-example ($\mathrm{IDCG}_i$) or per-class
($\pi_j w(1)$) denominators that differ across instances.
\end{remark}

% ============================================================
\subsection{Per-Class nDCG as a Novel Measure}
\label{sec:per_class_ndcg}
% ============================================================

The class-level measure $\mathrm{nDCG}_j^{\mathrm{class}}$
(Definition~\ref{def:idcg}, equation~(\ref{eq:ndcg_class}))
does not appear in the standard IR literature, which focuses on
the exemplar axis. It has the following interpretation.

\begin{proposition}[Per-class nDCG as conditional mean]
\label{prop:ndcg_class_mean}
Let $I$ be uniformly distributed over the $n$ observed examples, and assume $n_j^+>0$. The expectation below is the resulting empirical conditional average.
\begin{equation}
  \mathrm{nDCG}_{j}^{\mathrm{class}}
  \;=\; \frac{1}{w(1)}\,
        \mathbb{E}\!\left[w\!\bigl(\mathrm{rank}_{i}(j)\bigr)
        \;\middle|\; Y_{i,j}=1\right].
  \label{eq:ndcg_class_cond}
\end{equation}
\end{proposition}

\begin{proof}
From (\ref{eq:ndcg_class}):
$\mathrm{nDCG}_j^{\mathrm{class}}
 = \frac{1}{\pi_j w(1)}\cdot\frac{1}{n}\sum_i Y_{ij}
   w(\mathrm{rank}_i(j))
 = \frac{1}{w(1)}\cdot\frac{1}{n_j^+}\sum_{i:Y_{ij}=1}
   w(\mathrm{rank}_i(j))
 = \frac{1}{w(1)}\,\mathbb{E}[w(\mathrm{rank}_i(j))\mid Y_{ij}=1]$.
\end{proof}

\noindent The measure asks: on average, how highly does the
classifier rank label $j$ for examples where $j$ is truly
relevant? A classifier that always ranks label $j$
first among its relevant examples achieves
$\mathrm{nDCG}_j^{\mathrm{class}} = w(1)/w(1) = 1$, but this cannot hold for all labels at once
when examples have several relevant labels, since only one label per example is first under a strict (tie-broken) ordering: with the strict-order oracle used in the experiments (relevant labels of an example ordered uniformly at random, averaged over $20$ draws), the illustrative macro values are $0.81$ on Emotions and $0.61$ on Yeast (Section~\ref{sec:experiments}). The values depend on the specified tie-breaking convention. In contrast, if all truly relevant labels receive the same highest score and irrelevant labels are scored lower, the min-rank convention assigns every relevant label rank~$1$, giving per-class nDCG $1$ for each represented class; this is not the strict-order oracle used for the reported values. A
uniformly random ordering of the $m$ labels gives
$\frac1m\sum_{r=1}^{m}w(r)/w(1)$, which for the nDCG weight is $0.55$ for $m=6$ and $0.40$ for
$m=14$ and decreases slowly with $m$ ($0.12$ for $m=1000$). The measure is
analogous to recall
($\mathbb{E}[\hat{Y}_{ij}\mid Y_{ij}=1]$ for binary
predictions) in the Discrimination Framework.

\citet{prati2026unified} establish (in their proposition on ordinal AUC as a column-wise aggregation) that the ordinal
AUC across multiple classes can be expressed as a
column-wise aggregation of per-class binary AUCs. An
analogous structure holds here:
$\mathrm{DCG}^{\mathrm{micro}} = \frac{1}{m}\sum_j
\mathrm{DCG}_j^{\mathrm{class}}$ (Theorem~\ref{thm:dcg_collapse}(a))
expresses the global DCG as the unweighted mean of per-class
DCG values, another column-wise decomposition. This is not an assertion that micro-AUC equals macro-AUC: the latter are generally different because micro-AUC weights the per-class values by their positive-example counts.

% ============================================================
\subsection{Connection to the Discrimination Framework for Hard Predictions}
\label{sec:disc_hard}
% ============================================================

For binary hard predictions $\hat{Y}_{ij} \in \{0,1\}$, let
$k_i^{+}=\sum_j\hat{Y}_{i,j}$ be the number of predicted positives.
Under the min-rank convention of Definition~\ref{def:rank}, ties are \emph{not} broken:
\begin{equation}
  \mathrm{rank}_i(j)
  = \bigl|\{j': \hat{Y}_{i,j'} > \hat{Y}_{i,j}\}\bigr| + 1
  = \begin{cases}
      1 & \text{if }\hat{Y}_{i,j}=1,\\
      k_i^{+}+1 & \text{if }\hat{Y}_{i,j}=0.
    \end{cases}
  \label{eq:hard_rank}
\end{equation}
A different, explicitly tie-broken ordering can instead put the predicted positives at positions $r_{i,j}\in\{1,\ldots,k_i^+\}$ and the predicted negatives at positions $k_i^++s_{i,j}$, where $s_{i,j}\in\{1,\ldots,m-k_i^+\}$. These are tie-broken positions, not the min-ranks in Equation~\eqref{eq:hard_rank}; all statements below specify which convention they use.

\begin{proposition}[Hard-prediction limit]
\label{prop:hard_ranking}
For binary hard predictions $\hat{Y}_{ij} \in \{0,1\}$ with exactly one
predicted label per example ($k_i^{+} = 1$), let
$j^* = \arg\max_j \hat{Y}_{i,j}$ and let $k_i = \sum_j Y_{i,j}$ be the number of
relevant labels of example~$i$. Under the min-rank convention of
Definition~\ref{def:rank}, $\mathrm{rank}_i(j^*) = 1$ and every other label has
rank~$2$, so
\begin{equation}
  \mathrm{DCG}_i^{\mathrm{ex}}
  \;=\; \frac{Y_{i,j^{*}}\,w(1)}{m}
  \;+\;
  \underbrace{\frac{w(2)}{m}\,\bigl(k_i - Y_{i,j^*}\bigr)}_{R_i \;\geq\; 0}.
  \label{eq:hard_ranking_general}
\end{equation}
The correction term $R_i$ is the credit earned by the relevant labels
other than $j^*$. In particular, a wrong top prediction does not earn
zero: it earns $w(2)\,k_i/m$, which vanishes only if $w(2)=0$. (Under another
tie-breaking among the predicted negatives their ranks lie in
$\{2,\ldots,m\}$, and $R_i$ is positive for a strictly positive weight if and only if another relevant
label exists, $k_i>Y_{i,j^*}$.)
Consequently:
\begin{enumerate}
  \item[\textup{(a)}] In the multiclass case (one-hot $\mathbold{Y}$),
    $k_i = 1$, and the one-hot constraint forces $Y_{i,j}=0$ for every $j$
    other than the \emph{true} class $c_i$, which need not be $j^*$. If the
    prediction is correct, $R_i = 0$ and $\mathrm{DCG}_i^{\mathrm{ex}} = w(1)/m$;
    if it is wrong, $Y_{i,j^*}=0$ and $\mathrm{DCG}_i^{\mathrm{ex}} = w(2)/m$.
    The exemplar average is therefore
    \[
      \frac{1}{n}\sum_i \mathrm{DCG}_i^{\mathrm{ex}}
      \;=\;
      \frac{\mathrm{Precision}@1 \cdot w(1)
            + \bigl(1-\mathrm{Precision}@1\bigr)\cdot w(2)}{m},
    \]
    an affine function of $\mathrm{Precision}@1$ that reduces to
    $\mathrm{Precision}@1\cdot w(1)/m$ if and only if $w(2)=0$ (or
    $\mathrm{Precision}@1=1$). For the canonical weight $w(r)=1/\log_2(r+1)$,
    $w(2)\approx 0.63\,w(1)$, so wrong top predictions retain a large share of
    the credit.
  \item[\textup{(b)}] In the general multilabel case, since $R_i \geq 0$,
    $\mathrm{DCG}_i^{\mathrm{ex}} \geq Y_{i,j^*}\,w(1)/m$, hence
    \[
      \frac{1}{n}\sum_i \mathrm{DCG}_i^{\mathrm{ex}}
      \;\geq\;
      \mathrm{Precision}@1 \cdot w(1) / m,
    \]
    with equality iff $R_i = 0$ for every example, i.e.\ (for $w(2)>0$ and
    $k_i\ge 1$) iff every example's top-predicted label is its only relevant
    one. Precision@1 is therefore a lower bound on exemplar DCG, not an exact
    reduction.
\end{enumerate}
\end{proposition}

\begin{proof}
For $k_i^+=1$ the single predicted label $j^*$ has score~$1$ and every other
label score~$0$, so under the min-rank convention $\mathrm{rank}_i(j^*)=1$ and
$\mathrm{rank}_i(j)=2$ for $j\neq j^*$. Splitting the sum defining
$\mathrm{DCG}_i^{\mathrm{ex}}$ (Equation~\ref{eq:dcg_ex}) into the $j=j^*$ term
and the remaining terms gives~(\ref{eq:hard_ranking_general}), with
$R_i = \frac{w(2)}{m}\sum_{j\neq j^*}Y_{i,j} = \frac{w(2)}{m}(k_i-Y_{i,j^*})
\geq 0$ because $w\geq 0$. Part (a): for one-hot $\mathbold{Y}$ with true class
$c_i$ we have $k_i=1$ and $Y_{i,j^*}=\mathbf{1}[c_i=j^*]$, so
$\mathrm{DCG}_i^{\mathrm{ex}} = \bigl(w(1)\,\mathbf{1}[c_i=j^*] +
w(2)\,\mathbf{1}[c_i\neq j^*]\bigr)/m$; averaging over $i$, with
$\mathrm{Precision}@1 = \frac1n\sum_i\mathbf{1}[c_i=j^*_i]$, gives the stated
formula, whose difference from $\mathrm{Precision}@1\cdot w(1)/m$ is
$(1-\mathrm{Precision}@1)\,w(2)/m$. Part (b) follows from $R_i\geq 0$ by
averaging over~$i$.
\end{proof}

\begin{remark}[General $k_i^{+}$, and the role of the weight's support]
\label{rem:hard_ranking_general_k}
Part (b) generalises to $k_i^{+}=k$ predicted positives $j_1,\ldots,j_k$
(with a tie-breaking convention that assigns the predicted positives positions $1,\ldots,k$ and the predicted negatives positions $k+1,\ldots,m$, as distinguished from min-rank following Equation~\eqref{eq:hard_rank}): splitting
as before,
\[
  \mathrm{DCG}_i^{\mathrm{ex}}
  \;=\; \frac{1}{m}\sum_{l=1}^{k} Y_{i,j_l}\,w(l)
  \;+\;
  \underbrace{\frac{1}{m}\!\!\sum_{j\notin\{j_1,\ldots,j_k\}}\!\! Y_{i,j}\,
    w\!\bigl(\mathrm{rank}_i(j)\bigr)}_{R_i \;\geq\; 0}.
\]
For the canonical nDCG weight $w(r)=1/\log_2(r+1)$, which is strictly positive
at every rank, $R_i>0$ whenever a relevant label falls outside the top-$k$
predictions, so only the inequality of part (b) survives in general, in the
multiclass case as well (Proposition~\ref{prop:hard_ranking}(a): under min-rank a wrong top
prediction still earns $w(2)/m$; under the strict tie-break used here it earns $w(1+s)/m$, where
$s\in\{1,\ldots,m-1\}$ is the position of the true label among the predicted negatives). If $w(r)=0$ for every $r>k$, the residual term $R_i$ vanishes, but the remaining top-$k$ sum is generally \emph{weighted}:
\[
\mathrm{DCG}_i^{\mathrm{ex}} = \frac{1}{m}\sum_{l=1}^{k}Y_{i,j_l}w(l).
\]
The unweighted standard Precision@$k$ identity requires the indicator discount $w(r)=\mathbf{1}[r\le k]$ (under the stated strict top-$k$ tie-breaking convention), for which
\[
\mathrm{DCG}_i^{\mathrm{ex}} = \frac{1}{m}\sum_{l=1}^{k}Y_{i,j_l}
= \frac{k}{m}\,\mathrm{Precision}_i@k.
\]
If instead $w(r)=c$ for $1\le r\le k$ and $w(r)=0$ for $r>k$, the right-hand side is $(ck/m)\,\mathrm{Precision}_i@k$. For instance, at $m=3$, $k=2$, $w=(1,1/2,0)$, with only the second-ranked label relevant, the weighted DCG is $1/6$ whereas $(k/m)\mathrm{Precision}_i@k=1/3$. Thus compact support removes the tail, while constant top-$k$ weights are needed for the usual unweighted identity; the multiclass assumption alone does not imply it.
\end{remark}

\noindent More generally, for soft predictions, a rank-truncated positional operator with $w_k(r)=w(r)\,\mathbf{1}[r\leq k]$ yields a rank-weighted Precision@$k$ quantity; the untruncated DCG with a strictly positive tail does not equal a top-$k$ metric. This connects
the Ranking Framework to the standard precision--recall-at-$k$ family
used in IR evaluation \citep{manning2008introduction}.

\needspace{7\baselineskip}

% ============================================================
\subsection{Positional Operators Are Not Strictly Proper}
\label{sec:ranking_proper}
% ============================================================

\begin{proposition}[Positional operators are not strictly proper]
\label{prop:disc_improper}
For $m \geq 2$, no positional operator $D_w$ with non-increasing
$w$ induces a strictly proper scoring rule. Among reports with pairwise
distinct entries (the standing assumption of Remark~\ref{rem:ties}),
truthful reporting, with distinct $p_{i,j}$, attains the maximum expected
score (the rule is proper in the weak sense), but it is never the unique
maximiser. Without the distinctness restriction, weak properness need not hold:
under the min-rank convention a constant report puts every label at rank~$1$ and earns
$w(1)\sum_jp_{i,j}$. If $w(2)<w(1)$ and a lower-ranked label has positive marginal probability, this may strictly exceed the truthful score; a constant weight is an exception.
\end{proposition}

\begin{proof}
For the positional operator, the total expected score for
example $i$ under report $\hat{\mathbold{Y}}_{i,\cdot}$ is
\begin{equation}
  \mathbb{E}\Bigl[\sum_j D_w(Y_{i,j},\hat{\mathbold{Y}}_{i,\cdot})\Bigr]
  = \sum_j p_{i,j}\cdot w\!\bigl(\mathrm{rank}_i(j)\bigr),
  \label{eq:disc_expect}
\end{equation}
where $p_{i,j} = P(Y_{i,j}=1)$ and the ranks are determined by
the reported vector. Since $w$ is non-increasing, the rearrangement
inequality shows that (\ref{eq:disc_expect}) is maximised
by the rankings that sort labels in decreasing order of
$p_{i,j}$ (not necessarily uniquely so if $w$ itself has ties), among
reports whose entries are pairwise distinct, so that the rank is a permutation.
Reporting $\hat{\mathbold{Y}}_{i,\cdot} =
\mathbold{p}_i$ induces such an ordering, so the truthful report
attains the maximum among them and the rule is proper there. If tied entries are
allowed and ranked by the min-rank rule, a constant report assigns rank~$1$ to every label and
earns $w(1)\sum_jp_{i,j}$. For instance, with $m=2$, $\mathbold p_i=(0.7,0.3)$ and $w(r)=1/\log_2(r+1)$, the tied report earns $1$, whereas the truthful strict ordering earns $0.7+0.3/\log_2 3\approx0.889$; hence weak properness fails if such tied reports are admitted.

Strict properness fails: the score depends on the report only
through the induced ordering, so \emph{any} report that preserves
the ordering of $\mathbold{p}_i$ (for instance, any strictly
increasing transformation of the probabilities, such as $x\mapsto x/2$ or $x\mapsto x^{10}$) attains the same
expected score. For $m \geq 2$ these maximisers form an
uncountable set, so the truth is never the unique maximiser, and
the score cannot distinguish calibrated forecasts from arbitrary
order-preserving distortions of them.
\end{proof}

\begin{remark}[Normalised measures are not proper in general]
\label{rem:ndcg_not_proper}
Proposition~\ref{prop:disc_improper} concerns the raw positional operator. The normalised
$\mathrm{nDCG}_i=\mathrm{DCG}_i/\mathrm{IDCG}_i$ divides by a quantity depending on the random number
$k_i$ of relevant labels; consequently, truthful probability reporting need not maximise expected nDCG.
To avoid any ambiguity from tied true probabilities, consider $m=3$ and the label outcomes
$(1,0,0)$, $(0,1,1)$ and $(0,1,0)$ with respective probabilities $0.45$, $0.54$ and $0.01$. The true marginal vector is $\mathbold p_i=(0.45,0.55,0.54)$, whose entries are pairwise distinct and whose truthful ranking is $2\succ3\succ1$. For $w(r)=1/\log_2(r+1)$, the truthful report has expected nDCG $0.45\,w(3)+0.54+0.01=0.775$, whereas a report inducing $1\succ2\succ3$ has expectation
\[
  0.45+0.54\,\frac{w(2)+w(3)}{w(1)+w(2)}+0.01\,w(2)
  \approx 0.830760>0.775.
\]
Thus the failure of properness holds even when neither the truthful report nor the competing report has tied scores.
For $\mathrm{AUC}_j$, Proposition~\ref{prop:auc_improper} gives properness under independent labels (and for at most three examples under the stated conditions).
\end{remark}

\begin{remark}[Properness and evaluation purpose]
\label{rem:disc_purpose}
The failure of strict properness is not a defect of positional
operators; it reflects the fact that ranking evaluation asks a
different question than probabilistic
calibration~\citep{cortes2004auc}. A positional score is entirely
insensitive to order-preserving miscalibration: a classifier that
reports $\mathbold{p}_i^{\,10}$ (elementwise) scores identically
to one that reports $\mathbold{p}_i$. Conversely, a ranking
oracle that places every relevant label ahead of every irrelevant one (using distinct ranks) achieves the ideal per-example DCG but may be arbitrarily miscalibrated. The three frameworks
therefore target three distinct classifier
behaviours, none of which subsumes the others.
\end{remark}

\needspace{7\baselineskip}

% ============================================================
\subsection{Skew-Sensitivity of Ranking Measures}
\label{sec:rank_skew}
% ============================================================

We now characterise the skew-sensitivity of the DCG family.
The analysis differs from Section~\ref{sec:cal_skew} because
the positional operator acts on the \emph{label ranking within
an example}, not on per-class example counts. Skewing class
$j$ (rescaling $n_j^+$ vs.\ $n_j^-$) adds or removes
examples, but does not alter the rank of label $j$ for any
existing example. Two skew models are used in this section. The nDCG results
(Propositions~\ref{prop:ndcg_class_skew} and~\ref{prop:ndcg_ex_skew}, Corollary~\ref{cor:rank_skew_summary})
replicate whole examples, which leaves every example's row rank unchanged. The AUC results of
Proposition~\ref{prop:auc_skew}(b)--(c) use the per-class-summary convention of
Section~\ref{sec:cal_skew}: only the counts of one class change and the summaries of the other classes
are held fixed, because literally replicating examples also changes the positives and the composition
of the negatives of the other classes, so that their $\mathrm{AUC}_l$ would change.
$\mathrm{AUC}_j$ and $\mathrm{nDCG}_j^{\mathrm{class}}$ themselves are invariant under literal
replication. For exemplar averages in this subsection, we adopt the explicit convention $\mathrm{nDCG}_i=0$ when $k_i=0$, retaining such rows in the sample counts; dropping these rows instead defines a different average and requires recalculating its group counts. Per-class nDCG is reported only for $n_j^+>0$, while AUC requires both $n_j^+>0$ and $n_j^->0$. These conventions distinguish undefined normalisations from the implementation's zero-score convention for empty-label examples.

\begin{proposition}[Per-class nDCG is skew-invariant]
\label{prop:ndcg_class_skew}
$\mathrm{nDCG}_j^{\mathrm{class}}$, as defined in
(\ref{eq:ndcg_class}), is skew-invariant: under independent
rescaling of $n_j^+$ and $n_j^-$, it remains unchanged.
\end{proposition}

\begin{proof}
By Proposition~\ref{prop:ndcg_class_mean},
\[
  \mathrm{nDCG}_j^{\mathrm{class}}
  = \frac{1}{w(1)}\,\mathbb{E}[w(\mathrm{rank}_i(j))\mid Y_{ij}=1].
\]
This is a conditional mean over the population of positive
examples for class $j$. Since $\mathrm{rank}_i(j)$ depends
only on the predicted score vector $\hat{\mathbold{Y}}_{i,\cdot}$
of example $i$ (not on the class prevalence $\pi_j$ or the
overall sample counts), adding duplicate copies of positive
or negative examples (the skew operation) does not change the
within-class distribution of $\mathrm{rank}_i(j)$, and hence
does not change the conditional mean.
\end{proof}

\begin{proposition}[Exemplar nDCG is skew-sensitive]
\label{prop:ndcg_ex_skew}
Fix a class~$j$ with $n_j^+>0$ and $n_j^->0$, and let $\overline{\mathrm{nDCG}}^{+}$ and
$\overline{\mathrm{nDCG}}^{-}$ be the mean of $\mathrm{nDCG}_i$ over the
positive examples ($Y_{ij}=1$) and over the negative examples ($Y_{ij}=0$) of
class~$j$, respectively. Replicating the positive examples $\lambda$-fold
($\lambda>0$; for noninteger $\lambda$, interpreted as uniform weighting) changes the exemplar average
$\frac{1}{n}\sum_i \mathrm{nDCG}_i$ for some $\lambda$ if and only if
$\overline{\mathrm{nDCG}}^{+} \neq \overline{\mathrm{nDCG}}^{-}$; that is, the
exemplar average is skew-sensitive exactly when the two groups differ in mean
per-example nDCG.
\end{proposition}

\begin{proof}
After replicating the $n_j^+$ positive examples $\lambda$-fold, the exemplar
average is
\[
  \frac{\lambda n_j^{+}\,\overline{\mathrm{nDCG}}^{+}
        + n_j^{-}\,\overline{\mathrm{nDCG}}^{-}}
       {\lambda n_j^{+} + n_j^{-}},
\]
a weighted mean of the two group means whose weight on
$\overline{\mathrm{nDCG}}^{+}$ is strictly increasing in $\lambda$. For two
values $\lambda\neq\mu$, cross-multiplying, the two cross-products differ by
$n_j^+n_j^-(\lambda-\mu)\bigl(\overline{\mathrm{nDCG}}^{+}-
\overline{\mathrm{nDCG}}^{-}\bigr)$, so the two averages coincide if and only if
the group means do.
\end{proof}

\noindent The condition is a knife-edge one. A difference in the number of
relevant labels $k_i$ between the two groups, which may also change $\mathrm{IDCG}_i$, is
one way in which the group means can differ, but it is neither necessary nor
sufficient, and equal $k_i$ for all examples does not make the exemplar average
skew-invariant. For a concrete example with the standard logarithmic discount and three labels, suppose the sole relevant label of a class-$j$ positive example is ranked first, giving $\mathrm{nDCG}_i=1$, while the sole relevant label of a class-$j$ negative example is ranked third, giving $\mathrm{nDCG}_i=w(3)/w(1)=1/2$. The exemplar mean changes from $3/4$ to $5/6$ when the positive example is duplicated. By contrast, $\mathrm{nDCG}_j^{\mathrm{class}}$
conditions on the positive examples only and has no such dependence
(Proposition~\ref{prop:ndcg_class_skew}).

\begin{corollary}[Skew behaviour of the DCG family]
\label{cor:rank_skew_summary}
Fix a class~$j$ and replicate its positive examples $\lambda$-fold
($\lambda>0$). Write $\overline{(\cdot)}^{+}$ and $\overline{(\cdot)}^{-}$ for
means over the positive and the negative examples of class~$j$, and
$\mathrm{DCG}_i=\sum_{j'}Y_{ij'}\,w(\mathrm{rank}_i(j'))$. Assume $n_j^+>0$ and $n_j^->0$, and that the global micro-nDCG denominator is positive. The six
aggregations of the DCG family behave as follows; the value changes with
$\lambda$ for some $\lambda$ if and only if the stated condition \emph{fails}. In the last row, put $G^+=\{i:Y_{ij}=1\}$, $G^-=\{i:Y_{ij}=0\}$, and set $A^\pm=\sum_{i\in G^\pm}\mathrm{DCG}_i$, $B^\pm=\sum_{i\in G^\pm}\mathrm{IDCG}_i$. The cross-product condition does not require $B^->0$; if both group ideal sums are positive, it is equivalent to $A^+/B^+=A^-/B^-$.
\begin{center}
\small
\begin{tabular}{@{}lll@{}}
\toprule
Aggregation & Invariant for every data set? & Invariant if and only if \\
\midrule
$\mathrm{nDCG}_j^{\mathrm{class}}$ & yes, for every non-increasing $w$
  & (always; Prop.~\ref{prop:ndcg_class_skew}) \\[2pt]
$\mathrm{DCG}_j^{\mathrm{class}}$ & generally no
  & $\mathrm{nDCG}_j^{\mathrm{class}}=0$ \\[2pt]
$\mathrm{DCG}^{\mathrm{micro}}$, $\frac1n\sum_i\mathrm{DCG}_i^{\mathrm{ex}}$ & generally no
  & $\overline{\mathrm{DCG}}^{+}=\overline{\mathrm{DCG}}^{-}$ \\[2pt]
$\frac1n\sum_i\mathrm{nDCG}_i$ & generally no
  & $\overline{\mathrm{nDCG}}^{+}=\overline{\mathrm{nDCG}}^{-}$
    (Prop.~\ref{prop:ndcg_ex_skew}) \\[2pt]
$\mathrm{nDCG}^{\mathrm{micro}}$ & generally no
  & $A^+B^-=A^-B^+$ \\
\bottomrule
\end{tabular}
\end{center}
Thus $\mathrm{nDCG}_j^{\mathrm{class}}$ is invariant for every admissible non-increasing weight without an additional data-dependent equality. The other aggregations obey the exact conditions in the table; such conditions may hold automatically for particular weights. For instance, when $w(r)=1$ and every included example has $k_i>0$, both exemplar-mean and micro-nDCG are identically $1$ under skew. Hence no unqualified uniqueness claim is intended. The per-class profile $\{\mathrm{nDCG}_j^{\mathrm{class}}\}_{j=1}^m$
plays, in the Ranking Framework, the role that recall and specificity play in
the Discrimination Framework and that $(\alpha_j,\beta_j)$ plays in the
Calibration Framework.
\end{corollary}

\needspace{13\baselineskip}
\begin{proof}
The first row is Proposition~\ref{prop:ndcg_class_skew}. In the second,
$Y_{ij}w(\mathrm{rank}_i(j))$ vanishes on the negative examples, so
$\mathrm{DCG}_j^{\mathrm{class}}$ is the two-group average of the proof of
Proposition~\ref{prop:ndcg_ex_skew} with negative-group mean $0$ and
positive-group mean $w(1)\,\mathrm{nDCG}_j^{\mathrm{class}}$; the average is
independent of $\lambda$ if and only if the two group means coincide, i.e.\ if
and only if $\mathrm{nDCG}_j^{\mathrm{class}}=0$. Rows three and four are
the same two-group argument applied to the per-example quantities
$\mathrm{DCG}_i$ and $\mathrm{nDCG}_i$. In row three the two reported raw summaries coincide exactly, not merely up to a factor: $\mathrm{DCG}^{\mathrm{micro}}=(1/n)\sum_i\mathrm{DCG}_i^{\mathrm{ex}}=(1/(nm))\sum_i\mathrm{DCG}_i$, by Theorem~\ref{thm:dcg_collapse}(a). For the last row,
the replicated micro-nDCG is
$(\lambda A^{+}+A^{-})/(\lambda B^{+}+B^{-})$ with $A^{\pm}=\sum_{\pm}
\mathrm{DCG}_i$ and $B^{\pm}=\sum_{\pm}\mathrm{IDCG}_i$, where $B^+>0$ but $B^-$ may vanish; for $\lambda\neq\mu$
the cross-products $(\lambda A^{+}+A^{-})(\mu B^{+}+B^{-})$ and
$(\mu A^{+}+A^{-})(\lambda B^{+}+B^{-})$ differ by
$(\lambda-\mu)(A^{+}B^{-}-A^{-}B^{+})$, which vanishes if and only if
$A^+B^-=A^-B^+$; if $B^+B^->0$, this is equivalently $A^+/B^+=A^-/B^-$. \qedhere
\end{proof}

\needspace{6\baselineskip}

% ============================================================
\subsection{Column-Dependent Operators and the AUC Family}
\label{sec:col_dep}
% ============================================================

The row-dependent positional operator rewards label $j$ for being
ranked highly \emph{within example~$i$'s} score vector. The
dual operator asks whether example~$i$ is
ranked highly \emph{within class~$j$'s} score column.
This column-ranking perspective underlies AUC-based multilabel
evaluation~\citep{flach2003geometry}.

\begin{mydef}[Column rank and column-dependent positional operator]
\label{def:col_rank}
The \emph{column rank} of example $i$ for class $j$ is
\begin{equation}
  \mathrm{rank}_{j}(i)
  \;=\;
  \bigl|\{i' : \hat{Y}_{i',j} > \hat{Y}_{i,j}\}\bigr| + 1,
  \label{eq:col_rank}
\end{equation}
the position of $\hat{Y}_{i,j}$ in the descending ordering of
column $j$ of $\mathbold{\hat{Y}}$. A \emph{column-dependent
positional operator} with weight function $v$ is
\begin{equation}
  D_{v}\bigl(Y_{i,j},\,\hat{\mathbold{Y}}_{\cdot,j}\bigr)
  \;=\;
  Y_{i,j}\cdot v\!\bigl(\mathrm{rank}_{j}(i)\bigr).
  \label{eq:col_disc_op}
\end{equation}
\end{mydef}

\noindent The structural difference from the row-dependent case
is the axis of dependency: $D_v$ depends on the full predicted
column $\hat{\mathbold{Y}}_{\cdot,j}$ rather than the full
predicted row $\hat{\mathbold{Y}}_{i,\cdot}$.

\subsubsection*{AUC via the Wilcoxon--Mann--Whitney statistic}

The canonical column-dependent measure is AUC. For class $j$,
define the \emph{concordance fraction} of positive example $i$
as the fraction of negative examples it outranks:
\begin{equation}
  f(i,j)
  \;=\;
  \frac{1}{n_j^{-}}
  \sum_{i':\,Y_{i'j}=0}
  \Bigl(\mathbf{1}\!\left[\hat{Y}_{i,j} > \hat{Y}_{i',j}\right]
  +\tfrac12\,\mathbf{1}\!\left[\hat{Y}_{i,j} = \hat{Y}_{i',j}\right]\Bigr).
  \label{eq:concordance}
\end{equation}
This definition requires $n_j^->0$; the corresponding class AUC additionally requires $n_j^+>0$. The AUC cell matrix $\mathbold{A} \in [0,1]^{n\times m}$ is
\begin{equation}
  A_{i,j} \;=\; Y_{i,j}\cdot f(i,j),
  \label{eq:auc_cell}
\end{equation}
so $A_{i,j} = 0$ whenever $Y_{i,j}=0$.

\noindent At the per-class aggregate level, Proposition~\ref{prop:auc_meanrank} shows that AUC is an affine, decreasing transformation of the mean positive-example column rank obtained from the fixed-rank operator of Definition~\ref{def:col_rank} with $v(r)=r$. This aggregate identity does not equate the individual concordance cells $Y_{ij}f(i,j)$ with fixed rank-weighted cells $Y_{ij}v(\mathrm{rank}_j(i))$: $f(i,j)$ also depends on which other examples are negative. Thus the analogy with normalised DCG holds at the aggregate level, while the AUC cell construction retains its label-dependent concordance fractions.

\begin{mydef}[AUC family via the three aggregation operators]
\label{def:auc_family}
The ``micro-AUC'' defined below aggregates \emph{within-class} concordance values with positive-class-count weights. It is not the AUC of a single ROC curve obtained by pooling all class--example scores, which also involves cross-class positive--negative comparisons (Remark~\ref{rem:auc_micro_not_pooled}). The three aggregations of $\mathbold{A}$ yield:
\begin{align}
  \mathrm{AUC}_{j}
    &= \frac{\bigl[\Upsigma_{\mathrm{m}}(\mathbold{A})\bigr]_j}{n\,\pi_j}
     = \frac{1}{n_j^{+}}\sum_{i:\,Y_{ij}=1} f(i,j)
     = \mathbb{E}\!\left[f(i,j)\;\middle|\; Y_{ij}=1\right],
  \label{eq:auc_class}\\[4pt]
  \mathrm{AUC}^{\mathrm{micro}}
    &= \frac{\Upsigma_{\mathrm{1}}(\mathbold{A})}
            {\Upsigma_{\mathrm{1}}(\mathbold{Y})}
     = \frac{\sum_{j} n_j^{+}\,\mathrm{AUC}_j}
            {\sum_{j} n_j^{+}},
  \label{eq:auc_micro}\\[4pt]
  \mathrm{AUC}_{i}^{\mathrm{ex}}
    &= \frac{\bigl[\Upsigma_{\mathrm{n}}(\mathbold{A})\bigr]_i}{k_i}
     = \frac{1}{k_i}\sum_{j:\,Y_{ij}=1} f(i,j),
  \label{eq:auc_ex}
\end{align}
where $k_i = \sum_j Y_{ij}$. The per-example ratio is defined for $k_i>0$; empty-label examples require an explicit omission or zero-score convention. $\mathrm{AUC}_j$ is the standard
per-class AUC~\citep{fawcett2006introduction};
$\mathrm{AUC}^{\mathrm{micro}}$ is the prevalence-weighted
average of per-class AUCs\footnote{Throughout this paper,
$\mathrm{AUC}^{\mathrm{micro}}$ always refers to
Equation~\ref{eq:auc_micro}, the prevalence-weighted average of
per-class $\mathrm{AUC}_j$ values. This is \emph{not} the same
statistic as a globally-pooled single-ROC-curve AUC (computed by
ranking all classes' scores together, e.g.\ \texttt{scikit-learn}'s
\texttt{roc\_auc\_score(..., average="micro")}, or the
micro-level extension of \citet{prati2026unified}'s own
threshold-integral construction of AUC): the two may differ when
classes' score distributions are not on a common scale, which is
typical under a binary-relevance ensemble. See
Remark~\ref{rem:auc_micro_not_pooled} for the numerical gap and full
discussion.}; $\mathrm{AUC}_i^{\mathrm{ex}}$ is a
\emph{per-example AUC} (the average concordance fraction of
example~$i$ across its relevant labels), which does not appear
as a named measure in the existing literature. It compares example~$i$ with the negatives of each
relevant class (a column-axis quantity), and is not the per-example label-AUC of the row axis, which
equals one minus the multilabel ranking loss (see the grid below).
\end{mydef}

\noindent Equation~(\ref{eq:auc_class}) recovers the
Wilcoxon--Mann--Whitney statistic:
$\mathrm{AUC}_j = P(\hat{Y}_{i^+,j} > \hat{Y}_{i^-,j}) + \tfrac12 P(\hat{Y}_{i^+,j} = \hat{Y}_{i^-,j})$
for independently sampled positive $i^+$ and negative $i^-$ (the second term gives half credit to ties as in Equation~\eqref{eq:concordance}),
which equals the area under the ROC
curve~\citep{wilcoxon1945individual,mann1947test,hanley1982meaning}.

The next proposition identifies the exact aggregate-level rank relationship. It does not assert a cell-by-cell identity or that the affine AUC normalisation is literally an IDCG ratio.

\begin{proposition}[AUC as a normalised mean rank]
\label{prop:auc_meanrank}
Fix a class $j$ with $n_j^{+}$ positive and $n_j^{-}$ negative
examples, $n = n_j^{+} + n_j^{-}$, and let
\begin{equation}
  M_j \;=\; \frac{1}{n_j^{+}}\sum_{i:\,Y_{ij}=1}\mathrm{rank}_j(i)
  \label{eq:meanrank}
\end{equation}
be the mean descending column rank~(\ref{eq:col_rank}) of the
positive examples (the raw aggregate of the fixed-weight column
operator $D_v$ with $v(r)=r$). Assuming distinct scores within
column~$j$,
\begin{equation}
  \mathrm{AUC}_j
  \;=\;
  \frac{\,n - \tfrac{1}{2}(n_j^{+}-1) - M_j\,}{n_j^{-}},
  \label{eq:auc_meanrank}
\end{equation}
an affine, strictly decreasing function of $M_j$ whose
coefficients depend only on the counts $(n, n_j^{+}, n_j^{-})$.
Hence $M_j$ is skew-\emph{sensitive} in general (replicating negatives
can change the positive ranks and thus $M_j$), while
$\mathrm{AUC}_j$ is its skew-\emph{invariant} normalisation, the
affine map~(\ref{eq:auc_meanrank}) removing the prevalence
dependence.
\end{proposition}

\begin{proof}
A positive example $i$ outranks $n - \mathrm{rank}_j(i)$ of the
other examples. Summing over positives,
$\sum_{i:Y_{ij}=1}\!\bigl(n-\mathrm{rank}_j(i)\bigr)
 = n\,n_j^{+} - n_j^{+}M_j$ counts all pairs a positive outranks;
of these, $\binom{n_j^{+}}{2}=\tfrac12 n_j^{+}(n_j^{+}-1)$ join two
positives, leaving
$n\,n_j^{+} - n_j^{+}M_j - \tfrac12 n_j^{+}(n_j^{+}-1)$
positive--negative concordant pairs. Dividing by
$n_j^{+}n_j^{-}$ gives~(\ref{eq:auc_meanrank}); strict decrease in
$M_j$ and the count-only coefficients are immediate. With ties,
the same identity holds for mid-ranks and the $\tfrac12$ credit of~(\ref{eq:concordance}), the standard
Mann--Whitney correction; with the min-rank column rank of Definition~\ref{def:col_rank} it does not. The
formally verified statement assumes distinct scores.
\end{proof}

\needspace{6\baselineskip}

% ============================================================
\subsection{Row--Column Duality: nDCG and AUC}
\label{sec:duality}
% ============================================================

The two families share a common algebraic form, differing only
in the axis of the rank.

\begin{theorem}[Row--column duality]
\label{thm:duality}
The per-class measures of the nDCG and AUC families are both
conditional expectations over positive examples evaluating
rank quality along orthogonal axes:
\begin{align}
  \mathrm{nDCG}_{j}^{\mathrm{class}}
    &= \frac{1}{w(1)}\,\mathbb{E}\!\left[
       w\!\bigl(\mathrm{rank}_{i}(j)\bigr)
       \;\middle|\; Y_{ij}=1\right],
  \label{eq:ndcg_cond_dual}\\[4pt]
  \mathrm{AUC}_{j}
    &= \mathbb{E}\!\left[f(i,j)\;\middle|\; Y_{ij}=1\right].
  \label{eq:auc_cond_dual}
\end{align}
In~(\ref{eq:ndcg_cond_dual}), $\mathrm{rank}_{i}(j)$ is the
\emph{row rank}: position of label~$j$ in the descending
ordering of $\hat{\mathbold{Y}}_{i,\cdot}$.
In~(\ref{eq:auc_cond_dual}), $f(i,j)$ is the concordance fraction, the share of negative
examples that example~$i$ outranks. It is not by itself a function of the \emph{column rank}
$\mathrm{rank}_{j}(i)$, because the examples below $i$ include positives as well as negatives;
but its mean over the positives is, by Proposition~\ref{prop:auc_meanrank}, the affine
normalisation of the mean column rank. Both measures are therefore conditional means, over the
positive examples, of a rank-based quality: for nDCG a fixed function of the row rank, for AUC the
concordance fraction, whose average over the positives is an affine function of the mean column
rank.
\end{theorem}

\begin{proof}
Equation~(\ref{eq:ndcg_cond_dual}) is
Proposition~\ref{prop:ndcg_class_mean}. Equation~(\ref{eq:auc_cond_dual})
follows from Definition~\ref{def:auc_family},
equation~(\ref{eq:auc_class}). Both are conditional means over positive cells. For nDCG, the summand is a fixed discount of the row rank; for AUC, it is a positive--negative concordance fraction depending on the negative-score distribution. Only after aggregation over positives is the latter an affine function of the mean column rank (Proposition~\ref{prop:auc_meanrank}).
\end{proof}

\begin{example}[Running example: the two ranking axes disagree]
\label{ex:running_duality}
In Example~\ref{ex:running}, $A$ is ranked first in every row, so $\mathrm{nDCG}_A^{\mathrm{class}}=1$,
yet $\mathrm{AUC}_A=1/2$: its positives $(0.7,0.6)$ are not separated from its negatives $(0.8,0.4)$.
For $B$ it is the other way round: $\mathrm{AUC}_B=1$, since its positives $(0.6,0.2,0.3)$ all exceed
its negative $0.1$, but $B$ is never first in a row and $\mathrm{nDCG}_B^{\mathrm{class}}\approx0.587$.
Both are conditional means over the positives (Theorem~\ref{thm:duality}), of a row-rank and of a
column-rank quality respectively, and here they order the labels in opposite ways
($\mathrm{nDCG}_C^{\mathrm{class}}\approx0.631$, $\mathrm{AUC}_C=2/3$). The hard top-1 prediction is $A$
for every example, so $\mathrm{Precision}@1=1/2$ (Proposition~\ref{prop:hard_ranking}).
\end{example}

\begin{remark}[Two complementary ranking questions]
\label{rem:two_questions}
Theorem~\ref{thm:duality} makes explicit the distinct questions
the two families answer. $\mathrm{nDCG}_j^{\mathrm{class}}$
asks: \emph{for a randomly chosen positive example for label
$j$, how highly does the classifier rank $j$ among all labels
for that example?} $\mathrm{AUC}_j$ asks: \emph{for a randomly
chosen positive example for class $j$, what share of the negative
examples for class $j$ does the classifier rank below it?}
The two questions are independent: a classifier may rank each
example's positive labels highly (high nDCG) while failing to
separate positive from negative examples globally (low AUC), or
vice versa. A complete ranking evaluation requires both.
\end{remark}

The rest of this subsection discusses the weight $w$ itself and states no new formal
results.

The weight $w$ is an exogenous parameter of the positional
operator, and different choices recover the standard ranking
measures from the single template $Y_{i,j}\cdot w(\mathrm{rank})$,
in the same way that one $t$-norm generates
precision, recall and $F_1$ on the discrimination axis.
\begin{center}
\small
\begin{tabular}{@{}l l@{}}
\toprule
Weight $w(r)$ & Measure recovered \\
\midrule
$1/\log_2(r+1)$        & nDCG (canonical) \\
$\mathbf{1}[r \le k]$  & Precision@$k$ / recall@$k$ (hard-prediction limit) \\
$r$ (a cost, affinely normalised) & AUC (Proposition~\ref{prop:auc_meanrank}) \\
$1/r$                  & reciprocal rank; $\mathbb{E}[1/\mathrm{rank}_i(j)\mid Y_{ij}{=}1]$ is a per-class MRR \\
$\gamma^{\,r-1}$, $\gamma\in(0,1)$ & rank-biased precision (up to normalisation)~\citep{moffat2008rank} \\
\bottomrule
\end{tabular}
\end{center}
Two structural facts hold across the whole family. First,
skew-invariance is independent of $w$: for any
non-increasing $w$, Theorem~\ref{thm:duality} gives
$\mathrm{nDCG}_j^{\mathrm{class}} = \tfrac{1}{w(1)}
\mathbb{E}[w(\mathrm{rank}_i(j))\mid Y_{ij}{=}1]$, a conditional
mean over positives and hence skew-invariant; the weight sets
the discount profile but not the invariance. Second, no choice of
non-increasing $w$ makes the operator strictly proper
(Proposition~\ref{prop:disc_improper} holds for every
non-increasing $w$); the weight $w(r)=r$ is a cost to be minimised rather than a
gain, and AUC's non-strict properness is Proposition~\ref{prop:auc_improper}. The weight thus indexes named measures
without altering the structural profile of the class. The
geometric weight $\gamma^{\,r-1}$
carries a probabilistic user-persistence semantics (the user
inspects rank $r$ with probability $\gamma^{\,r-1}$) and a
generating-function form $\sum_{r\ge1} w(r)x^{r-1} = (1-\gamma x)^{-1}$,
and interpolates between top-heavy ($\gamma\to 0$) and uniform
($\gamma\to 1$) discounts through a single parameter.

The exogenous/effective distinction of
Section~\ref{sec:col_dep} is not specific to the column axis. On
the row axis, expected reciprocal rank~\citep{chapelle2009expected}
uses an \emph{effective} weight whose discount at position $r$
depends on the relevance of the higher-ranked labels (a cascade
user model), in the same way as the AUC weight depends on which
lower-ranked examples are negative. Expected reciprocal rank is
therefore to nDCG what AUC is to the raw mean rank: an
endogenous, label-dependent specialisation of the fixed-weight
operator. The exogenous-versus-effective dichotomy thus runs
along both axes, and in both cases the effective weight is a
normalisation of, or a data-dependent modulation of, the
underlying fixed-weight positional operator.

Crossing the two axes (row: rank labels within an example;
column: rank examples within a label) with the two
canonical exogenous weights (linear $w(r)=r$ and top-heavy
$w(r)=1/\log_2(r+1)$) places the named measures on a
$2\times2$ grid with one unfilled cell:
\begin{center}
\small
\begin{tabular}{@{}lcc@{}}
\toprule
 & linear weight & top-heavy weight \\
\midrule
row axis    & $1-\text{ranking loss}$ & nDCG \\
column axis & AUC & \emph{column-nDCG} (unfilled) \\
\bottomrule
\end{tabular}
\end{center}
The two standard measures sit on the anti-diagonal: nDCG is the
row axis under a top-heavy weight, and AUC (via
Proposition~\ref{prop:auc_meanrank}) is the column axis under a
linear weight, so they differ in both coordinates. The cells on the main diagonal are the duals with the weights
swapped. The row/linear cell corresponds to per-example label-pair ranking (one minus the multilabel ranking loss under matching conventions), which is distinct from the column-based $\mathrm{AUC}_i^{\mathrm{ex}}$ of Definition~\ref{def:auc_family}~\citep{schapire2000boostexter}. For a row with $0<k_i<m$ and distinct score ranks, this is the affine-normalised linear rank sum $1-\mathrm{RL}_i=1-\bigl(\sum_{j:Y_{ij}=1}\mathrm{rank}_i(j)-k_i(k_i+1)/2\bigr)/\bigl(k_i(m-k_i)\bigr)$; the raw cost $v(r)=r$ alone is not numerically equal to $1-\mathrm{RL}_i$.
The column/top-heavy cell, \emph{column-nDCG}, is not, to our knowledge,
a standard measure of multilabel evaluation (in bipartite ranking, the DCG of a class column is a member
of the linear rank-statistics family of \citet{clemencon2021concentration}): it asks whether the truly
positive examples of a label sit at the \emph{top} of that
label's ranked list, under a logarithmic discount. It is the
top-heavy alternative to AUC. Because AUC uses a linear rank
weight, it has no additional top-heavy discount beyond positive--negative pair ordering (although swapping a positive and a negative does change AUC), whereas column-nDCG (or its
rank-biased variant $w(r)=\gamma^{\,r-1}$) rewards concentrating
them at the top, the quantity that matters when the top-$k$
examples of a label are retrieved for downstream action. Close
relatives exist in retrieval and extreme classification (per-label
average precision, precision@$k$); the point here is structural:
the operator grid exposes column-nDCG as the
axis-swapped dual of nDCG rather than as an ad hoc metric.

% ============================================================
\subsection{Structural Properties of the AUC Family}
\label{sec:auc_props}
% ============================================================

\begin{theorem}[Collapse and non-collapse for AUC]
\label{thm:auc_collapse}
Assume $n_j^+>0$ and $n_j^->0$ for every class $j$, and $k_i>0$ for every example $i$, so that every exemplar AUC entering the mean over all $n$ examples is defined. Write $\widehat{\mathrm{Cov}}_j$ and $\widehat{\mathrm{Cov}}_i$ for the uniform empirical covariances over the $m$ classes and over the $n$ examples, respectively. Put $H=\sum_j n_j^+=\sum_i k_i>0$, $A_{ij}=Y_{ij}f(i,j)$, and denote the unweighted macro- and exemplar-means by
$\overline{\mathrm{AUC}}^{\mathrm{class}}=m^{-1}\sum_j\mathrm{AUC}_j$ and
$\overline{\mathrm{AUC}}^{\mathrm{ex}}=n^{-1}\sum_i\mathrm{AUC}_i^{\mathrm{ex}}$.
\begin{enumerate}
\item[\textup{(a)}] The raw cell aggregate satisfies
\begin{equation}\label{eq:auc_raw_collapse}
 \frac{\Upsigma_{\mathrm{1}}(\mathbold{A})}{nm}
 =\frac1m\sum_j\pi_j\mathrm{AUC}_j
 =\frac1n\sum_i\frac{k_i}{m}\mathrm{AUC}_i^{\mathrm{ex}}.
\end{equation}
\item[\textup{(b)}]
$\mathrm{AUC}^{\mathrm{micro}}=\overline{\mathrm{AUC}}^{\mathrm{class}}$
if and only if $\widehat{\mathrm{Cov}}_j(n_j^+,\mathrm{AUC}_j)=0$.
\item[\textup{(c)}]
$\mathrm{AUC}^{\mathrm{micro}}=\overline{\mathrm{AUC}}^{\mathrm{ex}}$
if and only if $\widehat{\mathrm{Cov}}_i(k_i,\mathrm{AUC}_i^{\mathrm{ex}})=0$.
\item[\textup{(d)}]
$\overline{\mathrm{AUC}}^{\mathrm{class}}=\overline{\mathrm{AUC}}^{\mathrm{ex}}$ if and only if
\begin{equation}\label{eq:auc_class_ex_eq}
\sum_{i,j:Y_{ij}=1} A_{ij}\left(\frac1{mn_j^+}-\frac1{nk_i}\right)=0.
\end{equation}
As an identity for \emph{every} real cell matrix supported on the positive cells of a fixed label matrix $Y$, this equality holds if and only if $mn_j^+=nk_i$ at every positive cell $(i,j)$. This stronger, universal statement is about supported cell arrays; it does not claim that arbitrary such arrays can be generated by AUC concordances.
\end{enumerate}
Thus the three normalised summaries need not coincide; their equality is governed by the stated conditions, rather than being excluded in all datasets.
\end{theorem}

\begin{proofsketch}
(a) sums the concordance cells $A_{ij}$ by rows and by columns. (b) and (c) are the covariance identity for
the weights $n_j^+$ and $k_i$. (d) expands both means as linear forms in the positive cells; equality for
every supported array forces each coefficient to vanish. The full proof is in
Appendix~\ref{app:proofs_rank}.
\end{proofsketch}

\begin{example}[Running example: collapse and non-collapse]
\label{ex:running_collapse}
The positional cells of Example~\ref{ex:running_cells} give the same raw DCG, $0.366$, whether averaged
over all cells, by labels or by examples (Theorem~\ref{thm:dcg_collapse}(a)); the normalised summaries
differ: micro-nDCG $0.835$, class mean $0.739$, exemplar mean $0.783$. Likewise the raw AUC aggregate is
$7/18$ in every order of summation, while macro-AUC is $13/18$, micro-AUC $7/9$ and the exemplar mean
$5/6$ (Theorem~\ref{thm:auc_collapse}). Micro and macro AUC differ because the class sizes
$n_j^+=(2,3,1)$ covary with $\mathrm{AUC}_j=(1/2,1,2/3)$.
\end{example}

\begin{proposition}[Per-class and macro-AUC are skew-invariant; micro-AUC may be skew-sensitive]
\label{prop:auc_skew}
Under independent $(\lambda,\mu)$ rescaling of positive and
negative example counts for class~$j$:
\begin{enumerate}
  \item[\textup{(a)}]
    $\mathrm{AUC}_j=P(\hat{Y}_{i^+,j}>\hat{Y}_{i^-,j})+\frac12P(\hat{Y}_{i^+,j}=\hat{Y}_{i^-,j})$ is skew-invariant (with the same half-credit tie convention as Definition~\ref{def:auc_family}).
  \item[\textup{(b)}]
    Macro-AUC $=\frac{1}{m}\sum_j\mathrm{AUC}_j$ is
    skew-invariant.
  \item[\textup{(c)}]
    Micro-AUC $=\frac{\sum_l n_l^+\mathrm{AUC}_l}{\sum_l n_l^+}$
    is skew-sensitive in general: scaling the positive examples of class~$j$
    by $\lambda$ changes it (for some $\lambda$) if and only if
    $\mathrm{AUC}_j$ differs from the micro-AUC itself. If the
    $\mathrm{AUC}_l$ are not all equal, such a class exists; if they are all
    equal, micro-AUC is invariant.
\end{enumerate}
\end{proposition}

\begin{proof}
\textit{(a)} Replicating positive examples $\lambda$-fold
leaves their score distribution unchanged; replicating
negatives $\mu$-fold leaves theirs unchanged. Hence both the strict-concordance probability and the tie probability are invariant; their half-credit combination, $\mathrm{AUC}_j$, is invariant.
\textit{(b)} Average of invariant terms.
\textit{(c)} Scaling the positives of class~$j$ by $\lambda$ replaces
$n_j^+$ by $\lambda n_j^+$ and leaves every $\mathrm{AUC}_l$ unchanged, by~(a).
With $S=\sum_l n_l^+\mathrm{AUC}_l$ and $N=\sum_l n_l^+$, micro-AUC becomes
$\frac{S+(\lambda-1)n_j^+\mathrm{AUC}_j}{N+(\lambda-1)n_j^+}$. For
$\lambda\neq\mu$, cross-multiplying, the two cross-products differ by
$n_j^+(\lambda-\mu)\bigl(\mathrm{AUC}_j\,N-S\bigr)$, which vanishes if and only
if $\mathrm{AUC}_j=S/N$, the micro-AUC. If $\mathrm{AUC}_j=S/N$ for every~$j$,
the $\mathrm{AUC}_j$ are all equal; otherwise some class has
$\mathrm{AUC}_j\neq S/N$.
\end{proof}

Proposition~\ref{prop:auc_skew} settles AUC's skew-invariance case by case. The same pattern has
now appeared in all three frameworks, and the next remark gives it a single explanation.

\begin{remark}[Unified derivation of three skew-invariance results]
\label{rem:auc_skew_known}
The skew-invariance of $\mathrm{AUC}_j$ is widely cited as
a practical advantage of AUC over accuracy-based
measures~\citep{fawcett2006introduction}. The present framework
provides a single algebraic explanation for the skew-invariance of recall
and $\mathrm{AUC}_j$, both known, and of the per-class $\mathrm{nDCG}_j^{\mathrm{class}}$
introduced here: all three are conditional means over positive
examples (Theorem~\ref{thm:duality} and the base
paper~\citep{prati2026unified}), and conditional means of a quantity that depends on the other
examples only through the within-class score distributions are preserved under replication of class
strata. (A conditional mean over positives of a rank within the whole column, such as the raw mean
rank $M_j$, is not preserved, because the rank depends on the number of negatives; $\mathrm{AUC}_j$ is
invariant because its summand is the share of negatives outranked, a functional of the negatives'
distribution. Recall, specificity and $\beta_j$ are conditional means over a class, positives or
negatives.) The Calibration Framework
yields the same conclusion for $(\alpha_j,\beta_j)$ via
Theorem~\ref{thm:cal_skew}. Skew-invariance thus holds for these class-conditional aggregations because both the conditioning strata and the quantities averaged remain unchanged under the specified group reweighting. Conditioning alone is not sufficient when the summand itself changes, as the raw mean column rank illustrates.
\end{remark}

Properness (Section~\ref{sec:proper_scoring}) is a property independent of skew-invariance; the
next proposition settles it for AUC.

\begin{proposition}[AUC is non-strict; truthful ranking may fail under dependence]
\label{prop:auc_improper}
Fix $n\ge2$ examples of one class, with Bernoulli labels $Y_1,\ldots,Y_n$ and marginal probabilities $p_i=P(Y_i=1)$. Reports $q\in[0,1]^n$ are evaluated by pairwise AUC with half credit for ties. For each label realisation, adopt the convention $\mathrm{AUC}(q,Y)=0$ whenever $n^+n^-=0$; otherwise use the ordinary concordance fraction. Then:
\begin{enumerate}
\item[\textup{(a)}] Under any joint law of the labels, the expected AUC cannot be uniquely maximised by a report: a distinct report with the same order and ties has the same value.
\item[\textup{(b)}] If the labels of different examples are independent, ordering scores by the truthful marginals $p_i$ maximises expected AUC (ties between equal marginals are allowed). Hence AUC is not strictly proper even in this setting.
\item[\textup{(c)}] For $n=2$ or $n=3$, the truthful ordering maximises expected AUC for \emph{every} joint law, without independence.
\item[\textup{(d)}] For every $n\ge4$, there exists a joint label distribution with distinct marginals for which a non-truthful ordering has strictly greater expected AUC. Thus truthful maximality is not guaranteed under arbitrary dependence.
\end{enumerate}
\end{proposition}

\begin{proof}
\noindent\textit{(a)} AUC depends on $q$ only through the strict comparisons and ties between reported scores, all preserved by a strictly increasing transformation $g:[0,1]\to[0,1]$. For any report $q\ne0$, take $g(t)=t/2$, so $g(q)\ne q$; if $q=0$, take $g(t)=(1+t)/2$. In either case the transformed report is different but has exactly the same AUC for every label vector, and hence the same expectation.

\noindent\textit{(b)} Put $Y_i^c=1-Y_i$ and, for $i\ne i'$, define
\[
 b_{ii'}=\mathbb E\left[\frac{Y_iY_{i'}^c}{n^+n^-}\,\mathbf 1[n^+n^->0]\right].
\]
A report ordering $q_i>q_{i'}$ receives $b_{ii'}$ from this ordered pair; a tie receives $(b_{ii'}+b_{i'i})/2$. Under label independence, conditioning on $Y_i=1$, $Y_{i'}=0$ leaves the other labels independent of this pair. On this event, $n^+=1+R^+$ and $n^-=1+R^-$, where $R^+,R^-$ count the labels of the remaining $n-2$ examples. Hence
\[
 b_{ii'}=p_i(1-p_{i'})K_{ii'},\quad b_{i'i}=p_{i'}(1-p_i)K_{ii'},\quad
 K_{ii'}=\mathbb E\left[\frac1{(1+R^+)(1+R^-)}\right]>0.
\]
The same $K_{ii'}$ occurs in both orientations, so
\[
 b_{ii'}-b_{i'i}=K_{ii'}(p_i-p_{i'}).
\]
Every pair is therefore optimally ordered by decreasing $p_i$, with ties harmless where $p_i=p_{i'}$. The truthful report $q=p$ achieves all these pairwise optima simultaneously.

Parts (c) and (d) are proved in Appendix~\ref{app:auc_improper_proof}.
\end{proof}

The next two remarks relate this column-dependent construction to the earlier threshold-integral
account of AUC in the Discrimination Framework and state how far the relation extends.

\begin{remark}[Threshold-integral and column-dependent views are equivalent]
\label{rem:auc_threshold}
\citet{prati2026unified} derive AUC within the Discrimination Framework via the
threshold-integral view: at each threshold~$\theta$, the
thresholded predictions produce cell matrices
$\mathbold{TP}(\theta)$, $\mathbold{FP}(\theta)$, etc., and
AUC integrates the (FPR$(\theta)$, TPR$(\theta)$) curve.
The column-dependent view introduced here is numerically
equivalent when the ROC curve uses the matching half-credit (mid-rank/trapezoidal) convention for tied scores (both then give the same $\mathrm{AUC}_j$), but it
exposes different structure. The threshold view embeds AUC
in the Discrimination Framework as a functional of coincidence-operator cell
matrices parameterised by~$\theta$. The column-dependent view
embeds it in the Ranking Framework as a conditional expectation,
making the duality with nDCG visible
(Theorem~\ref{thm:duality}) and yielding a uniform proof of
skew-invariance (Proposition~\ref{prop:auc_skew}) that
generalises across all three frameworks.
\end{remark}

\begin{remark}[Per-class AUC equivalence does not identify the two micro-AUC constructions]
\label{rem:auc_micro_not_pooled}
\begin{enumerate}
\item[\textup{(i)}] The threshold-integral and concordance views agree for each class $j$ when tied scores receive the same half-credit convention (Remark~\ref{rem:auc_threshold}). The micro-AUC in this paper is the positive-prevalence-weighted class mean from~\eqref{eq:auc_micro}:
\[
\mathrm{AUC}^{\mathrm{micro}}=\frac{\sum_j n_j^+\mathrm{AUC}_j}{\sum_j n_j^+}.
\]
Globally pooled ROC-AUC instead pools all class--example predictions and true labels before constructing a single ROC curve.
\item[\textup{(ii)}] The two need not agree. Pooled ROC-AUC includes cross-class positive--negative comparisons, whereas the weighted class mean uses only within-class comparisons. The resulting values \emph{may} differ, particularly when class score distributions are not directly comparable; unequal score scales alone do not force a nonzero gap.
\item[\textup{(iii)}] A seeded synthetic computation (supplementary script \texttt{pooled\_vs\_micro\_auc.py}: $n=200$ examples, five independent binary-relevance classes with prevalences in $[0.15,0.5]$, $1000$ draws) illustrates the size of the difference. When each class's scores pass through its own increasing logistic map, so that the classes are on different scales, the weighted micro-AUC exceeds pooled ROC-AUC by $0.072$ on average (standard deviation $0.042$ across draws; $5$--$95\%$ range of the absolute difference $0.012$--$0.146$). With a common scale for all classes, the absolute difference has median $0.002$ and $95$th percentile $0.006$. These values describe this generator only; they do not quantify the gap for other score distributions. In the running example (Example~\ref{ex:running}), pooled ROC-AUC is $43/72\approx0.597$ against the weighted micro-AUC $7/9\approx0.778$: pooling compares the high scores of $A$ on its negatives with the low scores of $B$ on its positives. Although scores are distinct within each row and each column, pooling creates three cross-class positive--negative ties, each assigned half credit. Of the $6\times6=36$ pooled positive--negative comparisons, 20 are concordant, 3 are tied and 13 are discordant, giving $\mathrm{AUC}_{\mathrm{pooled}}=(20+\frac12\cdot3)/36=43/72$. Throughout this paper, ``micro-AUC'' denotes only~\eqref{eq:auc_micro}, not pooled ROC-AUC.
\end{enumerate}
\end{remark}

\begin{table}[!t]
\centering
\small
\setlength{\tabcolsep}{3pt}
\renewcommand{\arraystretch}{1.18}
\begin{tabular}{@{}p{0.215\linewidth}|p{0.18\linewidth}p{0.245\linewidth}p{0.28\linewidth}@{}}
\hline
\textbf{Property} & \textbf{Discrimination} & \textbf{Calibration} & \textbf{Ranking} \\
\hline
Operator dependency & Pointwise & Pointwise & Row (nDCG) / Col (AUC) \\
Collapses (raw cell sums / normalised ratios) & Yes / No & Yes / Yes & Yes / No \\
Strictly proper scoring rule & $\times$ & Depends on loss (Brier: strict; MAE: not proper) & nDCG: not generally proper; AUC: non-strict and conditional \\
Skew-invariant basis & Recall, specificity & $(\alpha_j,\beta_j)$ & $\mathrm{nDCG}_j^{\mathrm{class}}$, $\mathrm{AUC}_j$ \\
Common structure of basis & Class-conditional rates & Positive/negative conditional error means & Positive-class conditional rank quality \\
Hard-prediction limit & Precision, recall & Hamming loss (complement: label accuracy) & Precision@$k$ (indicator weight) \\
\hline
\end{tabular}
\caption{Structural properties of the three frameworks.
For Ranking, nDCG and AUC are row-dependent and column-dependent subfamilies with raw aggregation collapse but generally different normalised summaries. Properness depends on the functional and its assumptions: normalised nDCG need not be proper, while AUC is never strictly proper and may fail truthful maximality under dependent labels. The skew-invariant bases use the appropriate conditioning strata, not only the positive examples.}
\label{tab:framework_comparison}
\end{table}

% ============================================================
%  References used in this section:
%  \cite{jarvelin2002cumulated} -- nDCG original paper
%  \cite{manning2008introduction} -- precision@k
%  \cite{fawcett2006introduction} -- ROC/AUC
%  \cite{hanley1982meaning} -- WMW = AUC
%  \cite{prati2026unified} -- prior framework
% ============================================================

% ============================================================
%  Cross-Framework Synthesis
% ============================================================
% ============================================================
%  Cross-Framework Synthesis
%  Draft for: "A Generalised Confusion Matrix Framework"
%  Author: Ronaldo C. Prati
% ============================================================

\section{Cross-Framework Synthesis}
\label{sec:synthesis}

Sections~\ref{sec:calibration} and~\ref{sec:ranking} developed
Frameworks~2 and~3 in full. This section draws the three frameworks
together, comparing their theoretical properties and identifying the
common algebraic template that underlies all three.

% ============================================================
\subsection{Properties of the Three Classes}
\label{sec:tax_properties}

Table~\ref{tab:op_properties} summarises the theoretical
properties of the three cell operator classes established in
Sections~\ref{sec:taxonomy}--\ref{sec:ranking}.

\begin{table}[!t]
\centering
\small
\setlength{\tabcolsep}{4pt}
\begin{tabular}{@{}>{\raggedright\arraybackslash}p{2.08in}|>{\centering\arraybackslash}p{1.02in}>{\centering\arraybackslash}p{1.02in}>{\centering\arraybackslash}p{1.02in}@{}}
\hline
\textbf{Property} &
  \textbf{Discrimination} &
  \textbf{Calibration} &
  \textbf{Ranking} \\
\hline
Pointwise in $(Y_{i,j}, \hat{Y}_{i,j})$
  & \checkmark & \checkmark & $\times$ \\
$D(y,y) = 0$ for all $y$
  & $\times$ & \checkmark & n/a$^*$ \\
Strictly proper scoring rule
  & $\times$ & \checkmark$^\ddagger$ & $\times$$^\dagger$ \\
Marginal consistency (Thm.~\ref{thm:partition_bg}, \citealt{prati2026unified})
  & $T_P$ only & n/a & n/a \\
Micro = macro = exemplar (scalar)
  & $\times$ & \checkmark & raw \checkmark{}; norm.\ $\times$ \\
Skew-invariant measures exist
  & \checkmark & \checkmark (via $(\alpha_j, \beta_j)$) & \checkmark (per-class)$^\S$ \\
\hline
\end{tabular}
\caption{Properties of the three cell operator classes.
Operator types: Discrimination = coincidence (Boolean conjunction,
generalised by $t$-norms); Calibration = deviation
($\phi(y-\hat{y})$); Ranking = positional
($Y \cdot w(\mathrm{rank})$).
$^*$~A positional operator is a gain, not a loss: $D = Y_{i,j}\,w(\mathrm{rank})$
vanishes only when $Y_{i,j}=0$ (irrelevant label) or when the weight does, and is
largest, $D = w(1)\,Y_{i,j}$, when the label is ranked first.
$^\dagger$~Raw positional operators are proper among reports with distinct entries (the truthful
report maximises the expected score) but never strictly
proper: any order-preserving distortion of the probabilities
attains the same score
(Proposition~\ref{prop:disc_improper}). With ties ranked by the min-rank rule, raw positional operators need not be proper (non-constant discounts admit counterexamples, while a constant weight makes every report a maximiser);
normalised nDCG is not proper in general (Remark~\ref{rem:ndcg_not_proper}).
$^\ddagger$~For the symmetric deviation operators $\phi(y-\hat{y})$ with $\phi$ even, convex, finite-valued and continuous on $[-1,1]$ and differentiable on $(0,1)$, properness holds exactly when $\phi'(t)/t$ is symmetric about $\tfrac12$; strictly positive weights additionally guarantee strict properness (Corollary~\ref{cor:savage_cases}(c)). Examples include squared error and the nonquadratic loss of Appendix~\ref{app:savage_examples}; log-loss satisfies the weight condition only in the extended-valued sense of Corollary~\ref{cor:savage_cases}(b). Convexity alone is
insufficient: $\phi(t)=|t|$ (MAE) yields a non-proper operator
whose minimiser is the conditional median rather than the
conditional mean (Remark~\ref{rem:mae_nonproper}).
$^\S$~Per-class nDCG$_j^{\mathrm{class}}$ and per-class
AUC$_j$ are skew-invariant
(Propositions~\ref{prop:ndcg_class_skew}
and~\ref{prop:auc_skew}); exemplar and micro summaries
are not in general.}
\label{tab:op_properties}
\end{table}

\noindent Several properties deserve comment.

First, coincidence operators have $D(y,y) \neq 0$ in general:
under the product $t$-norm, $T_P(y,y) = y^{2}$, which equals~1
only when $y = 1$ (certain positive) and~0 only when $y = 0$
(certain negative). For a soft label $y = 0.7$ with
equally soft prediction $\hat{y} = 0.7$, the TP cell value is
$0.49$, not~1. This is appropriate for discrimination: the
classifier correctly predicted the soft label, but there is still
a $0.3$ probability that the label is absent, so a full TP credit
would overstate certainty. Deviation operators, by contrast, give
$D(0.7, 0.7) = 0$: a prediction that exactly matches the soft
label incurs zero calibration error.

Second, the micro~=~macro~=~exemplar collapse survives normalisation only in the
Calibration Framework. For coincidence operators, micro- and macro-averaging differ
in general when per-class denominators are unequal (they agree for every numerator only if the
denominators are equal; Corollary~1 of \citealt{prati2026unified}).
For deviation operators, all three aggregation schemes give the same
scalar Brier score (Theorem~\ref{thm:brier_collapse}), because the
Brier score is a linear aggregation over cells with a fixed
denominator $nm$. For positional operators the raw aggregates collapse as well, since they are sums of cells
(Theorem~\ref{thm:dcg_collapse}(a)); what breaks the collapse is normalisation by
heterogeneous denominators, the ideal DCG of each example or class, so that normalised nDCG
can differ across the three schemes (Theorem~\ref{thm:dcg_collapse}(b)--(d)).

Third, properness selects a sub-family of the deviation operators within the
Calibration Framework: among symmetric deviation operators, properness holds
exactly when the weight $\phi'(t)/t$ is symmetric about $\tfrac12$
(Corollary~\ref{cor:savage_cases}(c)); squared error is the simplest member and
the only power-law one. Other choices such as
$\phi(t) = |t|$ (MAE) are structurally within the Calibration Framework and
still satisfy the micro~=~macro~=~exemplar collapse, but are not
proper: their minimiser is the conditional median rather than
the true probability (Remark~\ref{rem:mae_nonproper}), and the
Murphy decomposition does not apply. Proper alternatives such as
log-loss admit the Savage representation
(Theorem~\ref{thm:savage}); log-loss is itself a deviation operator, with
$\phi(t)=-\log(1-|t|)$.

Fourth, under class-indexed skew transformations that preserve the relevant conditional \emph{cell-quality distributions}, skew-invariance holds for the per-class building blocks in all three frameworks and for their unweighted means; pooled and exemplar aggregates are generally skew-sensitive. For per-class AUC, preservation of the positive- and negative-conditional score distributions suffices. For per-class nDCG, preservation of each label's own conditional score distribution \emph{alone} is insufficient: the conditional distribution of the within-row rank (or at least its expected discount) among the positives must also be preserved. In the Discrimination and
Calibration Frameworks, skew-invariant measures exist at the
per-class and macro levels: unweighted means of skew-invariant per-class
quantities remain skew-invariant, e.g.\ macro-recall and the
unweighted means of $\alpha_j$ and $\beta_j$ (note that
macro-$F_1$ is \emph{not} skew-invariant, since $F_1$ depends on
precision, and the individual Murphy terms are not skew-invariant in general
either, by Corollary~\ref{cor:murphy_skew}, up to the degenerate exceptions listed there,
such as $\mathrm{RES}_j\equiv0$ when $F_j^+=F_j^-$). In
the Ranking Framework, $\mathrm{nDCG}_j^{\mathrm{class}}$ (Proposition~\ref{prop:ndcg_class_skew})
and $\mathrm{AUC}_j$ (Proposition~\ref{prop:auc_skew}) are skew-invariant, and so are their unweighted means
(macro-AUC under the per-class convention of Section~\ref{sec:cal_skew}; the multiclass caveat of
Corollary~\ref{cor:micro_skew_base} applies to the per-class rates). Replicating or removing entire multilabel rows is a different operation: it may change the positive- or negative-conditional score distributions of other classes, so the class-indexed invariance statement need not hold for every affected class. Exemplar
and micro positional aggregations reintroduce label-prevalence weights and are therefore not skew-invariant in general, as with prevalence-weighted discrimination summaries; Brier averages can instead change because the individual classwise losses depend on prevalence.

% ============================================================
\subsection{A Unified Skew-Invariant Template}
\label{sec:synthesis_skew}

Across the three frameworks, the skew-invariant per-class measures share a common algebraic template: each is a conditional mean of a cell-level quality measure given the relevant positive \emph{or} negative class outcome, with the distribution of that conditional \emph{cell-quality quantity} held fixed.

\begin{itemize}
  \item In the Discrimination Framework, recall$_j = \mathbb{E}[\hat{Y}_{ij} \mid
    Y_{ij}=1]$ and specificity$_j = \mathbb{E}[1-\hat{Y}_{ij} \mid
    Y_{ij}=0]$ form the skew-invariant basis.
  \item In the Calibration Framework, $\alpha_j = \mathbb{E}[(1-\hat{Y}_{ij})^2
    \mid Y_{ij}=1]$ (miss-calibration of positives) and $\beta_j =
    \mathbb{E}[\hat{Y}_{ij}^2 \mid Y_{ij}=0]$ (false-alarm calibration)
    form the per-class analogue
    (Theorem~\ref{thm:cal_skew}).
  \item In the Ranking Framework, $\mathrm{nDCG}_j^{\mathrm{class}} =
    \frac{1}{w(1)}\mathbb{E}[w(\mathrm{rank}_i(j)) \mid Y_{ij}=1]$
    and $\mathrm{AUC}_j = \mathbb{E}[f(i,j) \mid Y_{ij}=1]$, with $f(i,j)$ the share of negatives of class~$j$
    that example~$i$ outranks, are conditional means of rank-based quality over positive
    examples (Propositions~\ref{prop:ndcg_class_skew}
    and~\ref{prop:auc_skew}).
\end{itemize}

Conditioning on the appropriate outcome ($Y_{ij}=1$ or $Y_{ij}=0$) removes explicit weighting by the class prevalence $\pi_j = P(Y_{ij}=1)$; invariance additionally requires the relevant conditional cell-quality distribution to remain unchanged under the skew transformation. Any measure that can be
expressed as $\mathbb{E}[q \mid Y_{ij}=1]$ (or $\mathbb{E}[q \mid Y_{ij}=0]$), where $q$ depends on the
rest of the data only through the within-class score distributions, is skew-invariant under
independent rescaling of positive and negative counts. The condition on $q$ is needed: the raw mean column rank
of the positives is a conditional mean over the positives but depends on the number of negatives,
and is skew-sensitive. Likewise, for per-class nDCG a change in the joint row-score configuration can alter $\mathrm{rank}_i(j)$ even when the positive- and negative-conditional distributions of each column's individual scores are unchanged; preservation of $F_j^+$ and $F_j^-$ alone is therefore not the relevant invariance hypothesis. The three
frameworks share this structure because in each of them the
per-class building blocks are cell values conditioned on the class label:
coincidence operators give TP (conditioned on $Y_{ij}=1$) and TN (on $Y_{ij}=0$);
deviation operators give the calibration errors $\alpha_j$ (on $Y_{ij}=1$) and
$\beta_j$ (on $Y_{ij}=0$); positional operators give a weighted gain that is
non-zero only on the positives. The template is therefore a pair of class-conditional
means, over the positives and over the negatives, with the ranking measures using only the
positive side.

Pooled aggregates reintroduce skew sensitivity, though not always for the same reason. For
the ratio measures of the Discrimination Framework and for micro-AUC, the pooled value is a
weighted mean of per-class values with prevalence-dependent weights. For the Brier score,
micro~=~macro, so the pooled value is the plain average of the per-class scores, each of which is
itself skew-sensitive ($\mathrm{BS}_j=\pi_j\alpha_j+(1-\pi_j)\beta_j$). The common remedy across all three paradigms is to
aggregate the per-class skew-invariant building blocks ($\mathrm{rec}_j$, $\mathrm{spec}_j$, $\alpha_j$,
$\beta_j$, $\mathrm{AUC}_j$, $\mathrm{nDCG}_j^{\mathrm{class}}$) and then average, ensuring
that each class contributes equally regardless of its frequency; for the Brier score this means averaging
$\alpha_j$ and $\beta_j$, not $\mathrm{BS}_j$.

\needspace{16\baselineskip}
\begin{example}[Running example: a change of prevalence]
\label{ex:running_skew}
Double the positive examples of $A$ in Example~\ref{ex:running} under the per-class convention, so that
$\pi_A$ rises from $1/2$ to $2/3$ and nothing else changes:
\begin{center}
\normalfont\small
\begin{tabular}{@{}lcc|lcc@{}}
\hline
Class-conditional & before & after & Prevalence-weighted & before & after\\
\hline
$\mathrm{rec}_A$, $\mathrm{spec}_A$ & $1$, $1/2$ & $1$, $1/2$ & $\mathrm{Precision}_A$ & $2/3$ & $4/5$\\
$(\alpha_A,\beta_A)$ & $(1/8,2/5)$ & $(1/8,2/5)$ & $\mathrm{BS}_A$ & $21/80$ & $13/60$\\
$\mathrm{AUC}_A$, $\mathrm{nDCG}_A^{\mathrm{class}}$ & $1/2$, $1$ & $1/2$, $1$ & micro-AUC & $7/9$ & $17/24$\\
macro-AUC & $13/18$ & $13/18$ & & & \\
\hline
\end{tabular}
\end{center}
Every class-conditional mean stays fixed; every summary that weights by prevalence moves. Duplicating the
two examples in which $A$ is relevant, as whole rows, is a different operation: it also changes the other
classes ($\mathrm{AUC}_C$ from $2/3$ to $3/4$, $\mathrm{nDCG}_B^{\mathrm{class}}$ from $0.587$ to $0.598$),
the coupling between classes seen in Experiment~3.
\end{example}

% ============================================================
%  Section 6 — Experiments
% ============================================================
% ============================================================
%  Section 6: Experiments
%  Draft for: "A Generalised Confusion Matrix Framework"
%  Author: Ronaldo C. Prati
% ============================================================

\section{Experiments}
\label{sec:experiments}

The seven experiments below illustrate the core theoretical claims of the paper.
Each uses at least one of the multilabel benchmark datasets described in
Section~\ref{sec:exp_setup}, and each is designed to produce a
qualitative demonstration rather than a performance benchmark:
the goal is to show that the structural properties derived in
Sections~\ref{sec:calibration}--\ref{sec:synthesis} hold empirically on real data
and are not artefacts of the algebraic framework alone.

The first four experiments share the Emotions dataset, whose six
human-interpretable labels make individual classifier behaviour easy to inspect
directly. Experiments~1--3 establish, in
turn, that the three frameworks disagree on classifier ranking, that they differ
in whether their own aggregation levels collapse to one scalar, and that the
skew-invariant per-class measures stay flat under a prevalence sweep
while the corresponding skew-sensitive aggregates do not. Experiment~4 then
shows that an order-preserving
distortion of a classifier's own probabilities is invisible to every ranking
measure while moving its Brier score and its MAE in opposite directions. The
remaining three experiments mostly use Yeast (Experiment~6 uses it for panel~(a) and Emotions for
panel~(b)), whose fourteen labels support a finer-grained dissection: Experiments~5 and~6 decompose the Brier score along
the class axis and the example axis respectively, and Experiment~7 closes the
section by comparing the two Ranking Framework measures directly on the same
labels, showing where they disagree.

% ============================================================
\subsection{Experimental Setup}
\label{sec:exp_setup}
% ============================================================

We use two multilabel benchmark datasets drawn from the MULAN
collection~\citep{tsoumakas2011mulan}, retrieved via
OpenML~\citep{vanschoren2014openml}. The
\textbf{Emotions} dataset~\citep{trohidis2008multi} consists of 593 music
clips described by 72 spectral features; each clip is annotated with one or more
of six emotional categories (amazed-surprised, happy-pleased, relaxing-calm,
quiet-still, sad-lonely, angry-aggressive), giving a label-set density of
approximately $0.31$.
The \textbf{Yeast} dataset~\citep{elisseeff2002kernel} consists of 2417 yeast
genes described by 103 expression-profile features; each gene is annotated with
one or more of 14 functional classes, with label-set density approximately
$0.30$. Emotions provides a small, human-interpretable label space suitable for
visualising individual class behaviour; Yeast provides a richer label space
($m = 14$) that yields a wider range of per-example label cardinalities.
Each dataset is split 70/30 into training and test sets using a fixed random seed.

Five classifiers are trained with the binary relevance
strategy~\citep{DBLP:journals/jdwm/TsoumakasK07} (one binary sub-classifier per label):
Logistic Regression (\textbf{LR}), Random Forest (\textbf{RF}),
Gradient Boosted Trees (\textbf{GBT}), an uncalibrated Multi-Layer Perceptron
(\textbf{MLP (uncal.)}), and the same MLP with sigmoid Platt scaling applied via
cross-validation (\textbf{MLP (cal.)}). The last pair is deliberately included
to isolate the effect of post-hoc calibration while keeping the base
classifier constant.
All feature vectors are min-max normalised to $[0,1]$.
Hard binary predictions are obtained by thresholding soft probabilities at~$0.5$.
Ranks are computed with ties broken by label index (NumPy's \texttt{argsort}), and AUC counts tied
positive--negative pairs as~$\tfrac12$ (scikit-learn). Ties arise mainly in the random forest, whose
scores tie within $36$ of the $178$ Emotions test rows and $524$ of the $726$ Yeast test rows. GBT repeats values
within a few columns ($1$ of $6$ labels in Emotions, $6$ of $14$ in Yeast), and LR and the two MLPs have no ties.
The tie rule therefore matters for RF and, in macro-AUC on Yeast, for GBT, which falls from $0.7050$ to $0.6966$
if ties count~$0$ and so changes places with the uncalibrated MLP ($0.6993$); for the other classifiers the
conventions agree (Experiment~1 and the files \texttt{tie\_sensitivity\_emotions.csv} and
\texttt{tie\_sensitivity\_yeast.csv} in the repository).

% ============================================================
\subsection{Framework Disagreement in Classifier Ranking}
\label{sec:exp1}
% ============================================================

\begin{figure}[!t]
  \centering
  \includegraphics[width=\linewidth]{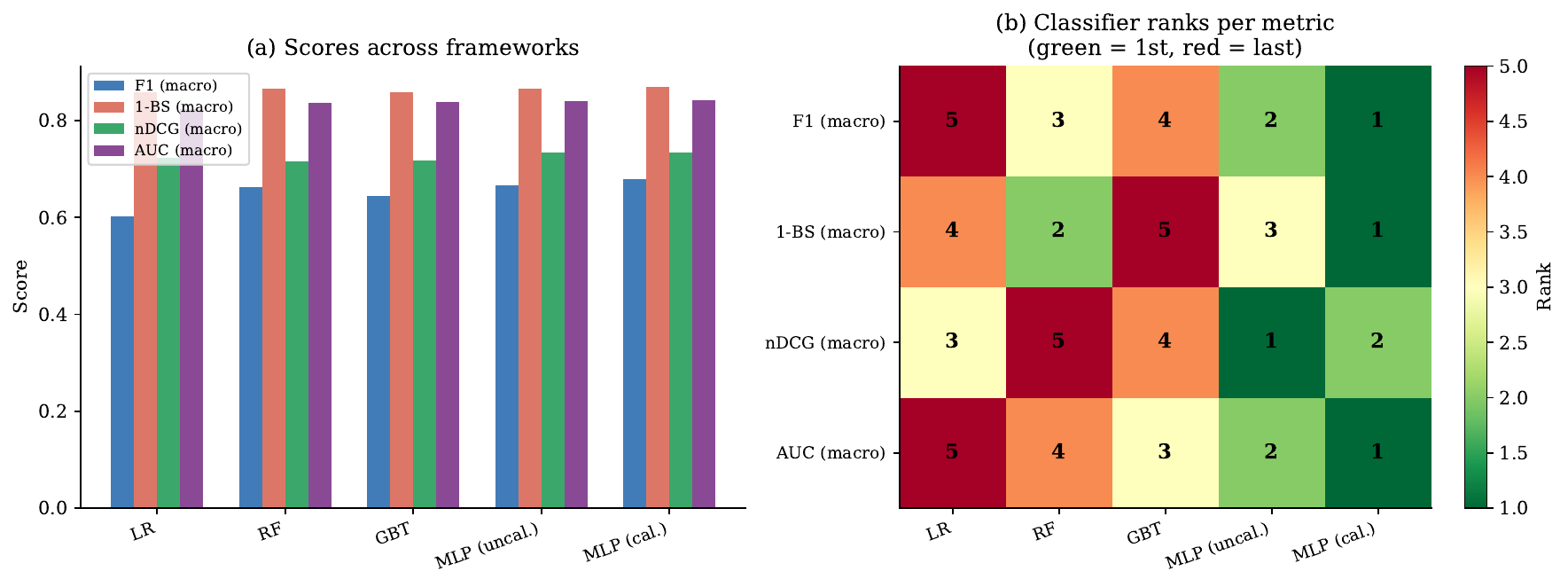}
  \caption{Experiment~1 (Emotions). (a) Scores across frameworks for five
    classifiers. (b) Rank heatmap (green = 1st, red = last). No two metrics
    produce the same classifier ordering.}
  \label{fig:exp1}
\end{figure}

The first experiment asks whether the choice of evaluation framework alters
the relative ranking of classifiers. We compute four scalar summary measures
on the Emotions test set: macro-$F_1$ (Discrimination Framework),
$1 - \mathrm{BS}^{\mathrm{macro}}$ (Calibration Framework, inverted so that
higher is better), macro-$\mathrm{nDCG}_j^{\mathrm{class}}$ (Ranking Framework,
row-dependent), and macro-$\mathrm{AUC}$ (Ranking Framework, column-dependent).
Figure~\ref{fig:exp1} shows the resulting scores and rank heatmap.

The heatmap reveals that no two metrics produce the same classifier ordering.
Kendall's $\tau$ values between the four metric-induced rankings are all strictly less
than 1, confirming pairwise disagreement. MLP (cal.) achieves the
best Brier score, as expected, since Platt scaling directly optimises calibration, and
is also first in $F_1$ and macro-AUC, but it is second in nDCG ($0.7339$ against $0.7343$
for the uncalibrated MLP). Below the top the orderings diverge strongly: RF is second in
Brier but fourth in macro-AUC and, with the implementation's tie rule, last in nDCG; GBT is last in Brier but third in
macro-AUC; LR is last in $F_1$ and in macro-AUC but third in nDCG. This disagreement is expected:
the three frameworks measure distinct classifier behaviours (coincidence, deviation, and
positional quality, Section~\ref{sec:synthesis}), and a classifier that is good in one
dimension need not be good in another; the theorems explain why the measures differ, not
which classifier each one favours. Two cautions apply. The random forest's probabilities contain
ties (a tied pair in $36$ of the $178$ Emotions test rows, and repeated values in every RF column), and the
position of RF in nDCG depends on how they are broken: its macro-nDCG$^{\mathrm{class}}$ is $0.7158$ with the
implementation's rule, $0.7211$ under the min-rank convention of Definition~\ref{def:rank} (where GBT, not RF,
is last) and $0.7172$ under random tie-breaking; counting ties as $0$ instead of $\tfrac12$ in the
concordance fraction lowers its macro-AUC from $0.8367$ to $0.8319$. And the margins in this single split
are small (nDCG $0.7339$ against $0.7343$), so only the disagreement among the four orderings, not
the individual positions, should be read from the figure.
Practitioners who report only one metric risk drawing conclusions that would be
reversed by an equally legitimate choice from a different framework.

% ============================================================
\subsection{Calibration Collapse vs. Ranking Non-Collapse}
\label{sec:exp2}
% ============================================================

\begin{figure}[!t]
  \centering
  \includegraphics[width=\linewidth]{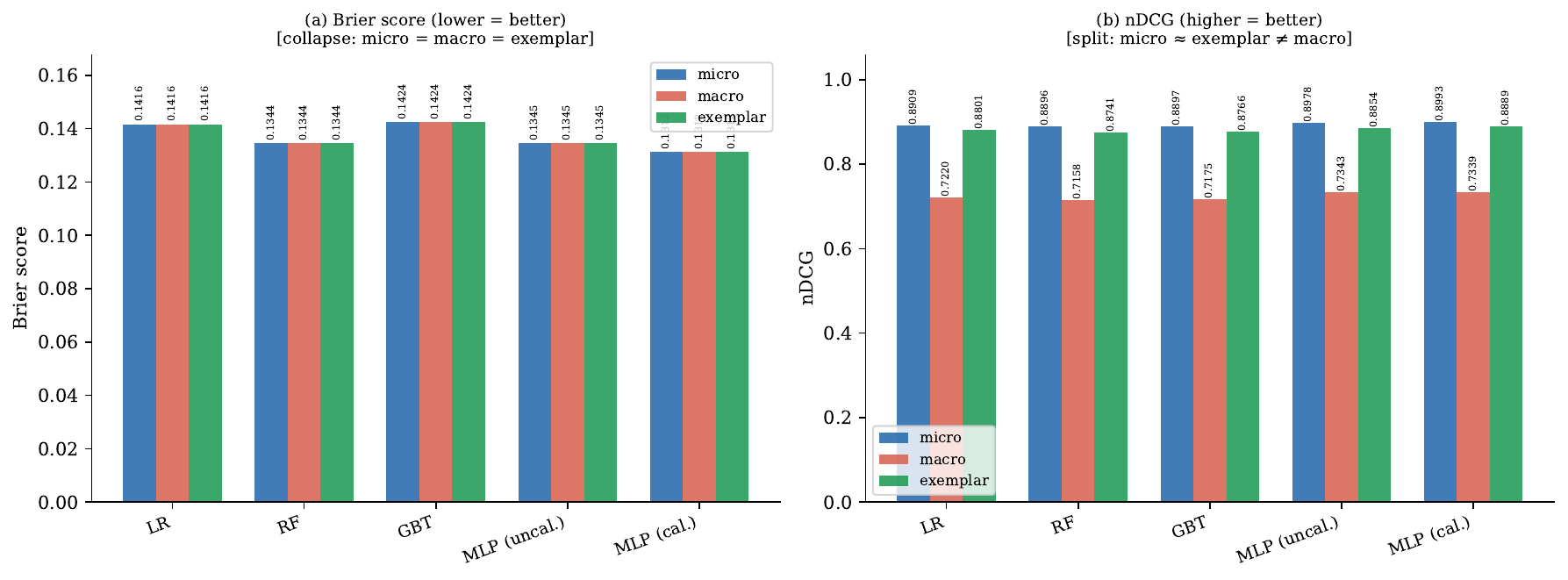}
  \caption{Experiment~2 (Emotions). (a) Brier score aggregations (micro, macro,
    exemplar) are numerically identical, confirming the collapse theorem.
    (b) nDCG aggregations: macro-class nDCG is well separated from the other two
    ($\approx0.72$ against $\approx0.88$--$0.90$) and micro and exemplar differ slightly,
    confirming non-collapse for normalised ranking measures.}
  \label{fig:exp2}
\end{figure}

The collapse theorem and its ranking-side failure make a joint prediction, tested
here on the Emotions dataset: $\mathrm{BS}^{\mathrm{micro}} =
\mathrm{BS}^{\mathrm{macro}} = \mathrm{BS}^{\mathrm{exemplar}}$
(Theorem~\ref{thm:brier_collapse}) while normalised nDCG does \emph{not} collapse
(Theorem~\ref{thm:dcg_collapse}). Figure~\ref{fig:exp2} confirms both.

Panel~(a) shows the three Brier aggregations for all five classifiers.
For each classifier, the three bars are numerically identical to at least four
decimal places, with max absolute difference below $10^{-14}$, within
floating-point rounding. Panel~(b) shows the three nDCG aggregations (micro,
macro-class, exemplar-mean). Macro-class nDCG is well separated from the other two
for every classifier ($\approx0.72$ against $\approx0.88$--$0.90$), while micro and exemplar
nDCG differ by about $0.01$. Macro-class nDCG is consistently lower than exemplar nDCG,
largely because the class ideal $\mathrm{IDCG}_j^{\mathrm{class}} = \pi_j w(1)$
cannot be attained for all labels at once: an oracle that scores every label by its truth, with ties
among the relevant labels of an example broken uniformly at random (mean over $20$ draws), reaches a macro
$\mathrm{nDCG}^{\mathrm{class}}$ of only $0.81$ on this test set. Under the min-rank convention of
Definition~\ref{def:rank} the same tied oracle would score $1$ on every represented class
(Section~\ref{sec:per_class_ndcg}). The splitting of the ranking aggregations comes from the non-uniform
IDCG normalisation, not from class imbalance alone, as shown in Theorem~\ref{thm:dcg_collapse}(b)--(d).

The contrast between the two panels illustrates the practical consequence of the
collapse theorem: for calibration measures, choosing between micro, macro, and
exemplar aggregation is a matter of convention. For ranking
measures, the choice changes the value and should be reported explicitly.

% ============================================================
\subsection{Skew-Invariance Under Prevalence Variation}
\label{sec:exp3}
% ============================================================

\begin{figure}[!t]
  \centering
  \includegraphics[width=\linewidth]{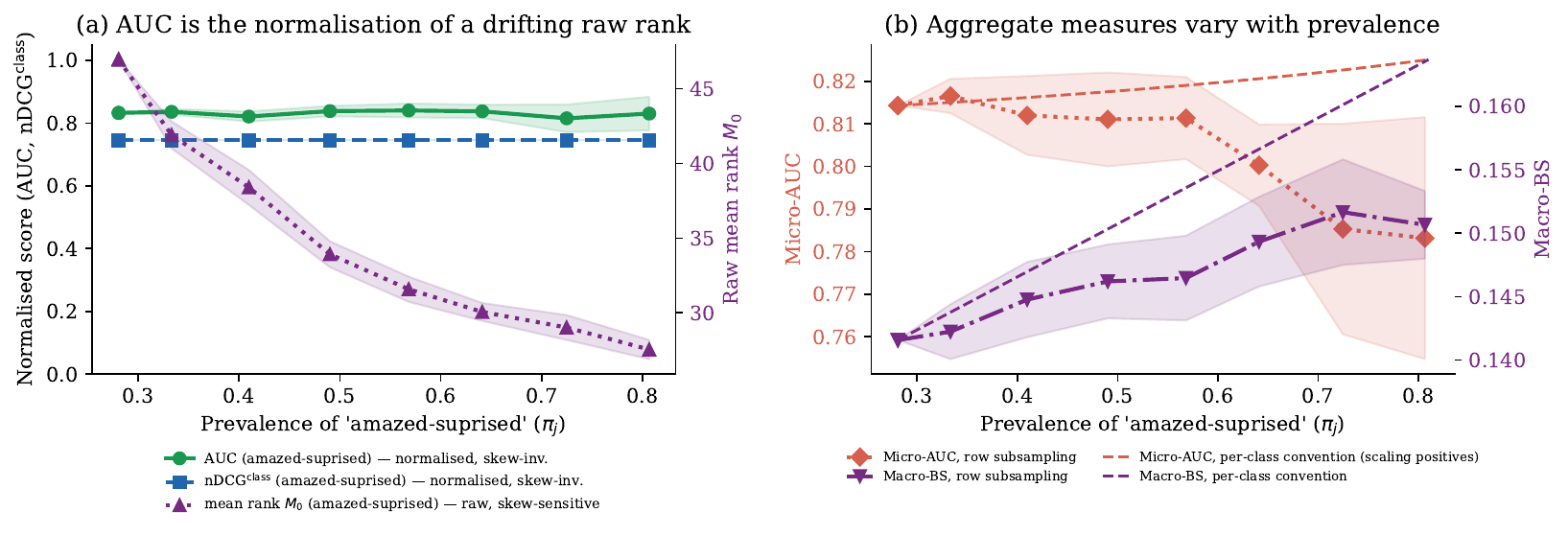}
  \caption{Experiment~3 (Emotions). Varying the prevalence of the first label
    from its natural $0.28$ up to $0.81$ (by removing negatives). (a) Per-class $\mathrm{nDCG}_j^{\mathrm{class}}$ is exactly constant in this subsampling design, whereas empirical per-class $\mathrm{AUC}_j$ remains approximately stable, with finite-sample variation (left axis),
    while the raw mean rank $M_0$ (right axis) drifts strongly; $\mathrm{AUC}_0$
    is its affine normalisation (Proposition~\ref{prop:auc_meanrank}).
    (b) Solid: micro-AUC and macro-BS under row subsampling, which couples the classes.
    Dashed: per-class-summary convention (micro-AUC by scaling the positives of class~0, macro-BS by the
    mixture formula, the other classes held fixed). Shaded bands: $\pm 1$ s.d.\ over 10 subsamples.}
  \label{fig:exp3}
\end{figure}

Propositions~\ref{prop:ndcg_class_skew} and~\ref{prop:auc_skew} establish that
$\mathrm{nDCG}_j^{\mathrm{class}}$ and $\mathrm{AUC}_j$ are skew-invariant,
while Corollary~\ref{cor:bs_skew} shows that $\mathrm{BS}^{\mathrm{macro}}$
is not and Proposition~\ref{prop:auc_skew}(c) that micro-AUC is not in general.
Figure~\ref{fig:exp3} tests these claims by varying the prevalence of the first
Emotions label (amazed-surprised, $\pi = 0.281$ in the original test set)
via subsampling of the negative examples, using logistic regression (LR) trained once on the
unchanged training set. Only negatives are removed, so the achievable
prevalence starts at the natural value (here $0.281$): the three lowest of the ten
targets $0.10,\dots,0.80$ all give $\pi=0.281$, and the sweep covers $0.28$ to $0.81$.
At each target prevalence,
the procedure is repeated ten times with different random subsamples; solid curves
show means and shaded bands show $\pm 1$ standard deviation.

Panel~(a) shows $\mathrm{AUC}_0$ (green) and $\mathrm{nDCG}_0^{\mathrm{class}}$
(blue), both trained on the unchanged training set and evaluated on the skewed
test subset. $\mathrm{nDCG}_0^{\mathrm{class}}$ is constant to machine precision, by design: only negative rows are
removed, so the row rank of every positive example is untouched, and this panel is an identity of the
design rather than an empirical test. $\mathrm{AUC}_0$ varies by $0.025$ over the sweep, with a
standard deviation up to $0.054$ at $\pi=0.8$.
Superimposed on a second axis is the \emph{raw} mean rank $M_0$ of the positive
examples (purple, right axis of panel~(a)), which is the unnormalised fixed-weight
column operator of Proposition~\ref{prop:auc_meanrank}. It drifts strongly with
prevalence (as negatives are removed, the positives rise in rank), yet its
affine Wilcoxon--Mann--Whitney normalisation, $\mathrm{AUC}_0$, is approximately stable across this finite-subsample sweep (with the nonzero variation and standard deviation reported above). Under ideal per-class reweightings that preserve the positive and negative conditional score distributions, AUC is exactly skew-invariant; finite negative subsampling only approximates that invariance. This distinction parallels the contrast between normalised and raw DCG. Ranking itself is not intrinsically prevalence-free. We verified the identity $\mathrm{AUC}_0 = (n - (n_0^+-1)/2 -
M_0)/n_0^-$ numerically at every subsample; the maximum deviation over all
trials was $2.2\times 10^{-16}$, at the level of floating-point round-off.
The same raw-to-normalised relationship holds on the row axis: the flat
$\mathrm{nDCG}_0^{\mathrm{class}}$ curve is the ideal-DCG normalisation of a raw
DCG that itself scales as $1/n$ with prevalence, the normalisation cancelling
the $n$, since $\mathrm{nDCG}_j^{\mathrm{class}} =
\tfrac{1}{w(1)}\,\mathbb{E}[w(\mathrm{rank}_i(j)) \mid Y_{ij}=1]$ does not
depend on $n$. Panel~(a) thus exhibits both positional axes obeying the same
pattern: a fixed-weight operator followed by a normalisation that removes the
prevalence dependence.
Panel~(b) shows micro-AUC (left axis) and macro-BS (right axis) under two schemes. The solid curves
are the row subsampling just described: micro-AUC falls from $0.814$ to $0.783$ (with a small initial rise
near $\pi=0.33$) and macro-BS rises from $0.142$ to about $0.152$, peaking near $\pi=0.72$. These curves are
confounded: removing negative rows of class~0 also removes positives and changes the negatives of the other
classes, so the other classes' summaries move too (in a separate diagnostic with $100$ subsamples at
$\pi\approx0.8$, the numbers of positives of the other five classes fall from $(55,84,34,50,60)$ to
$(22,11,3,5,33)$ and their AUCs change, for instance $0.63\to0.55$ and $0.91\to0.84$). The dashed curves use the
per-class-summary convention of Definition~\ref{def:cal_skew}, in which only the counts of class~0
change and the other classes' summaries are held at their full-test-set values. Scaling the negatives
leaves micro-AUC exactly unchanged, since $n_0^{+}$ and every $\mathrm{AUC}_l$ are untouched; scaling the
positives moves it towards $\mathrm{AUC}_0=0.83$, from $0.814$ to $0.825$ over the same prevalence range
(Proposition~\ref{prop:auc_skew}(c)); and macro-BS follows the mixture
$\mathrm{BS}_0=\pi\,\alpha_0+(1-\pi)\,\beta_0$ of Proposition~\ref{prop:bs_decomp} with the other classes
fixed, rising from $0.142$ to $0.164$. The rise of $\mathrm{BS}_0$ is not explained by
$\mathrm{UNC}_0=\pi(1-\pi)$, which peaks at $\pi=0.5$: it is the weight of the mixture moving to
$\alpha_0$, consistent with $\alpha_0>\beta_0$ ($0.32$ against $0.07$ on the full test set). The falling
micro-AUC of the subsampling curve is therefore a side effect of the coupling between classes, not the
effect described by Proposition~\ref{prop:auc_skew}(c). This experiment illustrates the mixture
mechanism for the Brier score and the stability of $(\mathrm{AUC}_j,\mathrm{nDCG}_j^{\mathrm{class}})$
under the subsampling, and shows that row subsampling couples the classes.

% ============================================================
\subsection{Properness, or the Limits of Order-Based Evaluation}
\label{sec:exp_properness}
% ============================================================

Proposition~\ref{prop:disc_improper} establishes weak properness for the raw positional operator among distinct-entry reports when the true marginals are also distinct. With tied reports scored by min-rank, weak properness can fail, although constant discounts are an exception. Remark~\ref{rem:ndcg_not_proper} and Proposition~\ref{prop:auc_improper} show that normalised nDCG and AUC are not universally proper. Nevertheless, any strictly order-preserving distortion of predicted probabilities leaves their realised rankings, and hence their nDCG and AUC values, unchanged; this demonstrates non-strictness, not universal properness. Theorem~\ref{thm:brier_proper} states
the opposite for the Brier score, and Remark~\ref{rem:mae_nonproper} that MAE, despite
belonging to the same structural family, is not proper at all. This experiment shows all three
claims on the same real predictions, using the LR classifier's predicted
probabilities $\hat{P}$ on the Emotions test set, and is the last of the Emotions-based
experiments.

For a sweep of exponents $\gamma \in [0.2, 5]$, every entry is distorted as
$\hat{P} \mapsto \hat{P}^{\gamma}$. Since $x \mapsto x^{\gamma}$ is strictly increasing on
$[0,1]$ for every $\gamma > 0$, this transformation preserves the rank of every entry within
any row and within any column, so it is an order-preserving distortion of the kind
Propositions~\ref{prop:disc_improper} and~\ref{prop:auc_improper} describe; for every $0<p<1$, exponents $0<\gamma<1$ raise the score ($p^\gamma>p$), whereas $\gamma>1$ lower it ($p^\gamma<p$). This one-sided power map does not generally move all scores toward $0.5$ or sharpen them toward both endpoints. The choice $\gamma=1$
recovers the untouched, classifier-reported probabilities.

\begin{figure}[!t]
  \centering
  \includegraphics[width=\linewidth]{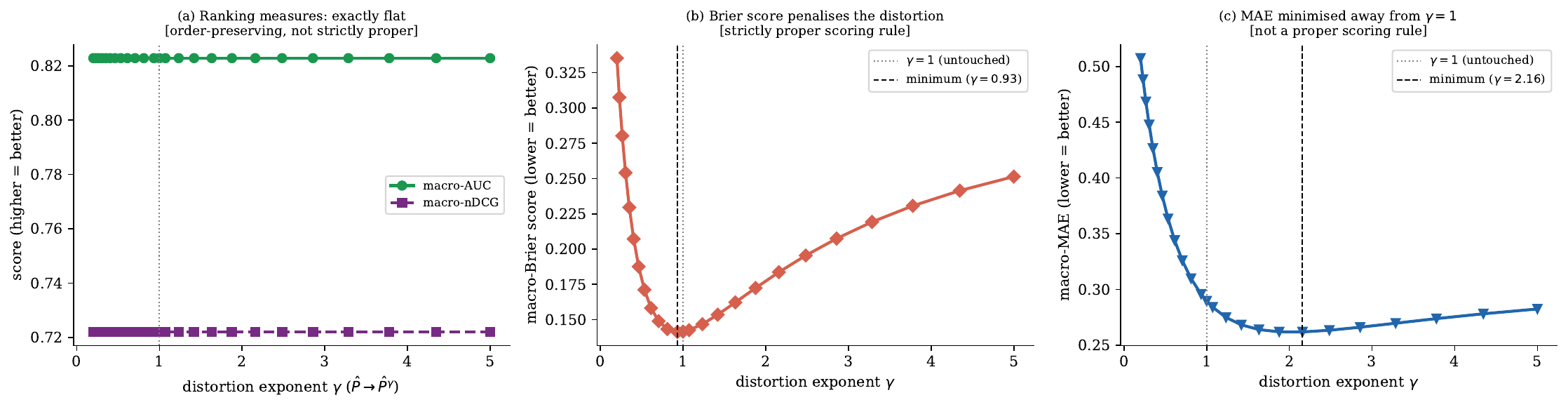}
  \caption{Experiment~4 (Emotions, classifier LR). Predicted probabilities distorted as
    $\hat{P}\to\hat{P}^{\gamma}$, an order-preserving transformation for every $\gamma>0$.
    (a)~macro-AUC and macro-nDCG$^{\mathrm{class}}$ are exactly flat (spread $0.00$) across the
    entire sweep. (b)~Macro-Brier score is minimised near $\gamma=1$ (the untouched
    probabilities), consistent with a strictly proper scoring rule applied to nearly calibrated
    scores. (c)~Macro-MAE is minimised at $\gamma=2.16$, well away from $\gamma=1$, consistent
    with Remark~\ref{rem:mae_nonproper}: MAE is not a proper scoring rule, and here it favours the
    downward-shifted scores of the power map ($\hat{P}^{\gamma}<\hat{P}$ for $\gamma>1$).}
  \label{fig:exp_properness}
\end{figure}

Figure~\ref{fig:exp_properness} confirms the three claims. Panel~(a): macro-AUC
and macro-nDCG$^{\mathrm{class}}$ are constant across the entire sweep to machine precision (a
spread of exactly $0.00$ over 25 values of $\gamma$): rank-based measures cannot
distinguish the untouched classifier from any monotonic distortion of it, whether
overconfident or not. Panel~(b): the macro-Brier score traces an asymmetric curve with its
minimum at $\gamma=0.93$, value $0.1413$, indistinguishable in practice from the untouched
$\gamma=1$ value of $0.1416$. This is consistent with Theorem~\ref{thm:brier_proper} given that LR's raw
output is, as expected, close to but not perfectly calibrated. Panel~(c): macro-MAE behaves
differently. Its minimum sits at $\gamma=2.16$, value $0.2618$, a
substantial improvement over the untouched value of $0.2895$ at $\gamma=1$. MAE
favours the downward-shifted scores produced by this particular power transformation, rather than necessarily sharper or more extreme probabilities. This is qualitatively compatible with the conditional-median target of MAE (Remark~\ref{rem:mae_nonproper}), but the experiment does not establish that the raw scores are the true conditional probabilities. On a single real
dataset, the three panels show that ranking measures cannot see calibration at
all, the Brier score rewards honesty, and structural membership in the Calibration Framework
(MAE satisfies the same aggregation collapse as Brier) is not sufficient for properness.

% ============================================================
\subsection{Per-Class Murphy Decomposition}
\label{sec:exp_murphy_class}
% ============================================================

\begin{figure}[!t]
  \centering
  \includegraphics[width=\linewidth]{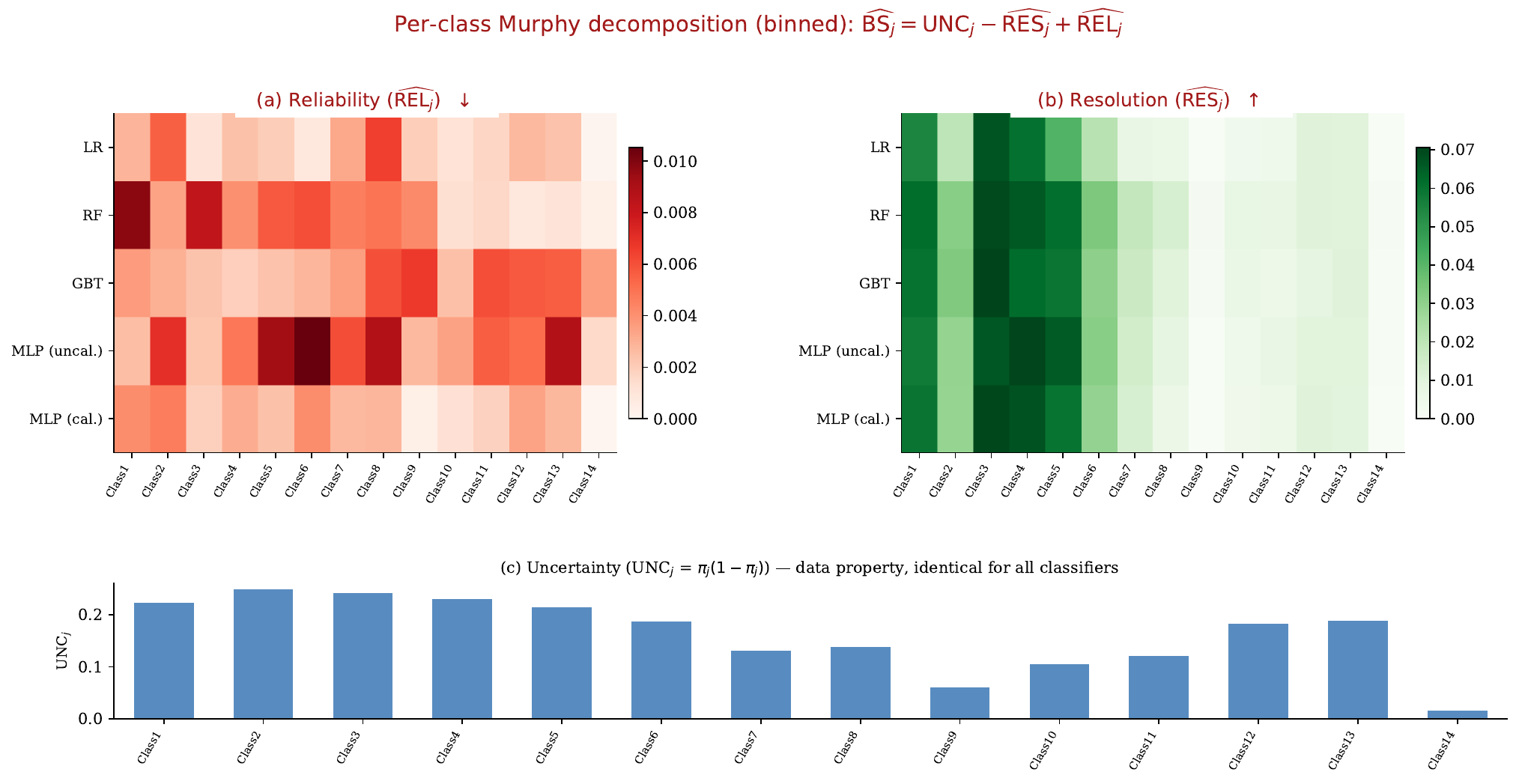}
  \caption{Experiment~5 (Yeast, 14 labels, 5 classifiers). Per-class binned Murphy
    decomposition $\widehat{\mathrm{BS}}_j=\mathrm{UNC}_j-\widehat{\mathrm{RES}}_j+\widehat{\mathrm{REL}}_j$.
    Top row: $\widehat{\mathrm{REL}}_j$ (miscalibration, Reds) and
    $\widehat{\mathrm{RES}}_j$ (resolution, Greens) heatmaps. Bottom: $\mathrm{UNC}_j$
    bar chart, identical across classifiers (data property only).}
  \label{fig:exp_murphy_class}
\end{figure}

The next two experiments move to Yeast for a finer-grained diagnosis, starting with the
column axis: Theorem~\ref{thm:murphy_col} decomposes the per-class Brier score exactly as
$\mathrm{BS}_j = \mathrm{UNC}_j - \mathrm{RES}_j + \mathrm{REL}_j$ when $f_j$
is evaluated at its exact argument; in practice $\mathrm{REL}_j$ and
$\mathrm{RES}_j$ are computed from $10$ equal-width calibration bins, so the
figure displays the binned decomposition of Proposition~\ref{prop:murphy_binned}
($\widehat{\mathrm{BS}}_j = \mathrm{UNC}_j - \widehat{\mathrm{RES}}_j +
\widehat{\mathrm{REL}}_j$, exactly, for the discretised predictor) rather than
Theorem~\ref{thm:murphy_col}'s pointwise version, which degenerates to
$\mathrm{RES}_j=\mathrm{UNC}_j$, $\mathrm{REL}_j=\mathrm{BS}_j$ on continuous
scores with no repeated values. Figure~\ref{fig:exp_murphy_class} displays this binned
decomposition for all five classifiers across the 14 Yeast labels.

The figure has three panels. The top-left heatmap shows $\widehat{\mathrm{REL}}_j$
(binned reliability, lower is better) for each classifier--label pair; darker red
indicates higher miscalibration. The top-right heatmap shows $\widehat{\mathrm{RES}}_j$
(binned resolution, higher is better); darker green indicates more discriminative
predictions. The bottom bar chart shows $\mathrm{UNC}_j = \pi_j(1-\pi_j)$,
which is identical across all classifiers by definition (it is a property
of the data, not the model) and is annotated accordingly.

Several patterns are consistent with theory. First, MLP (cal.) shows
lower REL$_j$ than MLP (uncal.) on 13 of the 14 labels, consistent with Platt scaling
reducing miscalibration as measured by the $\mathrm{REL}$ term while leaving
$\mathrm{UNC}_j$ unchanged. Second, GBT and RF exhibit higher RES$_j$ than LR on
13 of the 14 labels, reflecting their stronger discriminative power.
Third, RES$_j$ is strongly associated with UNC$_j$ across labels (Spearman $0.87$--$0.91$ for every
classifier), as expected since $\mathrm{RES}_j\le\mathrm{UNC}_j$; REL$_j$ shows no consistent association
with UNC$_j$ (Spearman between $-0.52$ and $0.71$ depending on the classifier). The dependence of RES on
prevalence is consistent with Corollary~\ref{cor:murphy_skew}(b); a comparison across labels with different prevalences is,
however, not a controlled test of that corollary; such a test would vary the prevalence of a single
label, as Experiment~3 does for the Brier mixture, and is not carried out for $\mathrm{RES}_j$ here.

The binned decomposition depends on the number of bins and, on continuous scores, replaces each score by a bin mean.
Table~\ref{tab:exp_isotonic} repeats it with the isotonic calibration map of Proposition~\ref{prop:murphy_isotonic}, which needs no bins, on the same
classifiers and labels. The identity $\mathrm{BS}_j=\mathrm{UNC}_j-\mathrm{RES}_j^{\mathrm{iso}}+\mathrm{REL}_j^{\mathrm{iso}}$ holds to rounding error
on all 70 classifier--label pairs, with $\mathrm{BS}_j$ the Brier score of the original scores, not of a discretised predictor. The two resolution terms are almost
the same ranking of the labels (Spearman $0.99$--$1.00$ for every classifier), and the isotonic one is somewhat larger on average. The reliability terms differ more: the isotonic $\mathrm{REL}_j^{\mathrm{iso}}$ is larger than
the binned one for every classifier, because the binned predictor absorbs the within-bin variation of the scores (Remark~\ref{rem:binning_gap}), and
the rank correlation across labels falls to $0.22$--$0.92$ depending on the classifier. Two of the observations above change with the estimator. MLP (cal.) has lower reliability error than
MLP (uncal.) on $12$ of the $14$ labels instead of $13$, and GBT has higher resolution than LR on $10$ of the $14$ labels instead of $13$ (RF, $13$ of $14$ in both).
The mean reliability error of RF and GBT is close under both estimators and swaps order ($0.0039$ and $0.0040$ binned, $0.0067$ and $0.0063$ isotonic), so we draw no
conclusion about their order. The two main findings, that Platt scaling lowers the reliability error and that the tree ensembles resolve more than the
logistic regression, hold under both estimators. The correlation of $\mathrm{RES}_j^{\mathrm{iso}}$ with $\mathrm{UNC}_j$ across labels is $0.88$--$0.92$, as for
the binned terms, and that of $\mathrm{REL}_j^{\mathrm{iso}}$ with $\mathrm{UNC}_j$ is positive ($0.38$--$0.87$), whereas it had no consistent sign for the binned terms.

\begin{table}[!t]
\centering
\small
\begin{tabular}{@{}lrrrr@{}}
\toprule
& \multicolumn{2}{c}{Reliability (lower is better)} & \multicolumn{2}{c}{Resolution (higher is better)}\\
\cmidrule(lr){2-3}\cmidrule(lr){4-5}
Classifier & binned & isotonic & binned & isotonic \\
\midrule
LR            & 0.0024 & 0.0053 & 0.0217 & 0.0255 \\
RF            & 0.0039 & 0.0067 & 0.0277 & 0.0317 \\
GBT           & 0.0040 & 0.0063 & 0.0265 & 0.0293 \\
MLP (uncal.)  & 0.0056 & 0.0081 & 0.0266 & 0.0295 \\
MLP (cal.)    & 0.0025 & 0.0056 & 0.0257 & 0.0297 \\
\bottomrule
\end{tabular}
\caption{Experiment~5 (Yeast), binned (10 equal-width bins) and isotonic Murphy terms, averaged over the 14 labels.
The isotonic terms satisfy $\mathrm{BS}_j=\mathrm{UNC}_j-\mathrm{RES}_j^{\mathrm{iso}}+\mathrm{REL}_j^{\mathrm{iso}}$ for the original scores.}
\label{tab:exp_isotonic}
\end{table}

% ============================================================
\subsection{Per-Exemplar Brier Decomposition}
\label{sec:exp_murphy_example}
% ============================================================

\begin{figure}[!t]
  \centering
  \includegraphics[width=\linewidth]{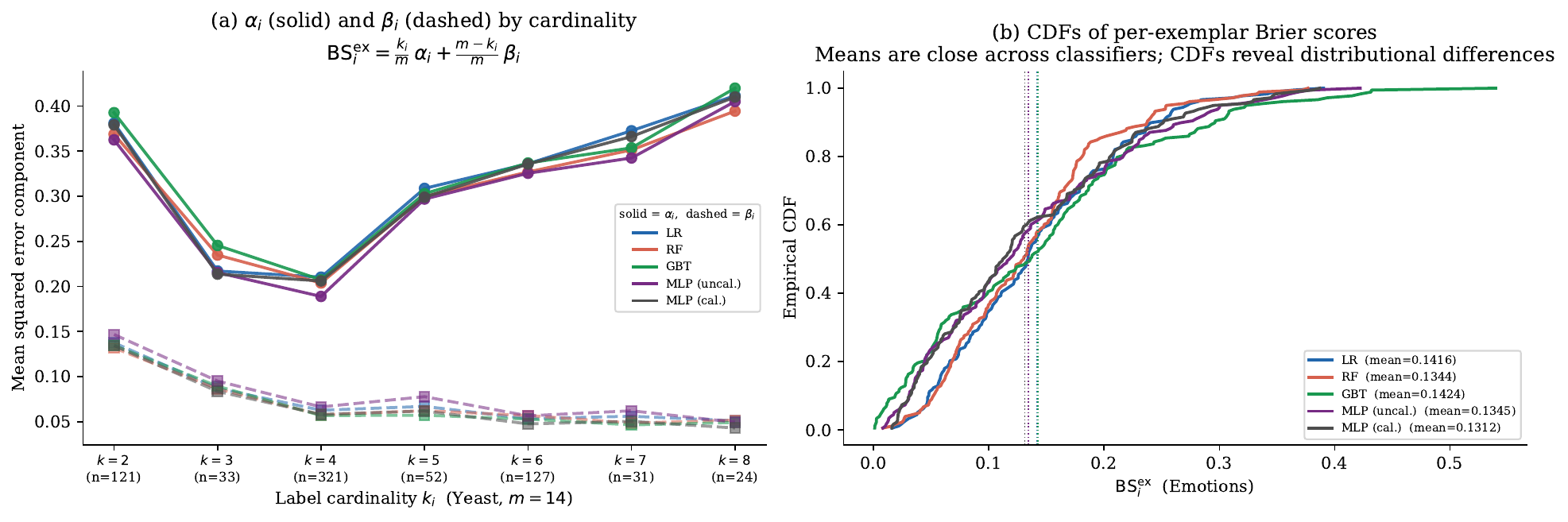}
  \caption{Experiment~6. (a) Mean $\alpha_i$ (solid) and $\beta_i$ (dashed)
    by label cardinality $k_i$ on Yeast; only bins with $\geq 15$ examples
    are shown. (b) Empirical CDFs of $\mathrm{BS}_i^{\mathrm{example}}$ on Emotions;
    dotted verticals mark each classifier's mean. The means are close
    ($0.131$--$0.142$) but the CDFs differ.}
  \label{fig:exp_murphy_example}
\end{figure}

Section~\ref{sec:three_murphy} introduces the row-axis counterpart, which decomposes the
per-exemplar Brier score
$\mathrm{BS}_i^{\mathrm{example}} = (k_i/m)\,\alpha_i + ((m-k_i)/m)\,\beta_i$
where $\alpha_i$ is the miss-calibration on example $i$'s positive labels and
$\beta_i$ is the false-alarm calibration on its negative labels (Proposition~\ref{prop:bs_decomp_row}).
Figure~\ref{fig:exp_murphy_example} provides two complementary views of this decomposition.

Panel~(a) plots $\alpha_i$ and $\beta_i$ by label cardinality on
Yeast, whose $m=14$ labels yield cardinalities $k_i$ from 2 to 8 with
adequate support (only bins with $\geq 15$ examples are shown; the sizes are
$121, 33, 321, 52, 127, 31, 24$ for $k_i=2,\dots,8$).
For each classifier, solid lines trace mean $\alpha_i$ and dashed lines trace mean
$\beta_i$ as a function of $k_i$. Two patterns emerge. First,
$\beta_i$ (false-alarm calibration) is low and broadly decreases with $k_i$ for all
classifiers, from about $0.13$--$0.15$ at $k_i=2$ to about $0.05$ at $k_i=8$:
examples with many relevant labels have fewer negative labels, and the classifier
assigns them low probability. Second, $\alpha_i$ (miss-calibration on positives) is
higher than $\beta_i$ at every cardinality and is not monotone in $k_i$: for LR it is
$0.38$ at $k_i=2$, falls to $0.21$ at $k_i=4$, and rises again to $0.41$ at $k_i=8$.
The two classifiers that differ only in post-hoc calibration, MLP (cal.) and MLP (uncal.),
differ mainly in $\beta_i$: Platt scaling lowers $\beta_i$ at every cardinality (for
example $0.134$ against $0.147$ at $k_i=2$), while $\alpha_i$ is slightly \emph{higher}
after calibration at six of the seven cardinalities (for example $0.379$ against
$0.363$ at $k_i=2$), so the effect of calibration at the exemplar level is a trade between
the positive and the negative side.

Panel~(b) shows empirical CDFs of $\mathrm{BS}_i^{\mathrm{example}}$ on
Emotions. The collapse theorem (Theorem~\ref{thm:brier_collapse}) guarantees that
$\frac{1}{n}\sum_i \mathrm{BS}_i^{\mathrm{example}}$ is the same for all
aggregation axes. Panel~(b) tests whether this equality of means conceals
distributional heterogeneity. The five empirical CDFs of
$\mathrm{BS}_i^{\mathrm{example}}$ are shown with vertical dotted lines marking the
classifiers' mean values.

The means are close across classifiers ($0.131$ to $0.142$; the collapse theorem
concerns the aggregations of a single classifier, not differences between classifiers), but the
CDFs differ. GBT has both the most mass near zero (5\% quantile $0.008$, against $0.040$ for LR
and RF) and the heaviest tail (95\% quantile $0.33$, maximum $0.54$); RF has the smallest
maximum ($0.38$) and MLP (cal.) the smallest median ($0.113$, against $0.134$ for LR). This divergence is invisible to any scalar Brier-score summary and can only be
revealed by examining the row-level distribution of $\mathrm{BS}_i^{\mathrm{example}}$.
The experiment therefore provides empirical support for the claim in
Remark~\ref{rem:cal_disc}: the row-level view, through the per-example errors
$(\alpha_i,\beta_i)$ and the distribution of $\mathrm{BS}_i^{\mathrm{example}}$, exposes
per-example calibration structure that the per-class summaries and the scalar Brier score
discard.

% ============================================================
\subsection{Per-Class nDCG as a Novel Measure}
\label{sec:exp_ndcg_novel}
% ============================================================

Proposition~\ref{prop:ndcg_class_mean} and Theorem~\ref{thm:duality} establish
that $\mathrm{nDCG}_j^{\mathrm{class}}$ and $\mathrm{AUC}_j$ are both
skew-invariant conditional means over positive examples, but along orthogonal
axes: $\mathrm{nDCG}_j^{\mathrm{class}}$ evaluates the \emph{row rank} of label
$j$ within each positive example's predicted label vector, while $\mathrm{AUC}_j$
evaluates the \emph{column rank} of positive examples above negatives for class
$j$. This closing experiment asks whether the two measures agree in practice on
the same Yeast labels just decomposed two ways above, and if not, what drives
their disagreement.

\begin{figure}[!t]
  \centering
  \includegraphics[width=\linewidth]{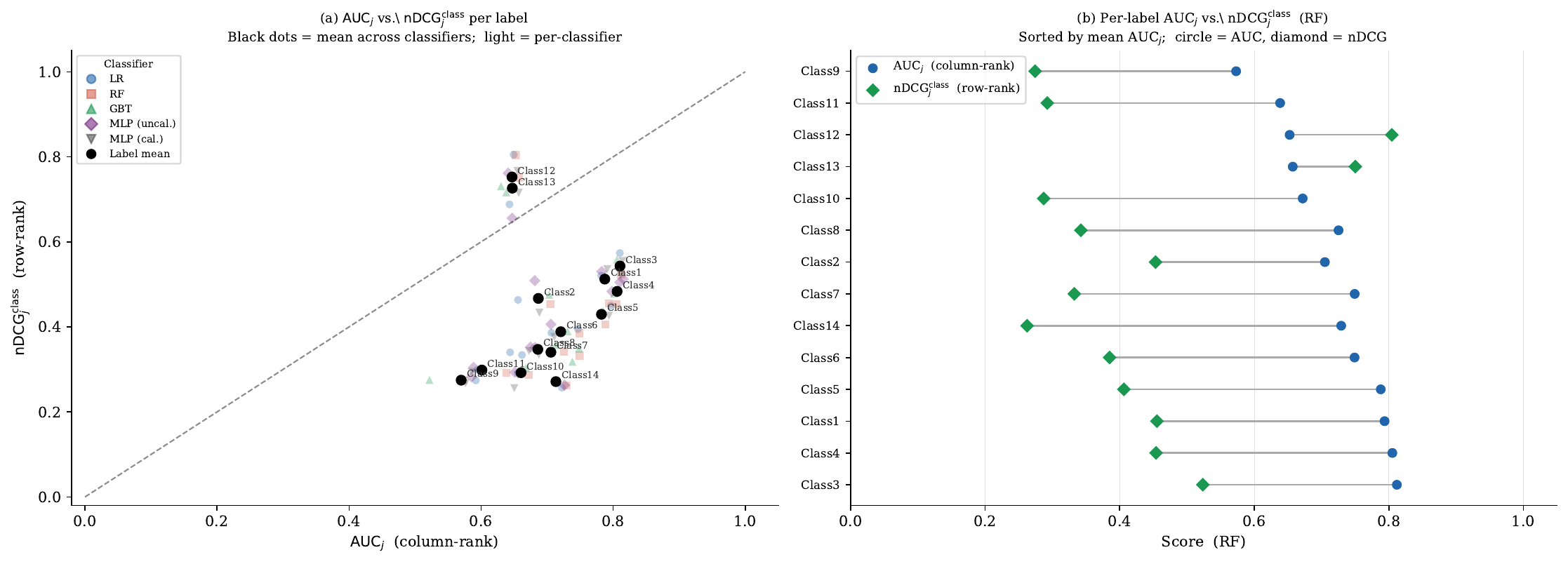}
  \caption{Experiment~7 (Yeast, $m=14$). (a) Scatter of label-mean
    $\mathrm{AUC}_j$ vs.\ $\mathrm{nDCG}_j^{\mathrm{class}}$, annotated with
    label names; light dots show per-classifier spread. Most labels fall below
    the diagonal $y=x$; Class12 and Class13, the two most prevalent labels, are exceptions
    ($\mathrm{nDCG}_j^{\mathrm{class}} > \mathrm{AUC}_j$).
    (b) Dumbbell plot for RF: circle = $\mathrm{AUC}_j$ (column-rank),
    diamond = $\mathrm{nDCG}_j^{\mathrm{class}}$ (row-rank), sorted by mean
    $\mathrm{AUC}_j$; line length = row-rank deficit.}
  \label{fig:exp_ndcg_novel}
\end{figure}

Figure~\ref{fig:exp_ndcg_novel} uses the Yeast dataset ($m = 14$ labels).
For all five classifiers and all 14 labels we compute $\mathrm{AUC}_j$ and
$\mathrm{nDCG}_j^{\mathrm{class}}$.

Panel~(a) shows the scatter of per-label means (black dots, annotated with label
names) against the diagonal $y = x$, with light per-classifier dots in the
background showing within-label spread across classifiers. Twelve of the fourteen label means fall
below the diagonal ($\mathrm{AUC}_j > \mathrm{nDCG}_j^{\mathrm{class}}$, mean gap $0.32$ over
these twelve labels and $0.26$ over all fourteen),
and two, Class12 and Class13, sit above it. These two are
by far the most prevalent labels in the test set ($\pi=0.76$ and $0.75$, against
at most $0.45$ for every other label). Across the 14 labels, $\mathrm{nDCG}_j^{\mathrm{class}}$ is
almost a monotone function of prevalence (Spearman $\rho=0.97$), whereas $\mathrm{AUC}_j$ is not
($\rho=0.22$), so the gap $\mathrm{AUC}_j-\mathrm{nDCG}_j^{\mathrm{class}}$ decreases with
prevalence ($\rho=-0.78$). At the other end, the four least prevalent labels
(Class14, Class9, Class10, Class11, $\pi\le0.14$) have $\mathrm{nDCG}_j^{\mathrm{class}}$ between
$0.27$ and $0.30$, below the $0.40$ that a uniformly random ordering of $14$ labels would give
(Section~\ref{sec:per_class_ndcg}; the ceiling for an oracle that scores every label by its truth, with random tie-breaking among
relevant labels as above, is $0.61$ on Yeast, itself increasing with prevalence, Spearman $0.49$, and dividing $\mathrm{nDCG}_j^{\mathrm{class}}$
by its oracle value leaves the Spearman correlation with prevalence at $0.97$), while their $\mathrm{AUC}_j$ ranges from $0.57$ to $0.71$ (Class14: $0.71$ and
$0.27$). Two labels with similar $\mathrm{AUC}_j$ can therefore have very different
$\mathrm{nDCG}_j^{\mathrm{class}}$ (Class2 and Class8, both $\approx0.69$, have $0.47$ and $0.35$;
Class13 and Class11, $0.65$ and $0.60$, have $0.73$ and $0.30$).

Panel~(b) shows a dumbbell plot for RF: each label has a circle
($\mathrm{AUC}_j$, column-rank) and a diamond ($\mathrm{nDCG}_j^{\mathrm{class}}$,
row-rank) connected by a line, listed by increasing $\mathrm{AUC}_j$ averaged over the classifiers
(top to bottom), so the two panels are directly comparable.
The line length is the row-rank deficit $\mathrm{AUC}_j-\mathrm{nDCG}_j^{\mathrm{class}}$, which for RF
is between $0.25$ and $0.47$ for twelve labels, largest for Class14 ($0.47$) and Class7 ($0.42$),
and negative for Class12 ($-0.15$) and Class13 ($-0.09$). The ordering of labels by $\mathrm{AUC}_j$
and by $\mathrm{nDCG}_j^{\mathrm{class}}$ differs: Class9 is last by $\mathrm{AUC}_j$ ($0.57$) and
also among the last by nDCG, but Class12 and Class13, which are 11th and 12th of 14 by
$\mathrm{AUC}_j$, are first and second by $\mathrm{nDCG}_j^{\mathrm{class}}$.

This is not in conflict with the skew-invariance of $\mathrm{nDCG}_j^{\mathrm{class}}$
(Proposition~\ref{prop:ndcg_class_skew}), which concerns a fixed classifier whose scores are
unchanged when the examples of one class are replicated; the experiment compares different
labels. The row rank of label $j$ is decided by the competition with the other labels' scores in
the same example, and a classifier fitted to the data produces scores whose scale follows each
label's base rate: a frequent label tends to receive high scores and often tops the row, a rare
one receives low scores and rarely does, whatever its separation of positives from negatives.
Two interventions test this explanation directly.

The first intervention is a prior-only predictor: give label $j$ the same score $\pi_j$ (the training prevalence)
for every example. Its $\mathrm{AUC}_j$ is $0.5$ by construction, so it carries the scale of the
scores and nothing else. Its $\mathrm{nDCG}_j^{\mathrm{class}}$ equals one plus the number of more
prevalent labels, put through the weight, so it is a monotone function of prevalence (Spearman
$\rho=1.00$) and averages to $0.40$, the value for a random ordering. It already tracks the
classifiers: across labels, the Spearman correlation between the two is $0.97$ (Pearson $0.89$,
$R^2=0.79$), the mean absolute difference is $0.08$, and the classifiers add on average only $0.04$
to the scale-only baseline ($0.44$ against $0.40$).

The second intervention is a column-wise normalisation: replace each column of scores by its mid-rank percentile.
This order-preserving, tie-preserving transform leaves every $\mathrm{AUC}_j$ exactly unchanged (the
largest change in the run is $0$). It places all labels on a common percentile scale and yields identical empirical percentile distributions when their scores have no ties, or tie groups of the same sizes at the same rank positions; with differing patterns of ties, mid-ranks need not produce identical distributions. The
Spearman correlation of $\mathrm{nDCG}_j^{\mathrm{class}}$ with prevalence falls from $0.97$ to
$-0.23$, and its correlation with $\mathrm{AUC}_j$ rises from $0.30$ to $0.86$ (Spearman; Pearson
$0.19$ to $0.69$). The normalisation is applied to each classifier and averaged over $20$ random
tie-breaks between labels. Both results are what the explanation predicts, and a fixed decision
rule (a normalised correlation with prevalence below $0.3$ in absolute value) was set before running
them. Caveats: one data set, one split and five classifiers; Class14 has about a dozen test
positives, so its values are noisy.

It follows that, across labels, $\mathrm{nDCG}_j^{\mathrm{class}}$ mixes the quality of the
label's score with the scale of its scores relative to those of the other labels, which is largely
set by the base rate; $\mathrm{AUC}_j$ compares only the scores of one label across examples, so it
is not affected by what the other labels receive. The two measures are therefore complementary, and
$\mathrm{nDCG}_j^{\mathrm{class}}$ should not be read as a prevalence-free measure of ranking
quality when comparing different labels; a complete ranking evaluation should report both.

% ============================================================
%  Section 7 — Conclusion
% ============================================================
% ============================================================
%  Section 7: Conclusion
%  Draft for: "A Generalised Confusion Matrix Framework"
%  Author: Ronaldo C. Prati
% ============================================================

\section{Conclusion}
\label{sec:conclusion}

This paper extended the indicator-matrix framework of
\citet{prati2026unified} from discrimination to calibration and ranking,
covering the three major paradigms of probabilistic multilabel evaluation
within a single algebraic language.

Replacing the $t$-norm cell operator
with a deviation operator $\phi(y - \hat{y})$ or a positional
operator $Y_{ij} \cdot w(\mathrm{rank})$ preserves the three-aggregation
architecture of the base framework (micro, macro, and exemplar) while
producing qualitatively different collapse behaviour and skew-sensitivity
profiles. The Calibration Framework is the only operator class whose measures
collapse to the same scalar under all three aggregation schemes even after
normalisation; the Discrimination Framework's ratio measures need not collapse, and
in the Ranking Framework the raw DCG and AUC aggregates collapse while the normalised
nDCG and AUC scalars need not, because of heterogeneous denominators (the ideal DCG, the
class sizes); equality holds only under the covariance conditions of the collapse theorems. The
Calibration Framework is also the only class that supports strictly proper
scoring rules, with the Savage representation providing the exact
characterisation; raw positional operators can be weakly proper under the stated ranking assumptions, but normalised nDCG and AUC are not universally proper; none is strictly proper.

These structural differences have practical consequences. Reporting
$\mathrm{BS}^{\mathrm{macro}}$ on an imbalanced dataset mixes calibration
quality with class prevalence in a way that recall and specificity do not; the
remedy is to report the class-conditional pair $(\alpha_j, \beta_j)$, which
plays the role in the Calibration Framework that recall and specificity
play in the Discrimination Framework and that per-class AUC and nDCG play in the
Ranking Framework. The seven experiments confirmed that these theoretical
distinctions are empirically visible and non-trivial: the classifier orderings induced by the three
frameworks differ (for instance RF is second in Brier score but fourth in AUC); pooled aggregates drift
under prevalence changes while per-class measures remain stable (macro-BS follows the mixture of
$\alpha_j$ and $\beta_j$, and row subsampling couples the classes); per-class nDCG tracks label
prevalence across labels, which AUC does not; the
scalar Brier score can conceal substantial distributional heterogeneity at the
per-exemplar level; and an order-preserving distortion of a classifier's
probabilities leaves every ranking measure exactly unchanged while raising
its Brier score (beyond a small neighbourhood of the original probabilities) and, in the
opposite direction, lowering its MAE, which makes the
properness distinction concrete.

The skew-invariant per-class
building blocks in all three frameworks share the same algebraic form: each is a
class-conditional expectation $\mathbb{E}[q \mid Y_{ij}=c]$ (over the positives for recall,
per-class AUC and nDCG, and $\alpha_j$; over the negatives for specificity and $\beta_j$). This template explains why
recall, specificity, per-class AUC, $\mathrm{nDCG}_j^{\mathrm{class}}$, and
$(\alpha_j, \beta_j)$ are skew-invariant under per-class reweightings that preserve
the conditional distribution of the cell quality $q$, and why their micro-averaged
counterparts are not in general (Section~\ref{sec:synthesis} states the exact hypotheses).
Skew-invariance is thus not a special property of AUC or of any particular measure, and
aggregations that reintroduce prevalence as a weight are skew-sensitive in general.

Most results of Sections~\ref{sec:calibration} and~\ref{sec:ranking} have additionally been mechanically
checked in the Lean~4 proof assistant against Mathlib, as part of the same formalization project verifying the
base framework of \citet{prati2026unified}. Each of the twelve theorems, propositions and corollaries of
Section~\ref{sec:ranking} is covered, in the form and under the conventions listed in the supplementary
verification report: scores are taken distinct (so ranks are permutations), except in the hard-prediction
results, which use min-rank; cell matrices range over arbitrary real matrices where the paper restricts them
to the relevant cells (the nDCG impossibility argument for relevant cells only is on paper, as is the
condition of Theorem~\ref{thm:auc_collapse}(d) in its relevant-cell form); replication is modelled by
weights, or by integer multiples of per-class score multisets for the skew characterisation; and expected
scores are taken over finite label scenarios, which the bridge to an arbitrary probability space shows to be
no restriction for binary labels. The properness of AUC (Proposition~\ref{prop:auc_improper}) is checked
under the paper's half-credit convention for ties: the non-uniqueness, the maximality for independent labels
and, for two and three examples, under any dependence (in both cases also when marginals tie), and the
counterexample for every number $n\ge4$ of examples, with its exact difference; only the final perturbation
to pairwise distinct marginals in that counterexample is on paper. The
hard-prediction limit of Proposition~\ref{prop:hard_ranking} is covered in full, including the remark on a
general number of predicted positives. In Section~\ref{sec:calibration}, the checked statements are the
Brier collapse and strict properness, both Murphy decompositions, the prevalence decomposition and its
corollaries, Theorem~\ref{thm:cal_skew} in a per-class model of score multisets under integer replication,
Corollary~\ref{cor:murphy_skew} (the exact form of the resolution term, the exact condition for perfect
calibration at two prevalences, and the degenerate exception for $\mathrm{REL}_j$), the Savage representation (both directions, for continuous sections), the
characterisation of the proper symmetric deviation operators (Corollary~\ref{cor:savage_cases}(c): the
equivalence with a symmetric weight and strictness for a positive weight, with $\phi$ differentiable and the
weight's nonnegativity assumed rather than derived from convexity), and
Remark~\ref{rem:class_weighted_brier}. Not formalised are, in that section, the framework bridge
(Theorem~\ref{thm:bridge}), Proposition~\ref{prop:unc_inequality}, Proposition~\ref{prop:tnorm_improper},
the row prevalence decomposition (Proposition~\ref{prop:bs_decomp_row}), parts (a), (b) and (d) of
Corollary~\ref{cor:savage_cases} together with the uniqueness of the squared error among power losses and
the MAE item, the binned and isotonic Murphy decompositions and the QWK remark; and, throughout, the remarks other than
those named above and all experiments. The formalization and the scripts and data of all experiments are
available at \url{https://github.com/rcprati/discrimination-calibration-ranking}.

The Calibration and Ranking Frameworks are implemented in the \texttt{aggmetrics.scores} subpackage of
\texttt{aggmetrics} (\url{https://github.com/rcprati/aggmetrics}): the Brier score and other deviation operators, the class-conditional
errors, the level-set, binned and isotonic Murphy decompositions, the DCG and AUC families with explicit tie conventions, and diagnostics
that report the covariance behind each gap and the pairs of labels that nDCG and AUC order in opposite ways. The subpackage reproduces the
running example of this paper and is tested against scikit-learn where the conventions coincide.

Several directions remain open for future work. First, the framework currently handles
binary labels ($Y_{ij} \in \{0,1\}$) and soft probability outputs
$\hat{Y}_{ij} \in [0,1]$; extending it to ordinal or real-valued labels would
require a generalised notion of the calibration function $f_j$ and of the rank
function. Remark~\ref{rem:qwk} sketches the connection to quadratic weighted
kappa for ordinal classes, but a full treatment is left for future work.
Second, the Murphy decomposition derived here applies to each label
independently; capturing cross-label calibration, that is, whether the joint
probability vector $(\hat{Y}_{i,\cdot})$ matches the joint label
distribution, would require multivariate proper scoring rules
\citep{gneiting2007strictly} and a corresponding matrix-level decomposition.
Third, the per-exemplar Brier decomposition (Section~\ref{sec:three_murphy})
and per-example AUC (Definition~\ref{def:auc_family}) are novel measures whose
properties (consistency, sample complexity, finite-sample behaviour) deserve
dedicated study. Fourth, this paper focuses on the \emph{evaluation} side;
connecting the framework to \emph{learning}, specifically to loss functions
used during training, would clarify which training objectives correspond to
minimising measures in each framework and what guarantees they carry for
calibration, ranking, and discrimination simultaneously.

% ============================================================
%  Appendix
% ============================================================
% ============================================================
%  Appendix: running example
% ============================================================
\appendix

\section{The Running Example in Full}
\label{app:running}

All values below are produced by \texttt{experiments/running\_example.py} (exact fractions where the value
is rational; the nDCG family involves $\log_2 3$ and is given to three decimals). Labels are ordered
$(A,B,C)$ and examples $1$--$4$, as in Example~\ref{ex:running}.

At threshold $0.5$ the hard predictions, by rows, are $(1,1,0), (1,0,0), (1,0,1), (0,0,0)$. The per-class counts
$(tp_j,fp_j,fn_j,tn_j)$ are $(2,\ 1,\ 0,\ 1)$, $(1,\ 0,\ 2,\ 1)$, $(0,\ 1,\ 1,\ 2)$; precision is
$(\tfrac{2}{3},\ 1,\ 0)$ (micro $\tfrac{3}{5}$, macro $\tfrac{5}{9}$),
and recall is $(1,\ \tfrac{1}{3},\ 0)$ (micro $\tfrac{1}{2}$, macro $\tfrac{4}{9}$).

For calibration, the squared-deviation cells are as in Example~\ref{ex:running_cells}, with $\mathrm{BS}_j^{\mathrm{class}}=(\tfrac{21}{80},\ \tfrac{13}{40},\ \tfrac{9}{40})$,
$\mathrm{BS}_i^{\mathrm{example}}=(\tfrac{13}{150},\ \tfrac{53}{300},\ \tfrac{59}{100},\ \tfrac{23}{100})$, micro $=$ macro $=$ exemplar $=\tfrac{13}{48}$;
$\alpha=(\tfrac{1}{8},\ \tfrac{43}{100},\ \tfrac{9}{25})$, $\beta=(\tfrac{2}{5},\ \tfrac{1}{100},\ \tfrac{9}{50})$, $\pi=(\tfrac12,\tfrac34,\tfrac14)$.
The squared scores $\hat P_{ij}^2$ give Brier $\tfrac{14039}{40000}$ with all ranks unchanged.
The same-Brier companion has Brier $\tfrac{13}{48}$, $\mathrm{AUC}_j=(1,\ \tfrac{1}{3},\ 0)$, macro-AUC $\tfrac{4}{9}$.

The row ranks, by rows, are $(1,2,3), (1,3,2), (1,3,2), (1,2,3)$, with ideal masses $\mathrm{IDCG}_i=(1+w_2,\,1+w_2,\,1,\,1)$, where
$w_2=1/\log_2 3$. For raw DCG, micro $=$ class mean $=$ exemplar mean $=0.366$; further,
$\mathrm{nDCG}_j^{\mathrm{class}}=(1,\ 0.587,\ 0.631)$ (mean $0.739$);
$\mathrm{nDCG}_i=(1,\ 1,\ 0.5,\ 0.631)$ (mean $0.783$); micro-nDCG $0.835$.
The top-1 label is $A$ for every example, and $\mathrm{Precision}@1=\tfrac{1}{2}$.

On the column axis, the concordance cells are as in Example~\ref{ex:running_cells}, with $\mathrm{AUC}_j=(\tfrac{1}{2},\ 1,\ \tfrac{2}{3})$, macro $\tfrac{13}{18}$,
micro $\tfrac{7}{9}$; $\mathrm{AUC}_i^{\mathrm{ex}}=(\tfrac{3}{4},\ \tfrac{7}{12},\ 1,\ 1)$, mean $\tfrac{5}{6}$;
raw aggregate $\tfrac{7}{18}$; pooled ROC-AUC over all twelve cells $\tfrac{43}{72}$.

Under the per-class convention with the positives of $A$ doubled, the values change as follows: $\pi_A$ $\tfrac{1}{2}\to\tfrac{2}{3}$,
precision$_A$ $\tfrac{2}{3}\to\tfrac{4}{5}$, $\mathrm{BS}_A$ $\tfrac{21}{80}\to\tfrac{13}{60}$,
micro-AUC $\tfrac{7}{9}\to\tfrac{17}{24}$; recall, specificity, $(\alpha_A,\beta_A)$, $\mathrm{AUC}_A$,
$\mathrm{nDCG}_A^{\mathrm{class}}$ and macro-AUC are unchanged. With whole rows $1$ and $2$ duplicated,
$\mathrm{nDCG}_j^{\mathrm{class}}=(1,\ 0.598,\ 0.631)$, micro-nDCG $0.898$, exemplar
mean $0.855$, $\mathrm{AUC}_j=(\tfrac{1}{2},\ 1,\ \tfrac{3}{4})$, micro-AUC $\tfrac{3}{4}$.

\section{Worked Examples for Section~\ref{sec:background}}
\label{app:bg_examples}

A binary classification task can be represented either by one indicator column for a designated positive class or by two complementary one-hot columns. Although these encodings contain equivalent label information, evaluating the positive class alone is different from pooling or macro-averaging both columns. For example, with true classes (Dog, Dog, Cat) and predicted classes (Dog, Cat, Cat), taking Dog as positive yields
\[
\mathrm{Precision}_{\mathrm{Dog}}=1,\qquad
\mathrm{Precision}^{\mathrm{micro}}=\frac23,\qquad
\mathrm{Precision}^{\mathrm{macro}}=\frac34
\]
when micro and macro are computed using the two-column encoding. The difference concerns the choice of aggregation, not an inconsistency between the encodings; see Definitions~\ref{def:indicator} and~\ref{def:aggregation}.

The next example shows why Theorem~\ref{thm:micro_macro} needs additivity. Consider $M(\mathbold{v})=\bigl(\mathrm{Recall}(\mathbold{v})\bigr)^2=tp^2/(tp+fn)^2$. For two labels with count vectors $\mathbold{v}_1=(1,0,1,2)^\top$ and $\mathbold{v}_2=(3,0,1,0)^\top$, the per-label scores are $M_1=1/4$ and $M_2=9/16$, with denominators $\mathrm{den}_1=4$ and $\mathrm{den}_2=16$. Pooling first and denominator-weighted averaging give, respectively,
\[
\begin{aligned}
M^{\mathrm{micro}}&=\left(\frac{1+3}{(1+1)+(3+1)}\right)^2=\frac49,\\
\frac{\sum_{j=1}^{2}\mathrm{den}_jM_j}{\sum_{j=1}^{2}\mathrm{den}_j}&=\frac{4(1/4)+16(9/16)}{4+16}=\frac12\neq\frac49.
\end{aligned}
\]
The squared numerator and denominator are not additive in the count vectors, so this measure lies outside the hypothesis of Theorem~\ref{thm:micro_macro}; merely expressing a measure as a ratio is insufficient.

The last example illustrates the cancellation in Corollary~\ref{cor:micro_skew_base}. Take two mutually exclusive actual labels with prevalences $p$ and $1-p$, where $0<p<1$, but allow multilabel predictions, which need not be one-hot. Keep the class-conditional rates fixed at $\mathrm{rec}_1=\mathrm{spec}_1=0.9$ and $\mathrm{rec}_2=\mathrm{spec}_2=0.1$. The positive-class weights are $(p,1-p)$ and the negative-class weights are $(1-p,p)$, so
\[
\begin{aligned}
\mathrm{rec}_{\mathrm{micro}}&=0.9p+0.1(1-p)=0.1+0.8p,\\
\mathrm{spec}_{\mathrm{micro}}&=0.9(1-p)+0.1p=0.9-0.8p,\\
\mathrm{BA}_{\mathrm{micro}}&=\frac12\bigl(\mathrm{rec}_{\mathrm{micro}}+\mathrm{spec}_{\mathrm{micro}}\bigr)=\frac12.
\end{aligned}
\]
For $p=1/4$ the pooled $(\mathrm{rec},\mathrm{spec})$ pair is $(0.3,0.7)$, whereas for $p=3/4$ it is $(0.7,0.3)$; balanced accuracy remains $0.5$. Thus unequal class-specific rates do not by themselves rule out micro skew-invariance.

\section{Examples for Corollary~\ref{cor:savage_cases}(c)}
\label{app:savage_examples}

The standard squared loss has $w(t)\equiv2$.
There also exist nonquadratic strictly proper examples: choosing
\[
 w(t)=1+\left(t-\tfrac12\right)^2
 \quad\text{gives}\quad
 \phi(t)=\tfrac58|t|^2-\tfrac13|t|^3+\tfrac14|t|^4
 \quad(|t|\le1).
\]
Here $\phi'(t)=t\,w(t)$ for $t>0$, and
$\phi''(t)=3(t-\tfrac13)^2+\tfrac{11}{12}>0$ there, so $\phi$ is
even and convex, while $w>0$ is symmetric.
It is not proportional to $t^2$.
The extended-valued log-loss example of Corollary~\ref{cor:savage_cases}(b) has
$w(t)=1/[t(1-t)]$, which is also symmetric and strictly positive,
although its $\phi$ lies outside the finite-valued assumptions of (c).
Finally, for the convex power losses $\phi(t)=|t|^s$ with $s\ge1$,
$w(t)=s\,t^{s-2}$ is symmetric about $1/2$ for every $0<t<1$
only when $s=2$. Thus MAE, with $w(t)=1/t$, and quartic error,
with $w(t)=4t^2$, are not proper; for the quartic case the
first-order condition is satisfied only at $p=1/2$.

\section{Proofs for Section~\ref{sec:calibration}}
\label{app:proofs_cal}

\subsection*{Proof of Theorem~\ref{thm:cal_skew}}
\textbf{(a) ($\Leftarrow$)} By Definition~\ref{def:cal_skew}, a
skew transformation rescales the subpopulation sizes while
keeping the within-class distributions of $\hat{Y}_{i,j}$
unchanged; any function of $\{F_j^{+}, F_j^{-}\}$ is therefore
invariant.

\textbf{(a) ($\Rightarrow$)} By the conventions above, a skew
transformation changes only the counts $(n^+_j, n^-_j)$ and leaves every
$F_j^{\pm}$ fixed, so it preserves $\{F_j^{+}, F_j^{-}\}$; conversely, two data
sets with the same $\{F_j^{+}, F_j^{-}\}$ and different counts are related by a
finite chain of skew transformations (class by class, through a common multiple of
the counts). The relation ``related by a chain of skew transformations'' is
therefore exactly ``having the same $\{F_j^{+}, F_j^{-}\}$''. A skew-invariant $M$
is constant on each equivalence class of this relation, hence a function of the
class representative $\{F_j^{+}, F_j^{-}\}$.

\textbf{(b) ($\Leftarrow$)} $\alpha_j$ and $\beta_j$ are the
means of $(1-\hat{y})^{2}$ under $F_j^{+}$ and of $\hat{y}^{2}$
under $F_j^{-}$, hence functions of $\{F_j^{+}, F_j^{-}\}$;
invariance follows from part (a).

\textbf{(b) ($\Rightarrow$)} A Calibration Framework measure is,
by hypothesis, a function
$M = g(\{\pi_j, \alpha_j, \beta_j\})$. Under rescaling
$(\lambda, \mu)$, the effective prevalence becomes
$\pi'_j = \lambda n^+_j / (\lambda n^+_j + \mu n^-_j)$ while
$(\alpha_j, \beta_j)$ is held fixed; as $\lambda/\mu$ ranges over
$(0,\infty)$, $\pi'_j$ sweeps the whole interval $(0,1)$.
Skew-invariance forces $g$ to be constant in each $\pi_j$
argument, class by class, on the set of attained values of
$(\pi_j,\alpha_j,\beta_j)$ (nothing is asserted outside what data can realise),
leaving a function of $\{\alpha_j, \beta_j\}$ alone. In particular, by
Proposition~\ref{prop:bs_decomp},
\begin{equation}
  \mathrm{BS}_{j}(\pi'_j)
  \;=\; \pi'_j\,\alpha_j + (1-\pi'_j)\,\beta_j
  \label{eq:bs_pi_prime}
\end{equation}
sweeps the open interval between $\alpha_j$ and $\beta_j$
whenever $\alpha_j \neq \beta_j$, which is why the per-class
Brier score itself is skew-sensitive
(Corollary~\ref{cor:bs_skew}).

\subsection*{Proof of Corollary~\ref{cor:murphy_skew}}
\textit{(a)} Immediate from the formula $\pi_j(1-\pi_j)$.

\textit{(b)} By Bayes' theorem,
\[
  f_j(p)
  = \frac{P(\hat{Y}_{i,j}=p\mid Y_{i,j}=1)\,\pi_j}
         {P(\hat{Y}_{i,j}=p\mid Y_{i,j}=1)\,\pi_j
          + P(\hat{Y}_{i,j}=p\mid Y_{i,j}=0)\,(1-\pi_j)}.
\]
Let $a_p$ and $b_p$ be the probabilities of the predicted value $p$ under the
within-class score distributions $F_j^{+}$ and $F_j^{-}$, so that the marginal
probability of $p$ is $P(p)=\pi_ja_p+(1-\pi_j)b_p$ and
$f_j(p)=\pi_ja_p/P(p)$. Since
$f_j(p)-\pi_j=\pi_j(1-\pi_j)(a_p-b_p)/P(p)$,
\[
  \mathrm{RES}_j=\sum_pP(p)\,(f_j(p)-\pi_j)^2
  =\pi_j^2(1-\pi_j)^2\sum_p\frac{(a_p-b_p)^2}{\pi_ja_p+(1-\pi_j)b_p},
\]
the sum running over the values $p$ with $P(p)>0$. Hence $\mathrm{RES}_j=0$ at every prevalence if and only if
$a_p=b_p$ for every $p$, equivalently $F_j^+=F_j^-$. This is the
no-separation case, in which positive and negative examples have
identical score distributions; no claim about a likelihood ratio is
needed (and $a_p/b_p$ may be undefined where $b_p=0$). Otherwise $\mathrm{RES}_j>0$ for every $\pi_j\in(0,1)$, while
$\mathrm{RES}_j\le\mathrm{Var}(Y_{i,j})=\pi_j(1-\pi_j)\to0$ as $\pi_j\to0$ or
$1$; so $\mathrm{RES}_j$ is not constant in $\pi_j$ and is skew-sensitive.

\textit{(c)} The calibration map is a conditional frequency, so it is a
function of the class balance: for a predicted value $p$ attained by both
classes ($a_p,b_p>0$), replicating the positive examples $\lambda$-fold gives
$f_j^{(\lambda)}(p)=\lambda\pi_ja_p/(\lambda\pi_ja_p+(1-\pi_j)b_p)$, strictly
increasing in $\lambda$. If $f_j^{(1)}(p)=p$, then $f_j^{(\lambda)}(p)\neq p$ for
$\lambda\neq1$, so some term of $\mathrm{REL}_j$ becomes positive. For a value $p$
attained by positives only, $f_j(p)=1$ for every $\lambda$; by negatives only,
$f_j(p)=0$. Hence calibration at two different prevalences forces every predicted value to be
$1$ (positive-only) or $0$ (negative-only); conversely, if every value is of this kind,
$f_j(p)=p$ at every prevalence, so $\mathrm{REL}_j=0$ throughout. For a concrete instance,
let three examples have score $2/3$, two of them positive, and three have score
$1/3$, one of them positive. Then $f_j(\hat{Y})=\hat{Y}$ and $\mathrm{REL}_j=0$.
Doubling the positives gives $f_j(2/3)=4/5$ and $f_j(1/3)=1/2$, and
$\mathrm{REL}_j=\frac19\bigl[5(\tfrac23-\tfrac45)^2+4(\tfrac13-\tfrac12)^2\bigr]=1/45$. Calibration is a property of the classifier
\emph{together with} the class balance of the data it is evaluated on, which is
why the skew-invariant summary is $(\alpha_j,\beta_j)$ and not a reliability term.

\section{Proofs for Section~\ref{sec:ranking}}
\label{app:proofs_rank}

\subsection*{Proof of Theorem~\ref{thm:dcg_collapse}}
\textit{(a)}
For raw DCG, by finite-sum rearrangement,
\begin{align*}
\mathrm{DCG}^{\mathrm{micro}}
&=\frac1{nm}\sum_{i,j}D_{ij}
=\frac1m\sum_j\left(\frac1n\sum_iD_{ij}\right)
=\frac1m\sum_j\mathrm{DCG}_j^{\mathrm{class}},\\
\mathrm{DCG}^{\mathrm{micro}}
&=\frac1n\sum_i\left(\frac1m\sum_jD_{ij}\right)
=\frac1n\sum_i\mathrm{DCG}_i^{\mathrm{ex}}.
\end{align*}
This is the same finite-sum argument as for Brier
(Theorem~\ref{thm:brier_collapse}).

\textit{(b)}
We compare micro and exemplar nDCG. Since $S_i=b_ir_i$, the difference between the weighted and
unweighted averages is
\[
\mathrm{nDCG}^{\mathrm{micro}}-\frac1n\sum_i r_i
=\frac{\sum_i b_ir_i}{\sum_i b_i}-\frac1n\sum_i r_i
=\frac{\widehat{\mathrm{Cov}}(b,r)}{\bar b},
\quad \bar b=\frac1n\sum_i b_i>0.
\]
Thus equality is equivalent to (\ref{eq:ndcg_cov}). If the $k_i$
are equal, so are the $b_i=\sum_{r=1}^{k_i}w(r)$, and the covariance
vanishes. Constant $r_i$ or other zero-covariance configurations
also yield equality.

\textit{(c)}
For micro versus class-mean nDCG, by the definitions,
\[
\mathrm{nDCG}^{\mathrm{micro}}=\frac{\sum_jT_j}{B},
\qquad \frac1m\sum_j\mathrm{nDCG}_j^{\mathrm{class}}
=\frac1m\sum_j\frac{T_j}{N_j}.
\]
Their difference is the left-hand side of (\ref{eq:ndcg_class_micro}).
If it vanishes for all freely chosen $T_j$, each coefficient
$1/B-1/(mN_j)$ must vanish, so $mN_j=B$ for all $j$;
the converse is immediate. Summing over $j$ then gives
$\sum_jN_j=B$. Yet
\[
\sum_jN_j=w(1)\sum_i k_i,\qquad
B=\sum_i\sum_{r=1}^{k_i}w(r)\le w(1)\sum_i k_i,
\]
with strict inequality when some $k_i\ge2$ and $w(2)<w(1)$.
This is a contradiction. The coefficient argument concerns the
linear extension to freely chosen class sums; the actual DCG sums
come from rankings and need not be freely variable.

\textit{(d)}
The class mean and the exemplar mean are $\sum_{i,j}D_{ij}/(mN_j)$ and
$\sum_{i,j}D_{ij}/(nb_i)$, respectively. Their difference gives
(\ref{eq:ndcg_class_ex}). As $D_{ij}=0$ whenever $Y_{ij}=0$,
the identity holds for every freely chosen relevant-cell matrix
if and only if the coefficients vanish at every relevant $(i,j)$:
$mN_j=nb_i$. Substituting $N_j=n_j^+w(1)$ gives
\[
\frac1{n_j^+}=\frac{mw(1)}{nb_i}\qquad(Y_{ij}=1).
\]
Summing over relevant cells in two ways, with every class nonempty,
yields
\[
m=\sum_j\sum_{i:Y_{ij}=1}\frac1{n_j^+}
=\frac mn\sum_i\frac{k_iw(1)}{b_i},
\quad\text{hence}\quad
\sum_i\left(\frac{k_iw(1)}{b_i}-1\right)=0.
\]
Each summand is nonnegative because $b_i\le k_iw(1)$,
and one is strictly positive if $k_i\ge2$ and $w(2)<w(1)$.
This contradiction establishes the last claim. As in (c),
the universal-coefficient statement is for admissibly supported
cell values, not an assumption that arbitrary such values can
all arise from rank vectors.

For a concrete non-collapse witness, take the usual $w(r)=1/\log_2(r+1)$ and consider
\[
Y=\begin{pmatrix}0&0&1\\1&1&0\end{pmatrix},\qquad
\hat Y=\begin{pmatrix}.9&.6&.2\\.9&.6&.2\end{pmatrix}.
\]
Here the two rows have strict ranks $(1,2,3)$. Direct evaluation
gives $\mathrm{nDCG}^{\mathrm{micro}}\approx0.809953$,
$\frac13\sum_j\mathrm{nDCG}^{\mathrm{class}}_j\approx0.710310$,
and $\frac12\sum_i\mathrm{nDCG}_i=0.750000$.
Thus the absence of a general collapse also occurs for
rank-generated cells, not just free cell matrices.

\subsection*{Proof of Theorem~\ref{thm:auc_collapse}}
\noindent\textit{(a)} For the raw collapse, by Definition~\ref{def:auc_family},
$\Upsigma_{\mathrm{1}}(\mathbold{A})=\sum_{i,j}A_{ij}$, while
$\sum_i A_{ij}=n_j^+\mathrm{AUC}_j$ and
$\sum_j A_{ij}=k_i\mathrm{AUC}_i^{\mathrm{ex}}$.
Divide the two ways of summing by $nm$ and use $\pi_j=n_j^+/n$ to obtain~\eqref{eq:auc_raw_collapse}.

\noindent\textit{(b)} For micro versus macro, by~\eqref{eq:auc_micro},
\[
\mathrm{AUC}^{\mathrm{micro}}
=\frac{\sum_j n_j^+\mathrm{AUC}_j}{\sum_j n_j^+},
\qquad
\overline{\mathrm{AUC}}^{\mathrm{class}}=\frac1m\sum_j\mathrm{AUC}_j.
\]
Writing $\overline{n^+}=m^{-1}\sum_j n_j^+>0$, their difference equals
$\widehat{\mathrm{Cov}}_j(n_j^+,\mathrm{AUC}_j)/\overline{n^+}$.
It vanishes exactly when the indicated covariance is zero; equal positive-class sizes are one sufficient case.

\noindent\textit{(c)} For micro versus exemplar mean, similarly,
\[
\mathrm{AUC}^{\mathrm{micro}}=\frac{\sum_i k_i\mathrm{AUC}_i^{\mathrm{ex}}}{\sum_i k_i},
\qquad \overline{\mathrm{AUC}}^{\mathrm{ex}}=\frac1n\sum_i\mathrm{AUC}_i^{\mathrm{ex}}.
\]
With $\bar k=n^{-1}\sum_i k_i>0$, their difference is
$\widehat{\mathrm{Cov}}_i(k_i,\mathrm{AUC}_i^{\mathrm{ex}})/\bar k$.
In particular, equal row cardinalities suffice for equality.

\noindent\textit{(d)} For macro versus exemplar mean, expanding the definitions gives
\[
\overline{\mathrm{AUC}}^{\mathrm{class}}
-\overline{\mathrm{AUC}}^{\mathrm{ex}}
=\sum_{i,j:Y_{ij}=1} A_{ij}
\left(\frac1{mn_j^+}-\frac1{nk_i}\right),
\]
proving~\eqref{eq:auc_class_ex_eq}. If the expression vanishes for \emph{every} supported real cell array, choose a single nonzero positive cell at a time: each coefficient must vanish, yielding $mn_j^+=nk_i$ on that support. The converse follows by substitution. Unlike the analogous nDCG restriction, these support conditions can hold on unbalanced data: with $n=m=4$ and positive-label sets $\{a,b\},\{a,b\},\{c\},\{d\}$, the row and column counts are both $(2,2,1,1)$, and the condition holds on every relevant cell.

\section{Proof of Proposition~\ref{prop:auc_improper}(c)--(d)}
\label{app:auc_improper_proof}

The pair weights $b_{ii'}$ are those defined in the proof of Proposition~\ref{prop:auc_improper}(b);
a report ordering $q_i>q_{i'}$ receives $b_{ii'}$ from the ordered pair.

\noindent\textit{(c)} We treat small samples without independence. For $n=2$, on $Y_i=1,Y_{i'}=0$, the denominator is $n^+n^-=1$. Thus $b_{ii'}-b_{i'i}=p_i-p_{i'}$, without independence. For $n=3$, the remaining label $z\in\{0,1\}$ gives
\[
 n^+n^-=(1+z)(2-z)=2
\]
under either orientation of the pair. Therefore $b_{ii'}-b_{i'i}=\tfrac12(p_i-p_{i'})$ under any joint law. Sorting by $p_i$ again maximises each pair contribution.

\noindent\textit{(d)} We give a dependent counterexample and then extend it. For $n=4$, consider the five label vectors and probability weights below:
\[
\begin{array}{c|ccccc}
 Y & (0,0,0,1)&(0,1,1,0)&(1,0,1,0)&(1,0,1,1)&(1,1,0,1)\\\hline
 11P(Y) &1&2&3&2&3
\end{array}
\]
Their marginal probabilities are $p=(8,5,7,6)/11$, all distinct. Directly summing the conditional AUC values gives
\[
\mathbb E[\mathrm{AUC}(1\succ3\succ4\succ2,Y)]=\frac{41}{66},\qquad
\mathbb E[\mathrm{AUC}(1\succ4\succ3\succ2,Y)]=\frac{83}{132}.
\]
The second ordering exceeds the truthful first ordering by $1/132>0$. For any $n=4+k$ with $k\ge0$, append $k$ examples that are positive with certainty and rank them above the original four in both reports. Only the relative order of indices $3,4$ differs, and the expected AUC for the truthful ordering minus the alternative is
\begin{align*}
D_k&=\frac1{11}\left(-\frac1{3(k+1)}+\frac5{2(k+2)}-\frac3{k+3}\right)\\
&=-\frac{5k^2+4k+3}{66(k+1)(k+2)(k+3)}<0.
\end{align*}
Finally, perturb the added certain-positive labels to independent Bernoulli variables with distinct probabilities $1-\varepsilon_l$ sufficiently close to $1$ (and different from the original four marginals). By continuity of the finite-scenario expected AUC, the strict inequality persists, now with all marginals distinct. This establishes the claim for every $n\ge4$.

\section{Ordinal Labels and Quadratic Weighted Kappa}
\label{app:qwk}

\begin{remark}[Ordinal extension and quadratic weighted kappa]
\label{rem:qwk}

The squared-error construction underlying Theorem~\ref{thm:bridge} also admits an ordinal multiclass extension, providing a connection to quadratic weighted kappa (QWK)~\citep{cohen1960coefficient}. The connection has four parts.

\begin{enumerate}
\item For a $K$-class ordinal problem with $K\geq 2$, one-hot encoding $\mathbold{Y},\mathbold{\hat{Y}}\in\{0,1\}^{n\times K}$, and ordinal levels indexed by $k=0,1,\ldots,K-1$, define the true and predicted ordinal label indices as
$y_i=\sum_{k=0}^{K-1}k\,Y_{ik}$ and
$\hat y_i=\sum_{k=0}^{K-1}k\,\hat Y_{ik}$.
Thus $y_i,\hat y_i\in\{0,\ldots,K-1\}$, and the maximum squared disagreement is $(K-1)^2$.

\item Applying the squared-error deviation to these ordinal indices gives the normalised per-example disagreement
\begin{equation}
  D_{\mathrm{ord}}(Y_{i,\cdot},\hat Y_{i,\cdot})
  \;=\;\frac{(y_i-\hat y_i)^2}{(K-1)^2}
  \;=\;\sum_{k=0}^{K-1}\sum_{l=0}^{K-1}
  \frac{(k-l)^2}{(K-1)^2}\,Y_{ik}\hat Y_{il},
  \label{eq:qwk_cell}
\end{equation}
where the second equality follows from the one-hot encoding. The average $\frac{1}{n}\sum_iD_{\mathrm{ord}}$ is exactly the \emph{observed weighted disagreement} in the QWK numerator.

\item Full QWK takes the form
\begin{equation}
  \kappa_w\;=\;1-\frac{\dfrac{1}{n}\displaystyle\sum_iD_{\mathrm{ord}}}
   {\displaystyle\sum_{k=0}^{K-1}\sum_{l=0}^{K-1}
    \frac{(k-l)^2}{(K-1)^2}\cdot\frac{n_k}{n}\cdot\frac{m_l}{n}},
  \label{eq:qwk_full}
\end{equation}
where $n_k=\sum_iY_{ik}$ and $m_l=\sum_i\hat Y_{il}$ are the column sums of the one-hot matrices. The expression is defined as written only when the expected weighted disagreement $E_w$, namely the denominator in~(\ref{eq:qwk_full}), is strictly positive. When $E_w=0$, a separate convention would be needed to assign a value to QWK.

\item The structure of QWK is analogous to the treatment of Cohen's $\kappa$ in \citet{prati2026unified}. In both cases the \emph{observed} term is a cell-operator aggregation (a coincidence aggregation for $\kappa$, an ordinal deviation aggregation for $\kappa_w$), whereas the \emph{expected-under-independence} term is computed from products of column marginals of $\mathbold Y$ and $\mathbold{\hat Y}$. As noted in \citet{prati2026unified} for $\kappa$, these marginals can be calculated as column sums of the indicator matrices and tracked as auxiliary quantities alongside the standard cell aggregations. QWK is generally skew-sensitive: changes in class prevalence can alter the observed weighted disagreement and its expected-under-independence normaliser, so invariance under prevalence shifts is not guaranteed.
\end{enumerate}

\end{remark}

% ============================================================
%  Bibliography
%  (bibliographystyle is set automatically by jmlr2e to plainnat)
% ============================================================
\bibliography{refs}

\end{document}